\documentclass{article} 
\usepackage{iclr2024_conference,times}

\usepackage{amsmath,amsfonts,bm}

\def\Figref#1{Figure~\ref{#1}}

\def\Secref#1{Section~\ref{#1}}

\def\eqref#1{equation~\ref{#1}}

\def\1{\bm{1}}

\DeclareMathAlphabet{\mathsfit}{\encodingdefault}{\sfdefault}{m}{sl}
\SetMathAlphabet{\mathsfit}{bold}{\encodingdefault}{\sfdefault}{bx}{n}

\usepackage{hyperref}
\usepackage{url}
\usepackage{graphicx}
\usepackage{tikz}
\usepackage{caption}
\usepackage{subcaption}
\usepackage{amsthm}

\hypersetup{
 colorlinks=true,
 linkcolor=red,
 citecolor=cyan,
 filecolor=magenta,
 urlcolor=cyan,
 }
\usepackage{booktabs}
\usepackage{multirow}
\usepackage{adjustbox}
\usepackage{float}

\title{Beyond Static Evaluations of\\Out-of-Distribution Generalization}
\title{A Comprehensive Evaluation of Out-of-Distribution Generalization needs Multiple Shift Intensities}
\title{Towards a More Comprehensive Evaluation of Out-of-Distribution Generalization under Multiple Shift Intensities}
\title{Robustness May be More Brittle than We\\Think under Different Degrees of\\Distribution Shifts}

\author{Kaican Li \\
Hong Kong University of Science and Technology \\
\texttt{klibf@connect.ust.hk} \\
\And
Yifan Zhang \\
National University of Singapore \\
\texttt{yifan.zhang@u.nus.edu\hphantom{ab}} \\
\And
Lanqing Hong \\
Huawei Noah's Ark Lab \\
\texttt{honglanqing@huawei.com} \\
\And
Zhenguo Li \\
Huawei Noah's Ark Lab \\
\texttt{li.zhenguo@huawei.com\hphantom{ab}} \\
\And
Nevin L.\@ Zhang \\
Hong Kong University of Science and Technology \\
\texttt{lzhang@cse.ust.hk} \\
}

\makepreprint

\newcommand{\comment}[1]{}  

\begin{document}

\maketitle

\begin{abstract}
Out-of-distribution (OOD) generalization is a complicated problem due to the idiosyncrasies of possible distribution shifts between training and test domains.
Most benchmarks employ diverse datasets to address this issue; however, the degree of the distribution shift between the training domains and the test domains of each dataset remains largely fixed.
This may lead to biased conclusions that either underestimate or overestimate the actual OOD performance of a model.
Our study delves into a more nuanced evaluation setting that covers a broad range of shift degrees.
We show that the robustness of models can be quite brittle and inconsistent under different degrees of distribution shifts, and therefore one should be more cautious when drawing conclusions from evaluations under a limited range of degrees.
In addition, we observe that large-scale pre-trained models, such as CLIP, are sensitive to even minute distribution shifts of novel downstream tasks.
This indicates that while pre-trained representations may help improve downstream in-distribution performance, they could have minimal or even adverse effects on generalization in certain OOD scenarios of the downstream task if not used properly.
In light of these findings, we encourage future research to conduct evaluations across a broader range of shift degrees whenever possible.

\end{abstract}

\section{Introduction}

Out-of-distribution (OOD) generalization is vital to the safety and reliability of machine learning applications in the real world.
However, the complexities of distribution shifts between the training domains and the real test domains make OOD generalization a challenging problem.
Numerous empirical studies~\citep{gulrajani2021in, wiles2022a, wenzel2022assaying, idrissi2022imagenet} have suggested that most algorithms only offer very little improvement in OOD performance over empirical risk minimization~(ERM)~\citep{vapnik1998statistical}.
Furthermore, algorithms performing better than ERM against one type of distribution shift often perform poorly against another~\citep{ye2022ood}.
The inconsistency suggests that it is important to consider various possible types of distribution shifts of a task when evaluating the OOD performance of a model; otherwise, the evaluation might lead to biased conclusions.

To address the issue, most OOD benchmarks~\citep{koh2021wilds,hendrycks2021many,gulrajani2021in,ye2022ood} incorporate multiple datasets exhibiting a diverse range of distribution shifts.
However, another potential source of evaluation bias is often overlooked: the test domains of these datasets only capture a largely fixed degree of each distribution shift.
For example, in \citep{li2017deeper,koh2021wilds,he2021towards,zhao2022ood}, each test domain represents a different ``direction'' of the potential distribution shifts of a task but there is no distinction between different degrees of shift on the same direction.
Similar problems can also arise when only the aggregate performance across multiple degrees is examined~\citep{hendrycks2018benchmarking}.
Such kind of evaluation can result in misconceptions about model performance on the same grounds as those of the evaluation based on limited types (or ``directions'') of distribution shifts.

\begin{figure}[t]
    \centering
    \includegraphics[width=0.55\linewidth]{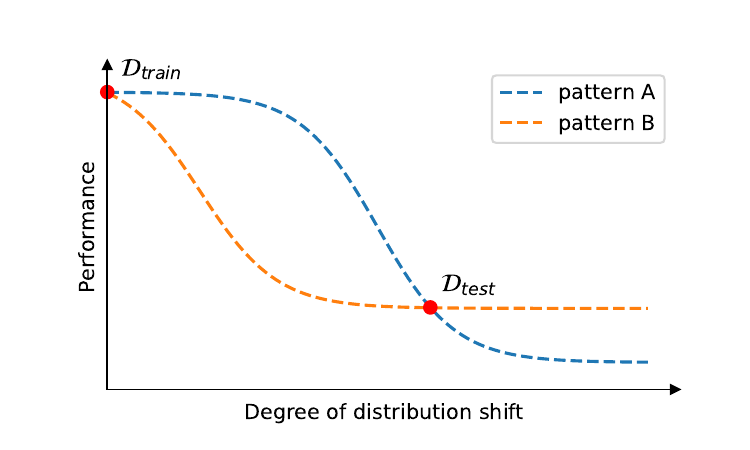}
    \caption{A typical situation where an evaluation under a limited number of degrees of a distribution shift cannot tell any difference between two distinct OOD generalization patterns (labeled as A and B) that can be realized by a model.}
    \label{fig:possible-models}
\end{figure}

Consider the situation (that we observed in this work) illustrated in \Figref{fig:possible-models}, where the performance of a model is evaluated in only two domains, one for in-distribution (ID) performance in the training domain $\mathcal{D}_\text{train}$, and the other for OOD performance in the test domain  $\mathcal{D}_\text{test}$.
In this case, the observed performance, which can be explained by at least two distinct generalization patterns as shown in the figure, presents an oversimplified summary of the OOD generalization ability of the model.
This simplification may lead to incorrect assumptions about model robustness under various degrees.
For example, when a model outperforms another model under a distribution shift of certain severity, it could leave the wrong impression that the first model is more robust in general, i.e., outperforming the other model under almost every possible degree of the concerned distribution shift, while in fact, the first model has much poorer worst-case performance.

In this study, we take a closer look at OOD generalization under distribution shifts of varying degrees.
We are interested in the behavior of different models under a broad range of shift degrees and also the relation between the performance of a model at different degrees.
Through extensive experiments, we make several observations about the generalization behavior of neural networks under the setting.
First, we highlight that the advantage of a model under some distribution shift may not apply to stronger shifts of the same type, even if the shift is just slightly stronger\footnote{We use ``mild/strong'' ``and low/high-degree'' interchangeably when describing a distribution shift.}.
Therefore, caution should be taken when interpreting evaluation results obtained under a limited range of shift degrees.
Second, we find that training a model with strongly shifted data can sometimes guarantee robustness to all milder shifts, while at other times it only has a limited impact on robustness and may even harm the OOD performance under milder shifts.

Lastly, the brittleness of robustness to different degrees of distribution shift is also observed in large-scale pre-trained models.
We find that while CLIP~\citep{radford2021learning} models are able to adapt to many novel tasks, achieving great (sometimes near-perfect) downstream ID performance, they can be extremely sensitive to downstream distribution shifts.
In the presence of a distribution shift that is rarely seen during pre-training, even a very mild degree of the shift can cause a disproportionate performance drop in CLIP models in comparison to models trained from scratch.
Interestingly, further adapting to the shift to which the models are sensitive significantly improves their general robustness.
We believe that such characterizations of the ``growth'' of the OOD generalization ability of a model is a generally good practice.
We encourage future research to adopt this kind of evaluation to generate more valuable insights into OOD generalization.

\section{Related work}
\paragraph{Out-of-distribution (OOD) generalization.} Deep neural networks have demonstrated incredible generalization on a variety of complicated tasks, sometimes exceeding human performance~\citep{russakovsky2015imagenet,silver2017mastering,openai2023gpt}, but they are shown to generalize very differently as we do and are very sensitive to all kinds of distribution shifts~\citep{szegedy2014intriguing,geirhos2020shortcut,wang2023robustness,yu2022dual,zhang2021deep}.
Such brittleness severely undermines the reliability of neural networks and hence limits their applications in the real world where the stakes can be very high.
For this reason, OOD generalization and related areas such as domain generalization~\citep{blanchard2011generalizing,zhou2021domain2,wang2022generalizing} and test-time adaptation~\citep{niu2022EATA,niu2022towards,zhang2022self} have gained much attention rapidly in recent years~\citep{shen2021towards}.

\paragraph{Distribution shifts and OOD benchmarks.} \cite{hendrycks2018benchmarking} proposed a benchmark based on images under different severity levels of common image corruption, however, no analysis was provided at the level of each individual severity level of corruption.
\citet{hendrycks2021many} further proposed OOD benchmarks under natural distribution shifts, but this time like DG benchmarks such as \citep{gulrajani2021in}, they still do not involve any discussion with regard to different degrees of distribution shifts.
\citet{koh2021wilds} proposed a diverse set of OOD benchmarks derived from real-world tasks but the degrees of distribution shifts are still largely fixed.
Similar examples include \citep{peng2019moment,he2021towards,liang2022metashift,zhao2022ood} which focus on incorporating as many diverse types of distribution shifts without considering different degrees of each type of distribution shift.
\citet{lynch2023spawrious} considered three levels of spurious correlation but did not discuss the connection between the model performance at each level.



\paragraph{Distribution shifts in model learning.}
On several datasets with controllable factors, \citet{schott2022visual} showed that models regardless of supervision signal and architectural bias could not learn the underlying mechanism that causes the distribution shifts.
In comparison, we find that learning the shifting-inducing mechanism of certain task is possible.
In a different context, \citet{shi2022robust} also studied OOD generalization under multiple degrees of distribution shift.
Their main focus is whether unsupervised methods can learn more robust representations than supervised learning.
They conducted evaluation against three different degrees of spurious correlation and found that unsupervised methods are generally more robust than supervised learning and the advantage grows as the degree of the distribution shift increases.

\paragraph{Robustness of CLIP.}
Pre-training usually has a great impact on generalization.
Foundation models such as CLIP~\citep{radford2021learning} leverage massive scale of training data to generalize to a great variety of downstream tasks.
At the same level of ID accuracy, zero-shot CLIP models are able to attain much higher OOD accuracy on several ImageNet variants than other models trained with a much smaller scale of data.
Later, it is shown that the main source of the remarkable robustness of CLIP is the diversity of its training data distribution~\citep{fang2022data}.
While CLIP can be made even more robust in some tasks after proper adaptation~\citep{wortsman2022robust}, what remains unclear in the literature is to what extent the robustness of CLIP and other foundation models can transfer to downstream tasks and how the models would behave as the degree of the downstream distribution shifts increases.

\section{Different degrees of distribution shifts}



The degree of a distribution shift can be quantified in many ways.
In this paper, we do not restrict our study to a particular way of quantification as different types of distribution shift may favor different ways of quantification.
Instead, we consider the degrees of only one type of distribution shift at a time, so the degrees can take arbitrary values as long as they preserve a certain ordering of a set of domains under the same type of distribution shift.

For simplicity, we use natural numbers to represent the order of a given set of domains under the same type of distribution shift, with smaller numbers indicating lower degrees of distribution shift.
We use $\mathcal{D}_d$ to denote a domain where $d$ is its degree and refer to $\mathcal{D}_0$ as a clean domain or a domain under no distribution shift.
A model is said to be more robust to a degree $d$ of distribution shift than another model if it attains better performance in $\mathcal{D}_d$.

\begin{figure}[t]
    \centering
    \includegraphics[width=\linewidth]{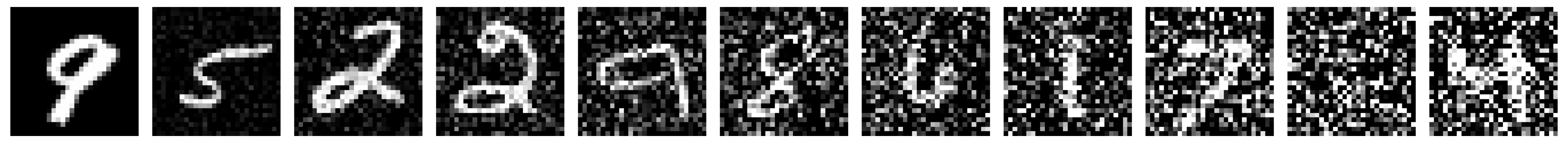}
    \caption{Examples of the \textsc{NoisyMNIST} dataset which consists of 11 subsets of MNIST, one of which is clean, while the other 10 subsets are affected by different degrees of Gaussian noise.}
    \label{fig:noisymnist}
\end{figure}

A simple example of problems under different degrees of distribution shift is shown in \Figref{fig:noisymnist}.
In this example, the degree of distribution shift in a domain corresponds to the intensity of pixel-level Gaussian noise in the images.
In practice, we may not have access to data under all possible degrees of a distribution shift and therefore must rely on data under a limited set of shift degrees to train and evaluate our models.
In this context, out-of-distribution generalization refers to the generalization from data under a given set of shift degrees to the rest of possible degrees.
This setting allows for a nuanced understanding of how different degrees of distribution shifts impact the learning and generalization capabilities of models.
Although we focus on image classification problems and use accuracy to measure model performance, we believe that the conclusions drawn from our experiment results also hold for more general problems.

\section{Seemingly robust models may be more brittle than we think}
\label{sec:scratch}
This section explores the complexities and nuances of model robustness under varying degrees of distribution shifts in neural networks. We first demonstrate that models exhibiting robustness under a certain degree of shift can experience substantial performance degradation under slightly higher degrees of shift. Then we explore whether models perform well under high degrees of shift is robust to lower ones, revealing contrasting results dependent on the specific task, thereby highlighting the inherent brittleness in neural network robustness under different degrees of distribution shift.

\subsection{Experiment setup}


\paragraph{Datasets.} The study in this section employs two altered versions of the MNIST dataset~\citep{lecun2010mnist}, herein referred to as \textsc{NoisyMNIST} and \textsc{RotatedMNIST}.
The \textsc{NoisyMNIST} dataset is generated by introducing Gaussian noise to the original images, resulting in 10 shifted domains under different degrees.
More specifically, the standard deviation of the noise is linearly spaced between 0 and 0.8, in increments of 0.08, at the pixel level, normalized to the pixel value range of 0 to 1. Any pixel value beyond this range is clipped to fit within the 0-1 boundary.
The \textsc{RotatedMNIST} dataset is created by rotating the original images, with degrees linearly spaced from 0 to 80, at intervals of 10 degrees, resulting in 8 shifted domains.
Note that our \textsc{RotatedMNIST} is different from the one in \citep{gulrajani2021in} which covers a smaller range of rotations.

Our study also employs an altered version of CIFAR10~\citep{krizhevsky2009learning}.
We call this dataset \textsc{LowLightCIFAR10} since the distribution shift in this dataset is a combination of two primitive types of distribution shifts which are shifts in brightness and shot-noise intensity.
Photos captured in darker environments tend to exhibit more intense shot noises, and hence this dataset simulates realistic photographic effects in photos captured under low-light conditions and hence is much more realistic than the MNIST variants.
We focus on \textsc{NoisyMNIST} and \textsc{LowLightCIFAR10} in Section~\ref{sec:lower-to-higher}, and further include \textsc{RotatedMNIST} in Section~\ref{sec:higher-to-lower}.

\paragraph{Algorithms.}
We experiment with Empirical Risk Minimization (\textbf{ERM}, \citealp{vapnik1998statistical}) and over 20 Domain Generalization (DG) algorithms, including but not limited to Invariant Risk Minimization (\textbf{IRM}, \citealp{arjovsky2019invariant}), Variance Risk Extrapolation (\textbf{VREx}, \citealp{krueger2020outofdistribution}), Spectral Decoupling (\textbf{SD}, \citealp{pezeshki2021gradient}), Deep Correlation Alignment (\textbf{CORAL}, \citealp{sun2017correlation}), Group Distributionally Robust Optimization (\textbf{GroupDRO}, \citealp{sagawa2019distributionally}), Representation Self Challenging (\textbf{RSC}, \citealp{huang2020self}), Domain-Adversarial Neural Networks (\textbf{DANN}, \citealp{ganin2016domain}), Inter-domain Mixup (\textbf{Mixup}, \citealp{yan2020improve}), and many others in various directions (see the full list in Appendix~\ref{app:algos}).




\subsection{Robustness may not even extrapolate to slightly higher degrees}
\label{sec:lower-to-higher}
\begin{figure}[t]
    \centering
    \includegraphics[width=0.6867\linewidth]{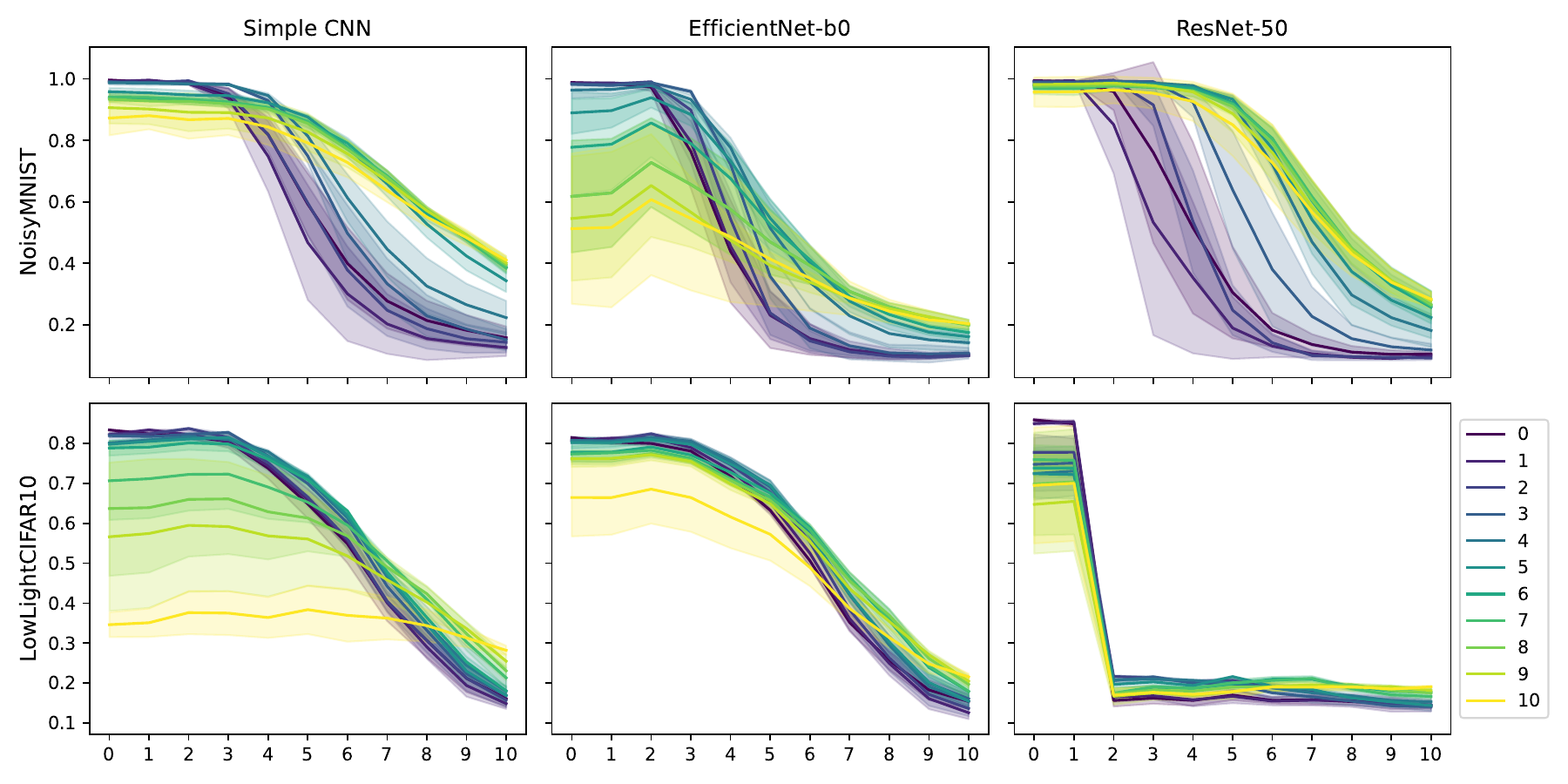}
    \includegraphics[width=0.3033\linewidth]{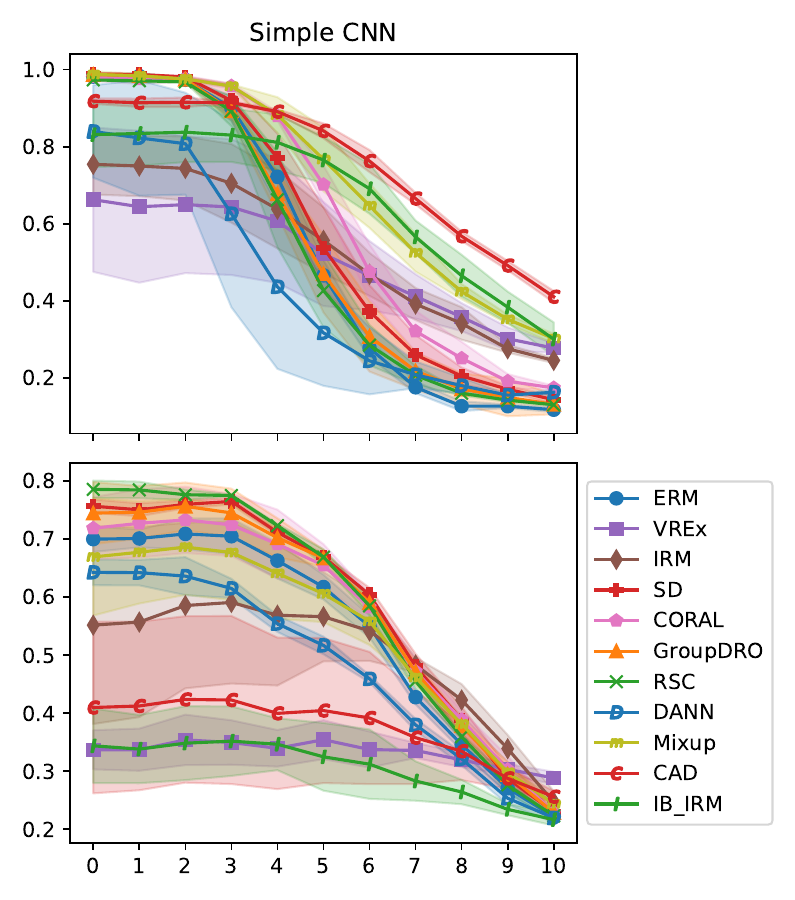}
    \caption{\textbf{(Left)} Performance of the best-performing models at each degree of \textsc{NoisyMNIST} and \textsc{LowLightCIFAR10}.
    The label of the curves denotes the domain on which the models perform best compared to the other models.
    The results are averaged over the top-5 models of all algorithms at each degree.
    \textbf{(Right)} Performance of ERM and representative domain generalization algorithms on the two datasets. The results are averaged over the top-3 models of each algorithm, selected by worst-domain performance.}
    \label{fig:dg}
\end{figure}

Data under strong distribution shifts are usually very rare in the real world.
We often face situations where we only have access to a reasonable amount of data under relatively mild distribution shifts.
With these data, we can be fairly certain about the performance of a model in mild situations, but this is hardly satisfactory for any application that demands a certain level of reliability also in worse situations.
Therefore, an important question is: how much can the performance of a model under some distribution shift tell us about its performance under stronger shifts?

To approach the question, we first constructed a dataset, \textsc{NoisyMNIST}, by gradually adding Gaussian noise to MNIST~\citep{lecun2010mnist}.
As illustrated in \Figref{fig:noisymnist}, \textsc{NoisyMNIST} consists of a clean subset $\mathcal{D}_0$ of MNIST and 10 subsets $\{\mathcal{D}_i\}_{i=1}^{10}$ under different degrees of noise.
While the construction process of \textsc{NoisyMNIST} is simple, the dataset is nonetheless representative of a wide range of distribution shifts that gradually corrupt predictive features in an image.
Intuitively, models that are more robust to relatively mild noises should also be more robust to stronger noises, at least to some extent.
If this is true, then in the case of \textsc{NoisyMNIST}, we should be able to rely on domains under only mild shifts such as $\mathcal{D}_4$, to pick the best-performing models in slightly worse domains such as $\mathcal{D}_5$ and $\mathcal{D}_6$ or even much worse domains such as $\mathcal{D}_{10}$.

For this investigation, we trained a pool of models on $\mathcal{D}_0$ and $\mathcal{D}_1$ with ERM and more than 20 domain generalization (DG) algorithms.
The models share the same architecture (a 4-layer CNN with roughly 0.37M parameters) but are trained with different initializations and hyperparameters in addition to the different learning algorithms.
The performance of the best-performing models in each domain is shown in \Figref{fig:dg}~(left).
The result indicates that models that are better under milder shifts are often significantly \emph{worse} than the other models under stronger shifts.
In particular, the average accuracy of the best-performing models in $\mathcal{D}_4$ has dropped by more than 10\% in $\mathcal{D}_5$ which is only under a slightly more intense noise than $\mathcal{D}_4$.
Similar phenomenon is also observed in \textsc{LowLightCIFAR10}.

\begin{table}[t]
    \small
    \centering
    \adjustbox{max width=\textwidth}{
    \begin{tabular}{lcccccccccccc}
        \toprule
        \multirow{3}{*}{\textbf{Algorithm}} & \multicolumn{3}{c}{\textbf{CNN}} & \multicolumn{3}{c}{\textbf{ResNet-50}} \\
        \cmidrule(lr){2-4} \cmidrule(lr){5-7}
         & $\mathcal{D}_4$ & $\mathcal{D}_5$ & $\mathcal{D}_6$ & $\mathcal{D}_4$ & $\mathcal{D}_5$ & $\mathcal{D}_6$\\
        \midrule
        ERM      & 77.8\scriptsize{$\pm$2.8} \small{(0.0)} & 47.7\scriptsize{$\pm$5.2} \small{(38.7)} & 26.5\scriptsize{$\pm$5.0} \small{(66.0)} & 97.4\scriptsize{$\pm$0.3} \small{(0.0)} & 84.0\scriptsize{$\pm$5.5} \small{(13.8)} & 54.8\scriptsize{$\pm$14.6} \small{(43.8)}\\
        VREx     & 90.1\scriptsize{$\pm$1.7} \small{(0.0)} & 74.3\scriptsize{$\pm$5.6} \small{(17.6)} & 53.4\scriptsize{$\pm$6.4} \small{(40.8)} & 64.6\scriptsize{$\pm$1.7} \small{(0.0)} & 32.1\scriptsize{$\pm$2.3} \small{(50.3)} & 17.4\scriptsize{$\pm$0.9} \small{(73.1)}\\
        IRM      & 78.7\scriptsize{$\pm$2.4} \small{(0.0)} & 57.6\scriptsize{$\pm$8.2} \small{(26.8)} & 38.0\scriptsize{$\pm$11.3} \small{(51.7)} & 95.7\scriptsize{$\pm$0.8} \small{(0.0)} & 82.4\scriptsize{$\pm$1.3} \small{(13.9)} & 56.6\scriptsize{$\pm$6.8} \small{(40.9)}\\
        SD       & 81.7\scriptsize{$\pm$1.2} \small{(0.0)} & 57.7\scriptsize{$\pm$2.3} \small{(29.4)} & 35.7\scriptsize{$\pm$2.1} \small{(56.4)} & 97.8\scriptsize{$\pm$0.4} \small{(0.0)} & 92.4\scriptsize{$\pm$2.1} \small{(5.5)} & 76.5\scriptsize{$\pm$6.0} \small{(21.7)}\\
        GroupDRO & 74.0\scriptsize{$\pm$1.7} \small{(0.0)} & 50.3\scriptsize{$\pm$5.7} \small{(32.1)} & 29.9\scriptsize{$\pm$8.4} \small{(59.6)} & 82.4\scriptsize{$\pm$9.0} \small{(0.0)} & 53.1\scriptsize{$\pm$17.7} \small{(35.5)} & 30.5\scriptsize{$\pm$10.4} \small{(63.0)}\\
        RSC      & 84.3\scriptsize{$\pm$4.6} \small{(0.0)} & 61.4\scriptsize{$\pm$7.2} \small{(27.2)} & 39.6\scriptsize{$\pm$7.1} \small{(53.0)} & 88.4\scriptsize{$\pm$4.0} \small{(0.0)} & 64.6\scriptsize{$\pm$8.6} \small{(26.9)} & 39.2\scriptsize{$\pm$8.0} \small{(55.6)}\\
        Mixup    & 93.2\scriptsize{$\pm$0.4} \small{(0.0)} & 84.1\scriptsize{$\pm$2.3} \small{(9.7)} & 69.2\scriptsize{$\pm$1.9} \small{(25.7)} & 85.4\scriptsize{$\pm$3.7} \small{(0.0)} & 49.2\scriptsize{$\pm$16.2} \small{(42.4)} & 26.6\scriptsize{$\pm$13.2} \small{(68.8)}\\
        CAD      & 94.1\scriptsize{$\pm$1.0} \small{(0.0)} & 78.7\scriptsize{$\pm$3.1} \small{(16.3)} & 58.6\scriptsize{$\pm$4.0} \small{(37.7)} & 78.8\scriptsize{$\pm$19.6} \small{(0.0)} & 50.6\scriptsize{$\pm$22.0} \small{(35.8)} & 30.5\scriptsize{$\pm$13.1} \small{(61.4)}\\
        IB-IRM   & 86.1\scriptsize{$\pm$5.6} \small{(0.0)} & 68.9\scriptsize{$\pm$9.7} \small{(19.9)} & 54.6\scriptsize{$\pm$13.3} \small{(36.5)} & 91.0\scriptsize{$\pm$2.9} \small{(0.0)} & 59.5\scriptsize{$\pm$12.0} \small{(34.6)} & 30.1\scriptsize{$\pm$12.3} \small{(66.9)}\\
        \bottomrule
    \end{tabular}}
    \caption{Performance of the best models in $\mathcal{D}_4$ of ERM and representative DG algorithms. The relative performance drops (\%) with respect to the performance in $\mathcal{D}_4$ are shown in the parentheses. All results are averaged over the top-3 models among 20 models with different initialization and hyperparameters for training.}
    \label{tab:brittleness-by-algos}
\end{table}

\begin{figure}[t]
    \centering
    \includegraphics[width=\linewidth]{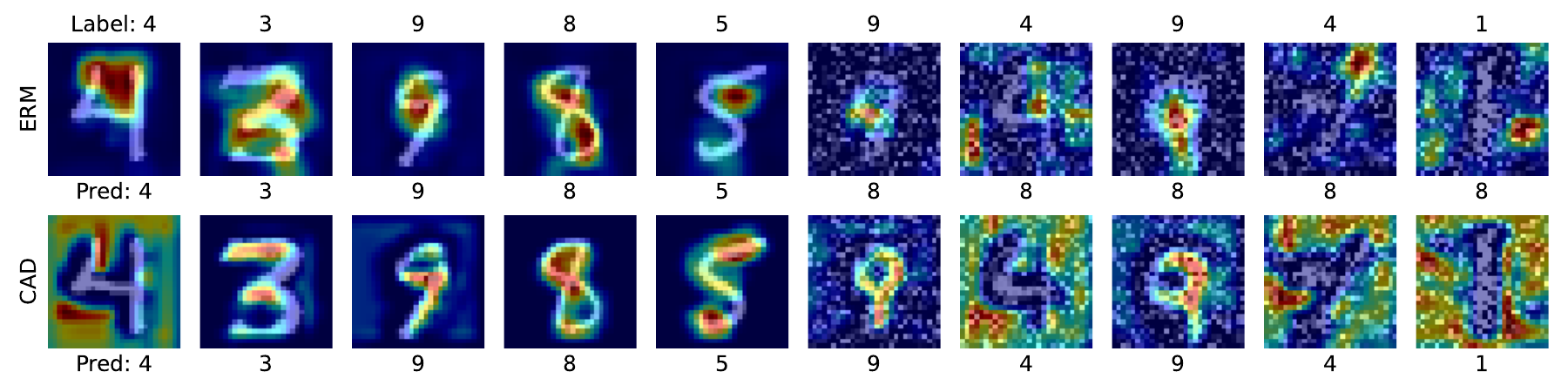}
    \caption{GradCAM visualization of model attention on random examples from $\mathcal{D}_0$ (leftmost) and $\mathcal{D}_7$ (rightmost) of \textsc{NoisyMNIST}. The two models (ERM and CAD) demonstrate distinctive generalization patterns, one relying on the local features while the other more on the global structures.}
    \label{fig:gradcam}
\end{figure}

We further experimented with EfficientNet-b0 ($\sim$5.3M parameters) \citep{tan2019efficientnet} and ResNet-50 ($\sim$25M parameters) \citep{he2016deep} to see if networks with much greater capacities can learn a representation that is generally more robust under all the considered levels of noise.
As shown in \Figref{fig:dg}~(left), the overall pattern still remains, although the difference among the best-performing models under the stronger end of noises has become less significant.
While this shows that larger networks help, the gap may never be closed by increasing the capacity of the network.
More importantly, \emph{the robustness of a model may be more brittle than we think: even under the same type of distribution shift, a slight increase in the degree of the shift may severely harm the performance of the model}.



Besides individual models, the brittleness of robustness under different shift degrees also has implications in evaluating different learning algorithms.
The performance of ERM and representative DG algorithms on \textsc{NoisyMNIST} are shown in \Figref{fig:dg} (right), where the algorithms exhibit very different generalization patterns that cannot be accurately captured by evaluations under only a limited set of shift degrees.
A number of DG algorithms, like ERM, are highly robust to low degrees of distribution shift but are quite brittle in the presence of higher degrees of shift.
In contrast, there are also algorithms that are significantly better than ERM under high shift degrees but are much worse in other cases.
When looking at the best-performing models in $\mathcal{D}_4$, the same brittleness can be generally observed for all the algorithms in Table~\ref{tab:brittleness-by-algos}, where the performance drop can be even more drastic than that is shown in \Figref{fig:dg}.
Astoundingly, the relative performance drop can go up to 50.3\% from $\mathcal{D}_4$ to $\mathcal{D}_5$ and 73.1\% from $\mathcal{D}_4$ to $\mathcal{D}_6$.
Furthermore, we find that adding more training domains can mitigate the issue but still far from resolving it (see Appendix~\ref{app:more-tr-domains}).

To better understand the observed brittleness, we focus on ERM and CAD, the worst-performing and the best-performing algorithms in the worst domain, $\mathcal{D}_{10}$, of \textsc{NoisyMNIST}.
We visualized the attention of the best model of ERM and CAD in terms of worst-domain performance using GradCAM~\citep{selvaraju2017grad}.
As shown in \Figref{fig:gradcam}, while both ERM and CAD models can make highly accurate predictions in the clean domain $\mathcal{D}_0$, they rely on radically different patterns to do so.
ERM prefers the most predictive features regardless of whether they are robust or not.
In the case of \textsc{NoisyMNIST}, these features turn out to be local features, which are easily corrupted by the noise, and thus no longer predictive when the noise becomes intense.

From this perspective, we can see that the brittleness manifests when the spurious correlation between the local features and the target labels reaches a breaking point.
However, where this breaking point is and how rapidly the correlation breaks seem to be totally dependent on the nature of the distribution shift and the task itself.
While \textsc{NoisyMNIST} and \textsc{LowLightCIFAR10} demonstrate two simple cases where the breaking point is at a moderate degree of distribution shift, there can be scenarios where the break happens at a much lower or higher degree of distribution shift and happens much more rapidly.
As a consequence, evaluations that only consider a narrow range of possible shift degrees would be highly unreliable in those scenarios.

\subsection{Robustness at higher degrees does not imply robustness at lower degrees}
\label{sec:higher-to-lower}
We have demonstrated in the previous section that models being more robust to milder distribution shifts does not imply more robustness to stronger distribution shifts, even when the shift is just slightly stronger.
In this section, we shed some light on the converse question: does models performing well under stronger distribution shifts imply them being more robust to milder shifts?

\begin{figure}[t]
    \centering
    \includegraphics[width=0.32\linewidth]{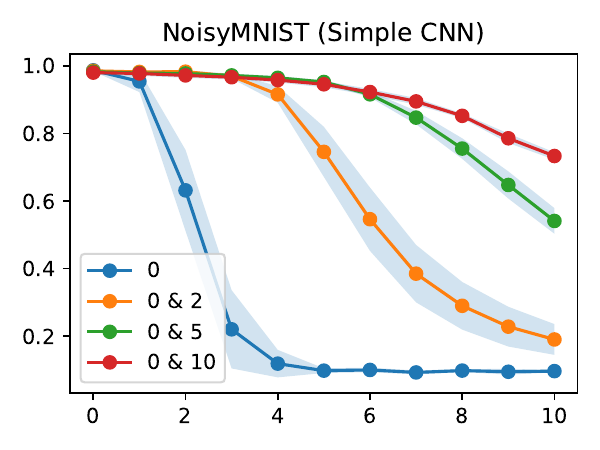}
    \hfill
    \includegraphics[width=0.32\linewidth]{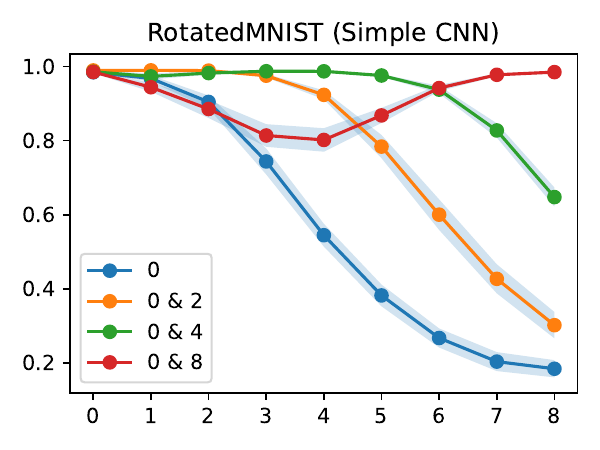}
    \hfill
    \includegraphics[width=0.32\linewidth]{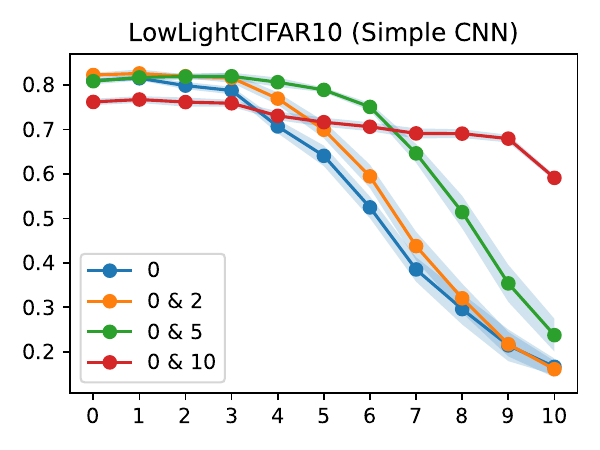}
    \includegraphics[width=0.32\linewidth]{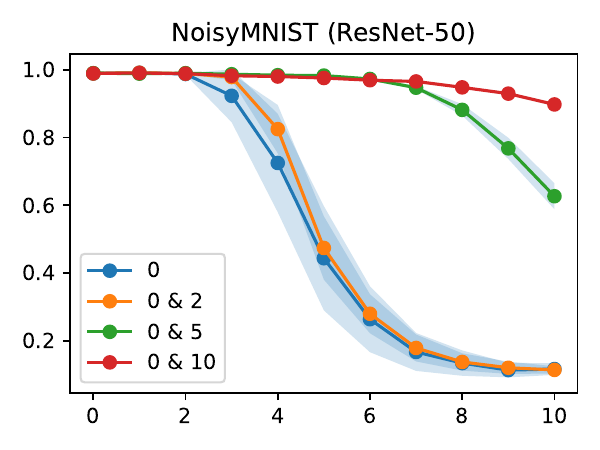}
    \hfill
    \includegraphics[width=0.32\linewidth]{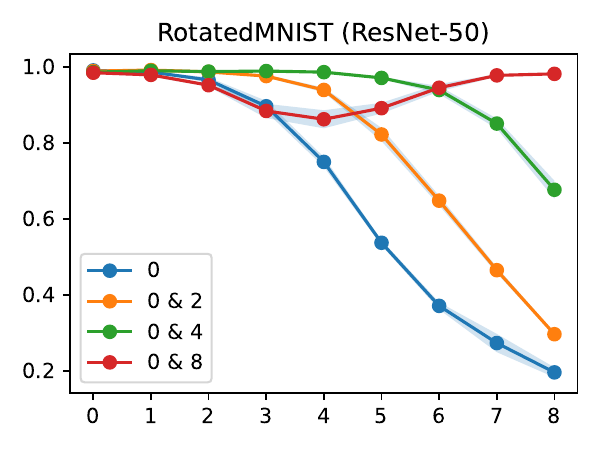}
    \hfill
    \includegraphics[width=0.32\linewidth]{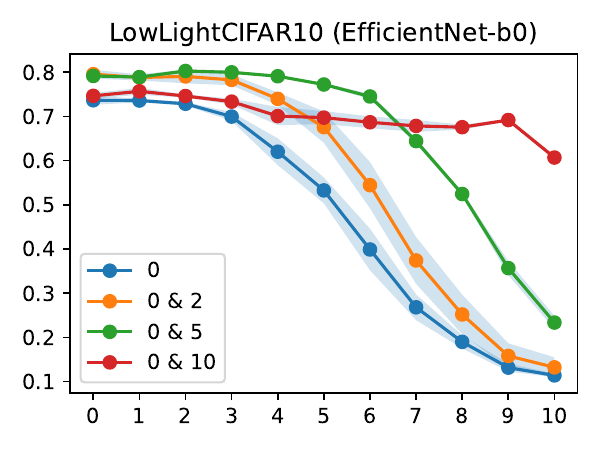}
    \caption{Performance of ERM models trained on domains under different shift degrees. The label of the curves denotes the indices of the training domains, e.g., ``0 \& 2'' means that the models are trained on $\mathcal{D}_0$ and $\mathcal{D}_2$ of the corresponding dataset. The results are averaged over 20 models with different initialization and training data sampling order.}
    \label{fig:tr-envs}
\end{figure}

To start with, we obtained models that are robust to strong distribution shifts by training the models on strongly shifted data together with clean data.
In \Figref{fig:tr-envs}, we compare these models with (i) models trained on much more mildly shifted data (also in addition to clean data) and (ii) models trained on clean data alone.
For \textsc{NoisyMNIST}, the answer to our question is affirmative.
The models that are more robust to stronger shifts are indeed more robust to milder shifts in general.
However, this is not the whole story as the pattern seems to depend on the specific task in consideration.

We experimented with another dataset, \textsc{RotatedMNIST}, which is constructed in a similar fashion to \textsc{NoisyMNIST} while replacing noise with rotation (see \Figref{fig:rmnist} for examples).
On the contrary to \textsc{NoisyMNIST}, robustness against higher shift degrees does \emph{not} guarantee robustness to lower degrees in the case of \textsc{RotatedMNIST} when the shift degree of the additional training data is high.
In comparison with models that trained on only clean data without any rotation, being more robust to the strongest shift may even harm generalization at the mildest degrees.
The results on \textsc{LowLightCIFAR10} demonstrate an even more interesting pattern. When the shift in the training data is not very strong (e.g., 0 \& 5), the generalization pattern follows that of \textsc{NoisyMNIST}---the models are robust to milder shifts. On the other hand, when the shift in the training data is very strong as in the case of 0 \& 10, the pattern looks like that of \textsc{RotatedMNIST}---performance under milder shifts is significantly worse. In particular, even the performance on clean data is compromised.

Notably, the observed patterns are largely consistent between the two model architectures which have a great difference in complexity; however, the improvement brought by model complexity is still far from closing the gap.
DG algorithms are helpful but only to a limited extent (see Fig.~\ref{fig:rmnist-dg}).
Again, this demonstrates the brittleness of the robustness of neural networks.

An important practical implication of the above finding is that, \emph{even for the same type of distribution shift, training on strongly shifted data may not be sufficient to obtain a model that is robust to milder shifts}.
Combined with our finding in \Secref{sec:scratch}, we arrive at the conclusion that the corresponding training data may be necessary to guarantee robustness at a certain degree of distribution shift for some tasks.
Meanwhile, we should also note that there are scenarios where obtaining a dataset under a sufficiently strong shift is able to guarantee robustness to all milder shifts as in the case of \textsc{NoisyMNIST}.
For these kinds of distribution shifts, it may require much less data to guarantee general robustness.

\section{Pre-trained representations are sensitive to novel downstream distribution shifts}
\label{sec:pre-trained}
Pre-training on large-scale datasets is one of the most effective ways that are known to consistently improve the generalization of neural networks across a wide range of tasks~\citep{taori2020measuring, miller2021accuracy}.
In particular, foundation models like CLIP~\citep{radford2021learning} have demonstrated remarkable zero-shot capability on a number of datasets that models trained on much smaller datasets fail to generalize to.
In this section, we further investigate how pre-training would influence the OOD generalization behavior of models on downstream tasks under multiple shift degrees.

\subsection{Experiment setup}

\paragraph{Datasets.}
In addition to \textsc{NoisyMNIST} and \textsc{RotatedMNIST}, we consider two more complicated datasets, \textsc{NoisyImageNet15} and \textsc{LR-ImageNet15}, which are modifications of a 15-category subset of ImageNet on bird species. \textsc{NoisyImageNet15} follows a similar construction to \textsc{NoisyMNIST}, introducing Gaussian noise on the pixel level, linearly spaced between 0 and 0.8, with values clipped to the 0-1 range. Meanwhile, \textsc{LR-ImageNet15} involves altering image resolution, first downsampling via bilinear interpolation and subsequently upsampling to $256{\times}256$, with the downsampled resolution in each domain corresponding to a factor of $0.8^d \cdot 256$, where $d$ represents the degree of distribution shift.
We extend \textsc{RotatedMNIST} to span from 0 to 100 degrees in the experiments of this section.

\paragraph{Pre-trained models.}
We use the implementation of ImageNet pre-trained ResNet-50~\citep{he2016deep} and ViT-B/32~\citep{dosovitskiy2021an} from torchvision, along with CLIP models of the same architectures released by OpenAI\footnote{\url{https://github.com/openai/CLIP}}. They are adapted to downstream tasks through linear probing, following \citep{radford2021learning}.

\subsection{Results}

\begin{figure}[t]
    \centering
    \includegraphics[width=\linewidth]{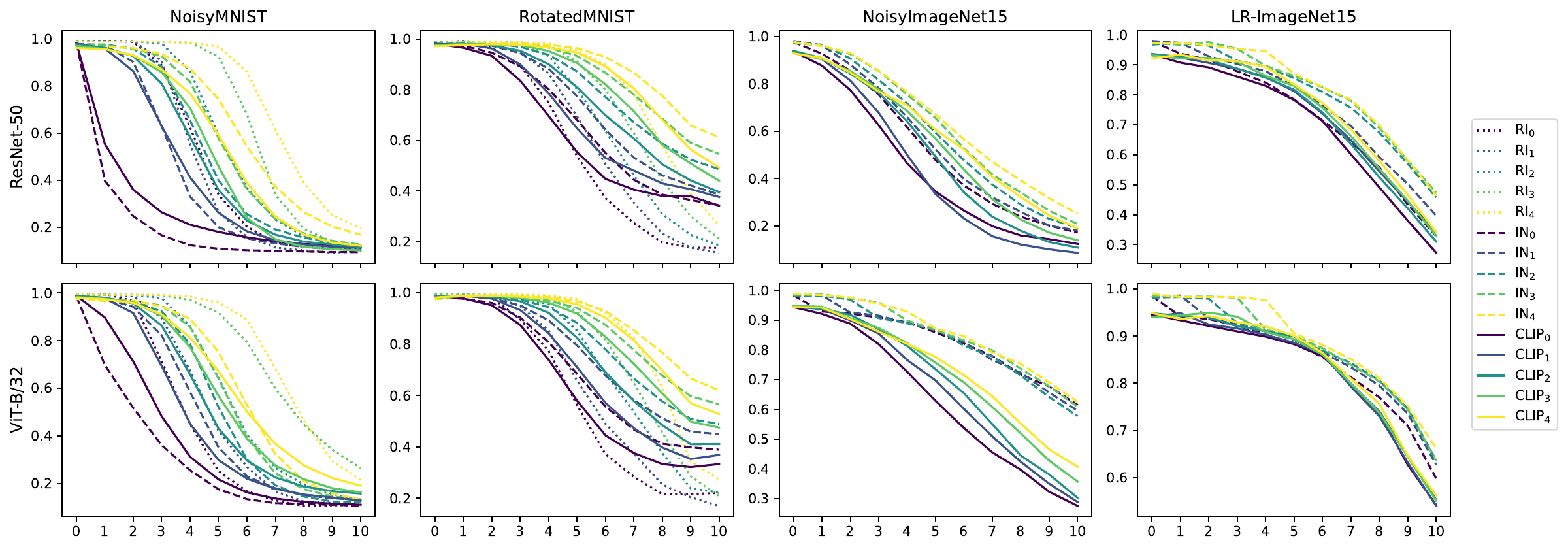}
    \caption{Performance of randomly initialized (\textbf{RI}) models, ImageNet (\textbf{IN}) pre-trained models, and \textbf{CLIP} models on different downstream tasks, evaluated over a broad range of shift degrees. The color of the curves indicates the domains used to train/adapt the models, e.g., RI$_d$ stands for models trained on $\{\mathcal{D}_0, \dots, \mathcal{D}_d\}$ from scratch. The pre-trained models are adapted to the downstream tasks through linear probing. The results are averaged over three runs. Error bars are omitted for clarity (see Appendix~\ref{sec:lp-details} for more details).}
    \label{fig:lp_vs_scratch}
\end{figure}

If large-scale pre-trained models like CLIP have learned generally more robust visual representations, they should be able to support a classifier that has better performance across a broad range of shift degrees than models trained from scratch as well as models trained on much smaller datasets, e.g., ImageNet~\citep{deng2009imagenet}.

In \Figref{fig:lp_vs_scratch}, we compare CLIP models with ImageNet pre-trained models and randomly initialized models that are trained from scratch on the downstream tasks to investigate the robustness of pre-trained representations.
On \textsc{NoisyMNIST}, \emph{although the pre-trained models perform equally well on clean domains as the randomly initialized models, they are surprisingly much more brittle to the distribution shift induced by the noise}.
Notably, the gap of accuracy between \textsc{CLIP$_0$} and \textsc{RI$_0$} increased by more than 40\% from $\mathcal{D}_0$ to $\mathcal{D}_1$ on ResNet-50.
This gap continued to increase until the shift reached a moderate degree.
Moreover, on ViT-B/32, a similar pattern is observed albeit slightly improved.
We hypothesize that the sensitiveness is largely because Gaussian noise is very rare in the training data of CLIP and also in ImageNet.
Evaluation under a more common type of distribution shift, rotation, has provided some evidence to support our hypothesis.
On \textsc{RotatedMNIST}, the pre-trained models are only slightly worse than the randomly initialized models under mild to moderate shifts while being much better under strong shifts.

In \Figref{fig:lp_vs_scratch}, we also compare CLIP models with ImageNet pre-trained models on harder problems: \textsc{NoisyImageNet15} and \textsc{LR-ImageNet15}.
On these two datasets, we observe that ImageNet pre-trained models are generally more robust than CLIP models under both distribution shifts.
Furthermore, the gap between the two model families starts out being small but gradually enlarges as the shift gets stronger.
This suggests that not only the nature of the downstream distribution shift (e.g., noise) but also the difference between the pre-training data and the downstream task itself plays a role in determining the robustness of the pre-trained models against downstream distribution shifts.
Our result complements the finding made in \citep{kumar2022fine} which shows that fine-tuning underperforms linear probing when the pre-trained features are ``good'' (the OOD data are covered by the pre-training data) and the downstream distribution shift is large.
In comparison, we have further showed that the gap can be huge when the OOD data is distinct from the pre-training data, and the resulting models adapted through linear probing can be extremely brittle.

Last but not least, we note that further adapting the pre-trained models to downstream distribution shifts can sometimes significantly improve their robustness as shown in \Figref{fig:lp_vs_scratch}, e.g., IN$_0$ vs.\@ IN$_1$ on \textsc{NoisyMNIST}.
On one hand, this corroborates existing findings that large-scale pre-trained representations are highly versatile.
On the other hand, this also suggests that unleashing the power of pre-trained representations may still require sufficiently diverse downstream task data that covers the potential distribution shifts.

\section{Conclusion}
In this work, we have shown that even when a model is robust to a moderate degree of distribution shift, a slight increase in the degree of the shift can be surprisingly detrimental to model performance.
In addition, training with data under only high degrees of distribution shifts may not be able to solve the problem as it does not always guarantee robustness to lower degrees of distribution shifts.
These findings suggest that robustness to certain degrees of distribution shifts may tell us very little about robustness to lower or higher degrees.
Furthermore, we observe that large-scale pre-trained models like CLIP are sensitive to downstream distribution shifts, especially unseen or rarely seen ones, indicating that naive adaptations such as linear probing on clean data only may not always lead to better downstream OOD performance.

In conclusion, our findings suggest that the robustness of neural networks under certain degrees of distribution shift can be quite brittle.
For this reason, we should be very careful when interpreting evaluation results based on data under a limited range of shift degrees, especially when making predictions about model performance under other shift degrees.
To further mitigate the issue, we encourage future research to adopt a more comprehensive evaluation of OOD generalization, e.g., considering multiple degrees of distribution shifts when it is possible to do so.
This should help us gain deeper insight into OOD generalization and develop more reliable and safe AI applications.



\bibliography{main}
\bibliographystyle{iclr2024_conference}

\newpage
\appendix
\section{Additional information about experiment setup}
\subsection{Datasets}
\begin{figure}[H]
    \centering
    \includegraphics[width=\linewidth]{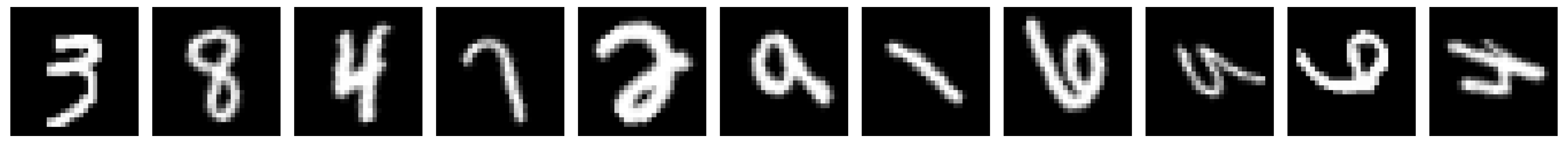}
    \caption{Examples of \textsc{RotatedMNIST}}
    \label{fig:rmnist}
\end{figure}

\begin{figure}[H]
    \centering
    \includegraphics[width=\linewidth]{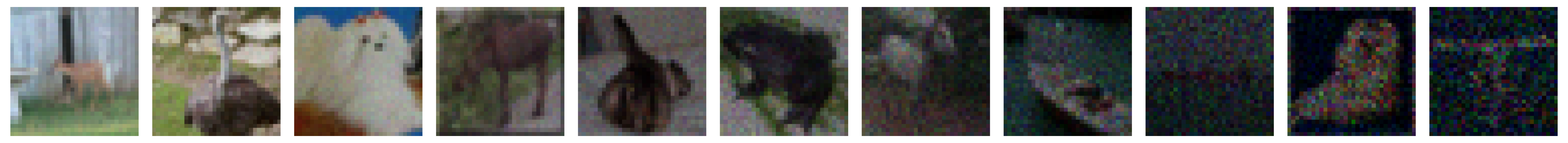}
    \caption{Examples of \textsc{LowLightCIFAR10}}
    \label{fig:lowlight-cifar10}
\end{figure}

\begin{figure}[H]
    \centering
    \includegraphics[width=\linewidth]{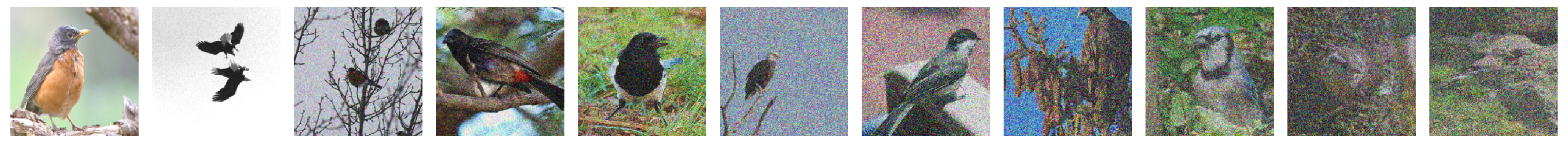}
    \caption{Examples of \textsc{NoisyImageNet15}}
    \label{fig:noisy-imagenet15}
\end{figure}

\begin{figure}[H]
    \centering
    \includegraphics[width=\linewidth]{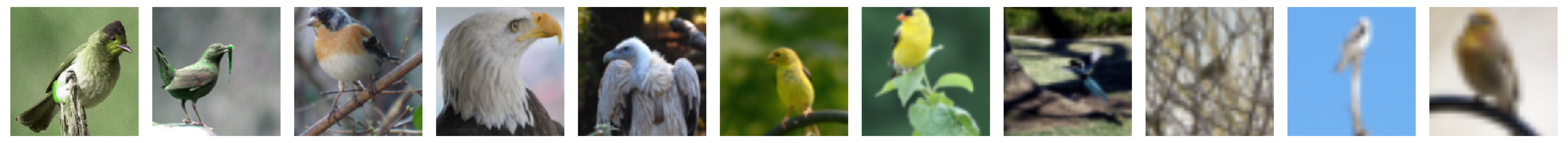}
    \caption{Examples of \textsc{LR-ImageNet15}}
    \label{fig:lr-imagenet15}
\end{figure}

Random examples drawn from each domain of the datasets we used (except \textsc{NoisyMNIST} which is shown in \Figref{fig:noisymnist}) are shown in \Figref{fig:rmnist}-\ref{fig:lr-imagenet15}.
The order of the examples is arranged according to the degree of the distribution shift from low to high.


For \textsc{NoisyMNIST} and \textsc{RotatedMNIST}, 60,000 images were divided into distinct training domains. For instance, in scenarios involving two training domains, each domain would encompass 30,000 images. Within the training domains, 20\% of the data is allocated for in-distribution validation, aiding model calibration and selection.
Every test domain of each altered dataset consists of 10,000 images, constructed using the same set of original images.

For \textsc{NoisyImageNet15} and \textsc{LR-ImageNet15}, we use the images in the training split of ImageNet to construct the training domains and the images in the validation split of ImageNet to construct the test domains.
Similarly, the training domains divide the total 15,000 images in the training split.
The test domains are constructed using the same set of original images, which consist of 750 images in total.

The 15 categories of birds we used in \textsc{NoisyImageNet15} and \textsc{LR-ImageNet15}, which correspond to indices 10 to 24 of the 1,000 categories of ImageNet, are ``brambling, Fringilla montifringilla'', ``goldfinch, Carduelis carduelis'', ``house finch, linnet, Carpodacus mexicanus'', ``junco, snowbird'', ``indigo bunting, indigo finch, indigo bird, Passerina cyanea'', ``robin, American robin, Turdus migratorius'', ``bulbul'', ``jay'', ``magpie'', ``chickadee'', ``water ouzel, dipper'', ``kite'', ``bald eagle, American eagle, Haliaeetus leucocephalus'', ``vulture'', and ``great grey owl, great gray owl, Strix nebulosa''.

In addition to the above datasets studied in the paper, we have also conducted preliminary experiments on another dataset called \textsc{ImpulseNoiseMNIST}. The dataset is constructed by gradually adding impulse noise to MNIST. Below are some of the examples of this dataset. The experiment results on this dataset are given in Appendix~\ref{app:impulse-noise-mnist}.

\begin{figure}
    \centering
    \includegraphics[width=\linewidth]{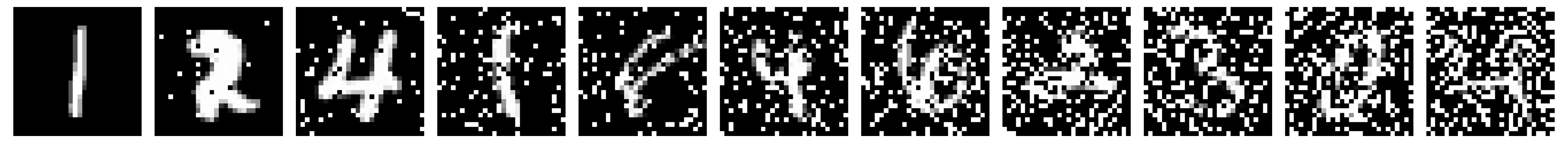}
    \caption{Examples of \textsc{ImpulseNoiseMNIST}}
    \label{fig:enter-label}
\end{figure}

\subsection{Algorithms}
\label{app:algos}
Here is the full list of domain generalization algorithms we used in this study:
\begin{itemize}
    \item Invariant Risk Minimization (\textbf{IRM}, \citealp{arjovsky2019invariant})
    \item Group Distributionally Robust Optimization (\textbf{GroupDRO}, \citealp{sagawa2019distributionally})
    \item Interdomain Mixup (\textbf{Mixup}, \citealp{yan2020improve})
    \item Marginal Transfer Learning (\textbf{MTL}, \citealp{blanchard2011generalizing})
    \item Maximum Mean Discrepancy (\textbf{MMD}, \citealp{li2018domain})
    \item Deep CORAL (\textbf{CORAL}, \citealp{sun2017correlation})
    \item Domain Adversarial Neural Network (\textbf{DANN}, \citealp{ganin2016domain})
    \item Conditional Domain Adversarial Neural Network (\textbf{CDANN}, \citealp{li2018deep})
    \item Style Agnostic Networks (\textbf{SagNet}, \citealp{nam2021reducing})
    \item Adaptive Risk Minimization (\textbf{ARM}, \citealp{zhang2021adaptive})
    \item Variance Risk Extrapolation (\textbf{VREx}, \citealp{krueger2020outofdistribution})
    \item Representation Self-Challenging (\textbf{RSC}, \citealp{huang2020self})
    \item Spectral Decoupling (\textbf{SD}, \citealp{pezeshki2021gradient})
    \item Learning Explanations that are Hard to Vary (\textbf{AND-Mask}, \citealp{parascandolo2021learning})
    \item Smoothed-AND mask (\textbf{SAND-mask}, \citealp{shahtalebi2021sand})
    \item Out-of-Distribution Generalization with Maximal Invariant Predictor (\textbf{IGA}, \citealp{koyama2020out})
    \item Gradient Matching for Domain Generalization (\textbf{Fish}, \citealp{shi2021gradient})
    \item Self-supervised Contrastive Regularization (\textbf{SelfReg}, \citealp{kim2021selfreg})
    \item Learning Representations that Support Robust Transfer of Predictors (\textbf{TRM}, \citealp{xu2021learning})
    \item Invariance Principle Meets Information Bottleneck for Out-of-Distribution Generalization (\textbf{IB-ERM} \& \textbf{IB-IRM}, \citealp{ahuja2021invariance})
    \item Optimal Representations for Covariate Shift (\textbf{CAD} \& \textbf{CondCAD}, \citealp{ruan2021optimal})
    \item Quantifying and Improving Transferability in Domain Generalization (\textbf{Transfer}, \citealp{zhang2021quantifying})
    \item Invariant Causal Mechanisms through Distribution Matching (\textbf{CausIRL} with CORAL or MMD, \citealp{chevalley2022invariant})
    \item Empirical Quantile Risk Minimization (\textbf{EQRM}, \citealp{eastwood2022probable})
\end{itemize}
We use the DomainBed~\citep{gulrajani2021in} implementation for all the above algorithms.

\subsection{Implementation details}
\label{app:imp-details}
The experiments in Section~\ref{sec:scratch} employed three neural network architectures: a simple 4-layer Convolutional Neural Network (CNN); a slightly more complex network, EfficientNet-b0~\citep{tan2019efficientnet}; and a much more complex network, ResNet-50~\citep{he2016deep} model. All the models were implemented without any form of pre-training.
For optimization purposes, we utilized the Adam optimizer~\citep{kingma2014adam} with a static learning rate of 0.001. The total batch size was fixed at 64, and was evenly divided across each training domain.
No weight decay or dropout was applied during the training process.
Training iterations were set to a maximum of 5,000 for the 4-layer CNN and 10,000 for the ResNet-50 to ensure convergence. No form of data augmentation was used throughout the training process, preserving the inherent distribution and characteristics of the datasets. 
To ensure the reliability of our results, we conducted a thorough random search for hyperparameters, repeated 20 times.
Except for learning rate, batch size, weight decay, and dropout as mentioned in the paper, the search of other hyperparameters follows that of DomainBed~\citep{gulrajani2021in}.

For experiments in Section~\ref{sec:pre-trained}, we use training-domain validation to select the best models among different iterations.
We do not use any data augmentation to train randomly initialized ResNet-50 on MNIST-based datasets.
The models were trained for 10,000 maximum iterations under a fixed learning rate 0.001 using Adam.
For randomly initialized ViT-B/32, random affine transformations were applied to MNIST-based datasets, with rotation and shearing disabled for \textsc{RotatedMNIST}.
The models were trained for 200,000 maximum iterations under a fixed learning rate 0.00003 using Adam.
For both randomly initialized ResNet-50 and ViT-B/32, we used batch size 64 evenly divided for each training domain, and no weight decay or dropout was applied.
We do not use any data augmentation for the experiments on \textsc{NoisyImageNet15} and \textsc{LR-ImageNet15}.
All experiments are repeated three times with different random seeds.

For experiments utilizing the ResNet-50 and the ViT-B/32 architecture, all MNIST-based datasets were resized to a resolution of $224{\times}224$ pixels.
For pre-trained models, all datasets are normalized according to the statistics of their respective pre-training datasets, while for randomly initialized models, the datasets are normalized according to their own statistics.

For linear probing, we follow \citep{radford2021learning} and use logistic regression implemented by scikit-learn\footnote{\url{https://scikit-learn.org/stable/modules/generated/sklearn.linear_model.LogisticRegression.html}}.
We also use the same number of maximum iteration, 1,000, and the same hyperparameter search scheme for regularization strength across all experiments.
For all datasets, we use 10\% of held-out data from training domains for training-domain validation and another 10\% held-out data for evaluating ID performance.

\section{Additional experiment results}

\subsection{Effects of increasing the number of training domains}
\label{app:more-tr-domains}
In \Figref{fig:dg}, we have shown the best-performing models (trained on $\mathcal{D}_0$ and $\mathcal{D}_1$) at each degree of \textsc{NoisyMNIST} and \textsc{LowLightCIFAR10}.
In \Figref{fig:dg-mult-tr-domain-spans} and \Figref{fig:dg-worst-case-mult-tr-domain-spans} below, we show more results on these datasets, with models trained on wider ranges of domains.
The results show that more training domains help. As more training domains are added, the gaps between the best-performing models gradually decrease; however, the gaps seem to be closing at a slow rate.
The discrepancy between the best-performing models is still largely present.
In addition, the advantage of DG methods over ERM becomes less pronounced as the number of training domains increases.

\begin{figure}[h]
    \centering
    \includegraphics[width=\linewidth]{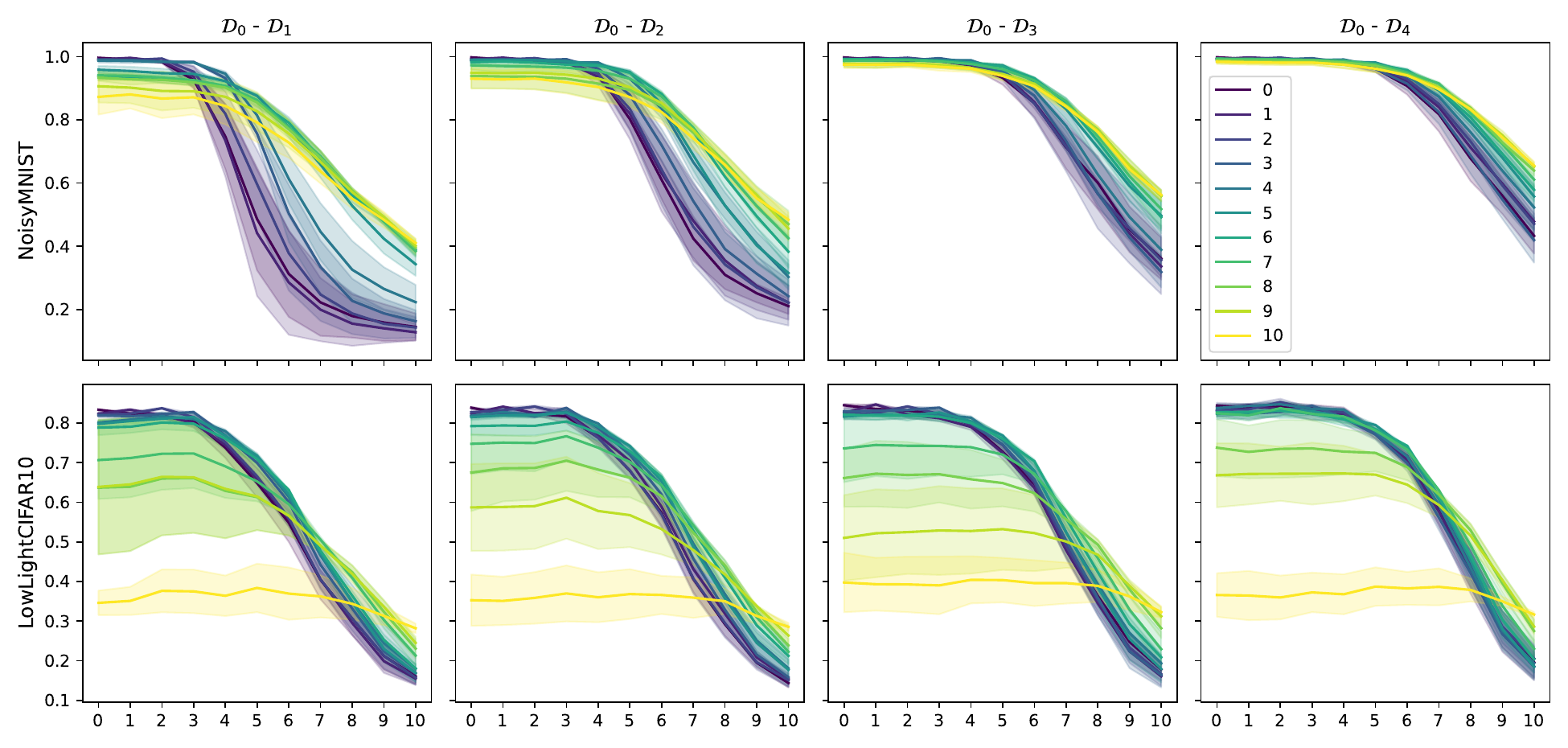}
    \caption{Performance of the best-performing models at each degree of \textsc{NoisyMNIST} and \textsc{LowLightCIFAR10}.
    The label of the curves denotes the domain on which the models perform best compared to the other models.
    The title of each column of the subplots denotes the training domains (e.g., ``$\mathcal{D}_0$ - $\mathcal{D}_2$'' means that the models are trained on $\mathcal{D}_0, \mathcal{D}_1, \mathcal{D}_2$).
    The results are averaged over the top-3 models of all algorithms at each degree.}
    \label{fig:dg-mult-tr-domain-spans}
\end{figure}

\vspace{-1mm}
\begin{figure}[h]
    \centering
    \includegraphics[width=\linewidth]{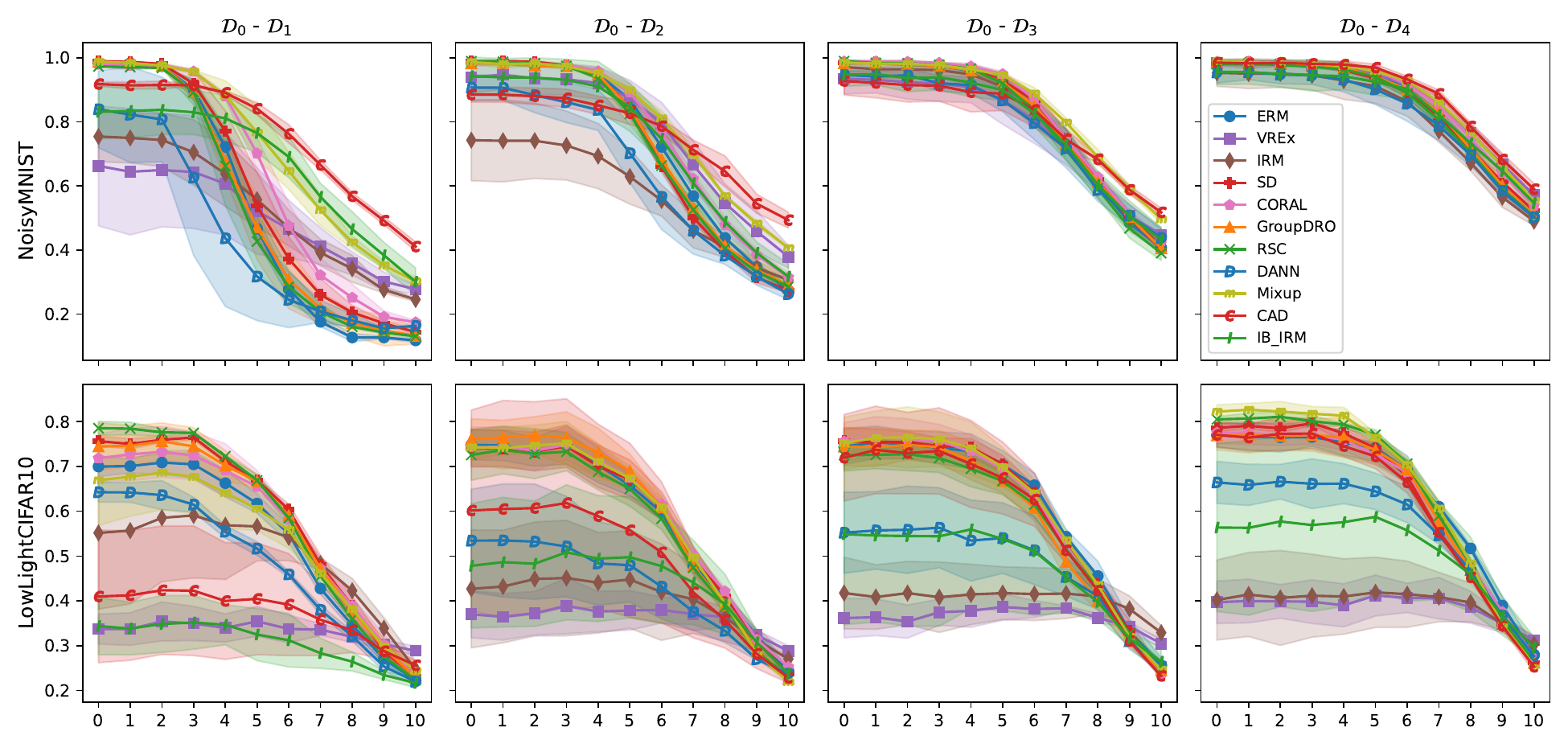}
    \caption{Performance of ERM and representative domain generalization algorithms on \textsc{NoisyMNIST} and \textsc{LowLightCIFAR10}.
    The title of each column of the subplots denotes the training domains (e.g., ``$\mathcal{D}_0$ - $\mathcal{D}_2$'' means that the models are trained on $\mathcal{D}_0, \mathcal{D}_1, \mathcal{D}_2$). The results are averaged over the top-5 models of each algorithm, selected by worst-domain performance.}
    \label{fig:dg-worst-case-mult-tr-domain-spans}
\end{figure}

\subsection{Experiment results on ImpulseNoiseMNIST}
\label{app:impulse-noise-mnist}
\begin{figure}[h]
    \centering
    \includegraphics[width=0.324\linewidth]{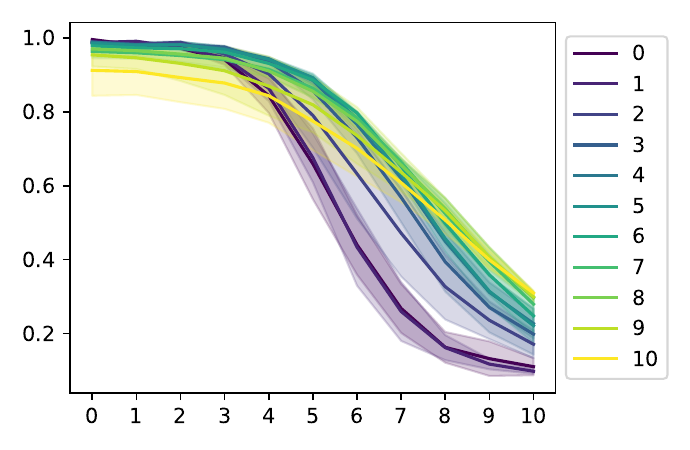}
    \includegraphics[width=0.373\linewidth]{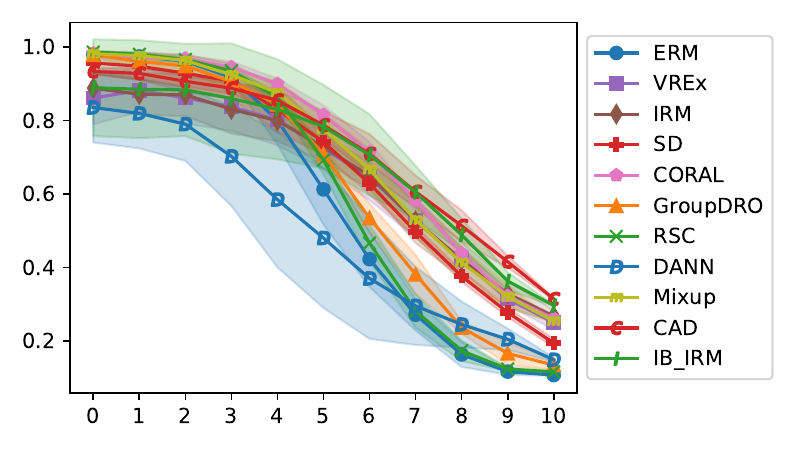}
    \includegraphics[width=0.282\linewidth]{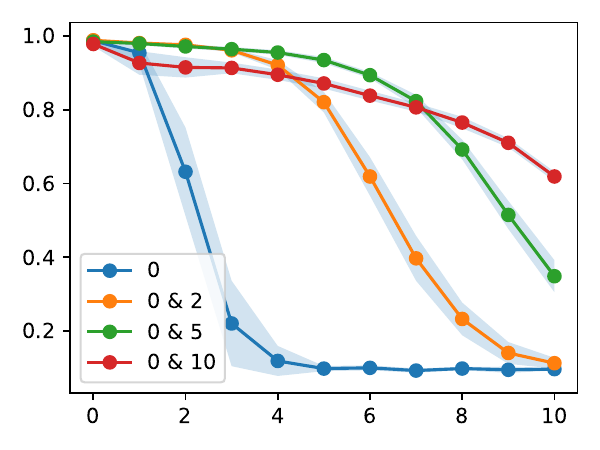}
    \caption{Results on \textsc{ImpulseNoiseMNIST} (Simple CNN).
    \textbf{(Left)} Performance of the best-performing models at each shift degree.
    The label of the curves denotes the domain on which the models perform best;
    \textbf{(Middle)} Performance of ERM and representative domain generalization algorithms on the two datasets;
    \textbf{(Right)} Performance of ERM models trained on domains under different shift degrees. The label of the curves denotes the indices of the training domains, e.g., ``0 \& 2'' means that the models are trained on $\mathcal{D}_0$ and $\mathcal{D}_2$ of the corresponding dataset.
    The implementation details of these experiments follow those of \textsc{NoisyMNIST}.}
    \label{fig:impulse-noise-mnist-results}
\end{figure}

In \Figref{fig:impulse-noise-mnist-results}, we show the experiment results on the \textsc{ImpulseNoiseMNIST} dataset.
The results on the generalization from milder to stronger shifts are largely consistent with the results in \Figref{fig:dg}.
As for the results on the generalization from stronger to milder shifts, the pattern looks like that of \textsc{LowLightCIFAR10}.

\subsection{Strong-to-mild generalization performance of DG algorithms}
\begin{figure}[h]
    \centering
    \includegraphics[width=0.7\linewidth]{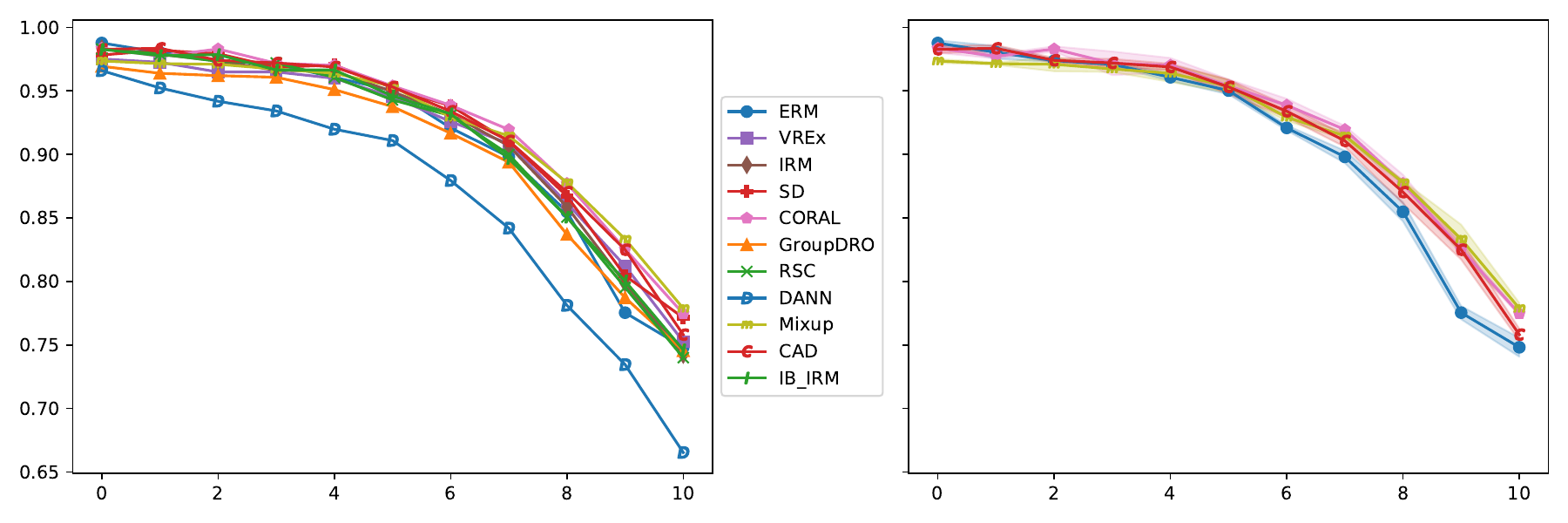}
    \caption{Average accuracy of the top-3 models of ERM and various DG algorithms trained on $\mathcal{D}_0$ and $\mathcal{D}_{10}$ of \textsc{NoisyMNIST}. The models are selected via training-domain validation. Error bars are omitted in the left sub-figure for clarity. The best-performing DG algorithms are compared with ERM in the right sub-figure. The results are averaged over 3 runs.}
    \label{fig:noisy-mnist-dg}
\end{figure}

\begin{figure}[h]
    \centering
    \includegraphics[width=0.7\linewidth]{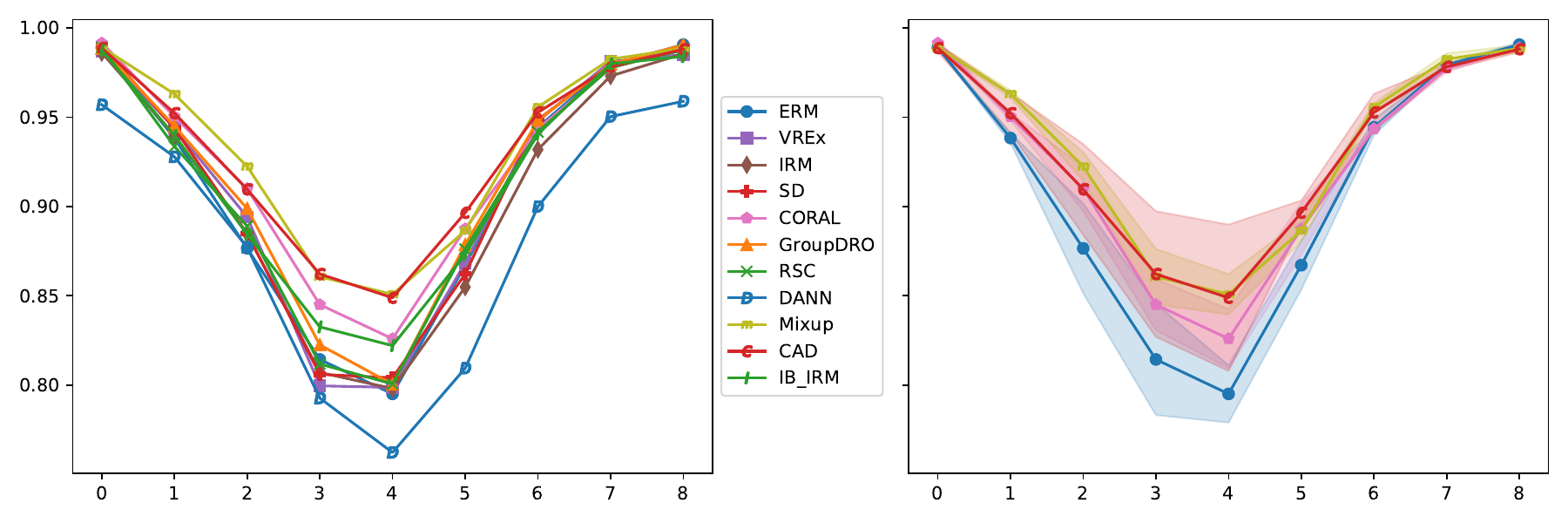}
    \caption{Average accuracy of the top-3 models of ERM and various DG algorithms trained on $\mathcal{D}_0$ and $\mathcal{D}_8$ of \textsc{RotatedMNIST}. The models are selected via training-domain validation. Error bars are omitted in the left sub-figure for clarity. The best-performing DG algorithms are compared with ERM in the right sub-figure. The results are averaged over 3 runs.}
    \label{fig:rmnist-dg}
\end{figure}

\begin{figure}[h]
    \centering
    \includegraphics[width=0.7\linewidth]{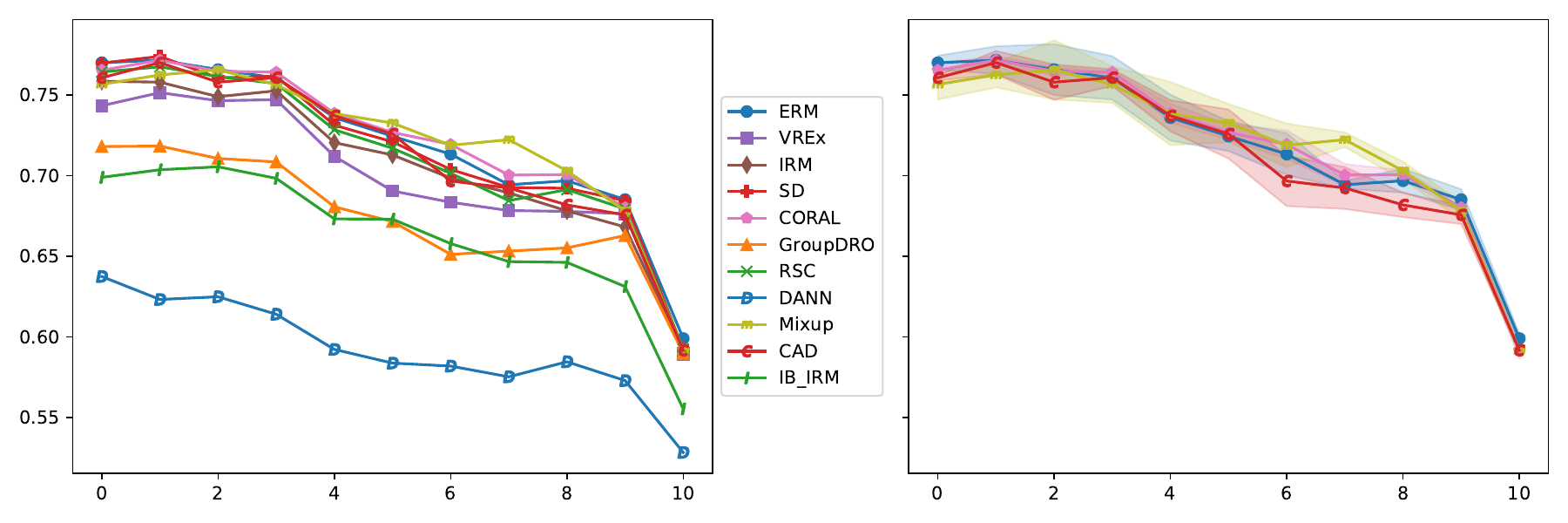}
    \caption{Average accuracy of the top-3 models of ERM and various DG algorithms trained on $\mathcal{D}_0$ and $\mathcal{D}_{10}$ of \textsc{LowLightCIFAR10}. The models are selected via training-domain validation. Error bars are omitted in the left sub-figure for clarity. The best-performing DG algorithms are compared with ERM in the right sub-figure. The results are averaged over 3 runs.}
    \label{fig:lowlight-cifar10-dg}
\end{figure}

In this section, we show the performance of DG algorithms for the hardest generalization case in Figure.~\ref{fig:tr-envs}. From Figure~\ref{fig:noisy-mnist-dg} to Figure~\ref{fig:lowlight-cifar10-dg}, we can see that most DG algorithms are helpful to the generalization from stronger shifts to milder shifts in the case of \textsc{RotatedMNIST}, although only to a limited extent.
On the other two datasets, only some of the DG algorithms are able to improve the generalization performance by a very small margin.


\subsection{Full numerical results of the experiments conducted in Section~\ref{sec:scratch}}
From Table~\ref{tab:noisymnist-cnn-0-3} to Table~\ref{tab:lowlightcifar10-rn50-8-10}, we show the full numerical results of the experiments we conducted in Section~\ref{sec:scratch}.
The first column of each table shows the domain on which the accuracy is used to select the top models.

\begin{table}[t]
    \centering
    \adjustbox{max width=\textwidth}{
    \begin{tabular}{llcccccccccccc}
        \toprule
        & Algorithm & $\mathcal{D}_{0}$ & $\mathcal{D}_{1}$ & $\mathcal{D}_{2}$ & $\mathcal{D}_{3}$ & $\mathcal{D}_{4}$ & $\mathcal{D}_{5}$ & $\mathcal{D}_{6}$ & $\mathcal{D}_{7}$ & $\mathcal{D}_{8}$ & $\mathcal{D}_{9}$ & $\mathcal{D}_{10}$\\
        \midrule
        \multirow{28}{*}{$\mathcal{D}_{0}$}
        & ERM        & 99.37$\pm$0.07 & 98.65$\pm$0.14 & 97.25$\pm$0.43 & 82.72$\pm$3.85 & 48.36$\pm$9.88 & 21.98$\pm$6.53 & 12.88$\pm$2.55 & 9.95$\pm$1.38 & 10.82$\pm$0.74 & 9.84$\pm$0.68 & 10.14$\pm$1.14\\
        & VREx       & 99.20$\pm$0.15 & 98.88$\pm$0.30 & 97.56$\pm$0.21 & 88.10$\pm$5.71 & 59.81$\pm$12.09 & 32.04$\pm$10.96 & 18.57$\pm$6.66 & 13.43$\pm$2.78 & 10.55$\pm$1.98 & 10.68$\pm$1.31 & 10.31$\pm$0.77\\
        & IRM        & 99.09$\pm$0.17 & 98.55$\pm$0.30 & 97.66$\pm$0.41 & 83.63$\pm$5.51 & 53.36$\pm$9.50 & 25.05$\pm$7.84 & 12.74$\pm$2.59 & 10.41$\pm$0.91 & 10.61$\pm$0.48 & 9.50$\pm$0.58 & 9.58$\pm$0.84\\
        & SD         & 99.36$\pm$0.17 & 98.82$\pm$0.14 & 98.18$\pm$0.37 & 84.91$\pm$4.27 & 49.75$\pm$13.37 & 22.59$\pm$13.01 & 14.06$\pm$6.46 & 10.91$\pm$3.01 & 10.58$\pm$1.40 & 10.03$\pm$1.35 & 9.97$\pm$0.83\\
        & GroupDRO   & 99.20$\pm$0.14 & 98.79$\pm$0.28 & 98.03$\pm$0.54 & 85.13$\pm$2.85 & 50.68$\pm$8.50 & 24.15$\pm$10.77 & 15.60$\pm$6.99 & 12.04$\pm$2.89 & 10.00$\pm$1.17 & 9.83$\pm$0.67 & 9.72$\pm$0.87\\
        & RSC        & 99.23$\pm$0.12 & 98.65$\pm$0.42 & 97.75$\pm$0.36 & 88.07$\pm$4.69 & 55.94$\pm$9.22 & 25.31$\pm$5.46 & 15.27$\pm$2.01 & 11.10$\pm$1.17 & 10.03$\pm$1.22 & 10.22$\pm$0.82 & 10.00$\pm$0.48\\
        & Mixup      & 99.47$\pm$0.12 & 98.92$\pm$0.39 & 97.97$\pm$0.41 & 93.77$\pm$2.34 & 81.08$\pm$6.99 & 64.72$\pm$8.22 & 49.45$\pm$8.73 & 38.68$\pm$7.03 & 30.17$\pm$5.88 & 25.55$\pm$5.45 & 21.43$\pm$4.18\\
\bottomrule
    \end{tabular}}
    \caption{Average accuracy of top-3 models of each algorithm at each shift degree of \textsc{NoisyMNIST} (Simple CNN). This table shows the results on shift degrees from $\mathcal{D}_{0}$ to $\mathcal{D}_{3}$.}
    \label{tab:noisymnist-cnn-0-3}
\end{table}
\begin{table}[t]
    \centering
    \adjustbox{max width=\textwidth}{
    \begin{tabular}{llcccccccccccc}
        \toprule
        & Algorithm & $\mathcal{D}_{0}$ & $\mathcal{D}_{1}$ & $\mathcal{D}_{2}$ & $\mathcal{D}_{3}$ & $\mathcal{D}_{4}$ & $\mathcal{D}_{5}$ & $\mathcal{D}_{6}$ & $\mathcal{D}_{7}$ & $\mathcal{D}_{8}$ & $\mathcal{D}_{9}$ & $\mathcal{D}_{10}$\\
        \midrule
        \multirow{28}{*}{$\mathcal{D}_{4}$}
        & ERM        & 98.27$\pm$0.59 & 98.35$\pm$0.73 & 98.02$\pm$0.43 & 93.38$\pm$1.85 & 75.27$\pm$3.76 & 44.87$\pm$5.37 & 24.09$\pm$5.11 & 14.84$\pm$3.01 & 11.97$\pm$1.31 & 11.59$\pm$1.05 & 10.66$\pm$0.80\\
        & VREx       & 97.42$\pm$1.27 & 97.45$\pm$0.89 & 97.17$\pm$1.57 & 95.60$\pm$1.66 & 86.71$\pm$4.83 & 67.91$\pm$10.97 & 48.03$\pm$11.35 & 33.99$\pm$11.69 & 23.92$\pm$8.46 & 19.20$\pm$7.85 & 16.08$\pm$5.44\\
        & IRM        & 91.01$\pm$6.89 & 90.93$\pm$6.54 & 90.64$\pm$6.14 & 86.89$\pm$5.72 & 76.81$\pm$2.95 & 58.44$\pm$6.45 & 39.81$\pm$9.09 & 27.34$\pm$8.96 & 19.75$\pm$7.77 & 15.74$\pm$5.60 & 13.66$\pm$3.91\\
        & SD         & 97.67$\pm$1.51 & 97.86$\pm$0.96 & 97.56$\pm$0.89 & 93.96$\pm$2.14 & 80.64$\pm$1.61 & 55.83$\pm$2.90 & 32.72$\pm$3.98 & 21.04$\pm$3.04 & 15.52$\pm$3.16 & 13.47$\pm$2.64 & 12.25$\pm$1.91\\
        & GroupDRO   & 98.11$\pm$0.82 & 98.13$\pm$0.34 & 97.63$\pm$0.45 & 91.93$\pm$2.47 & 72.33$\pm$2.47 & 47.58$\pm$6.60 & 28.07$\pm$8.41 & 17.80$\pm$6.56 & 13.88$\pm$4.85 & 13.07$\pm$4.47 & 11.93$\pm$2.68\\
        & RSC        & 97.52$\pm$0.35 & 97.22$\pm$0.73 & 96.54$\pm$0.99 & 93.03$\pm$2.02 & 79.26$\pm$7.09 & 54.78$\pm$9.93 & 36.24$\pm$7.15 & 23.98$\pm$3.88 & 17.69$\pm$2.05 & 14.81$\pm$1.26 & 12.72$\pm$0.63\\
        & Mixup      & 98.38$\pm$0.64 & 98.27$\pm$0.57 & 97.91$\pm$0.54 & 96.78$\pm$0.62 & 92.41$\pm$1.16 & 81.12$\pm$4.09 & 64.54$\pm$6.19 & 49.15$\pm$6.91 & 37.09$\pm$6.29 & 30.17$\pm$5.13 & 25.86$\pm$5.06\\
\bottomrule
    \end{tabular}}
    \caption{Average accuracy of top-3 models of each algorithm at each shift degree of \textsc{NoisyMNIST} (Simple CNN). This table shows the results on shift degrees from $\mathcal{D}_{4}$ to $\mathcal{D}_{7}$.}
    \label{tab:noisymnist-cnn-4-7}
\end{table}
\begin{table}[t]
    \centering
    \adjustbox{max width=\textwidth}{
    \begin{tabular}{llcccccccccccc}
        \toprule
        & Algorithm & $\mathcal{D}_{0}$ & $\mathcal{D}_{1}$ & $\mathcal{D}_{2}$ & $\mathcal{D}_{3}$ & $\mathcal{D}_{4}$ & $\mathcal{D}_{5}$ & $\mathcal{D}_{6}$ & $\mathcal{D}_{7}$ & $\mathcal{D}_{8}$ & $\mathcal{D}_{9}$ & $\mathcal{D}_{10}$\\
        \midrule
        \multirow{28}{*}{$\mathcal{D}_{8}$}
        & ERM        & 98.24$\pm$0.71 & 97.96$\pm$0.76 & 96.81$\pm$1.07 & 88.38$\pm$3.51 & 68.27$\pm$5.31 & 41.90$\pm$7.41 & 25.72$\pm$5.45 & 17.42$\pm$3.65 & 14.64$\pm$1.20 & 11.98$\pm$1.25 & 10.82$\pm$1.28\\
        & VREx       & 78.21$\pm$20.60 & 77.33$\pm$21.97 & 77.28$\pm$20.42 & 76.51$\pm$20.24 & 71.75$\pm$18.50 & 60.86$\pm$15.31 & 50.79$\pm$9.15 & 43.21$\pm$5.80 & 34.94$\pm$3.40 & 29.51$\pm$2.81 & 25.68$\pm$2.77\\
        & IRM        & 78.35$\pm$8.93 & 77.94$\pm$9.17 & 76.86$\pm$7.17 & 70.09$\pm$7.41 & 63.19$\pm$8.99 & 54.17$\pm$9.59 & 43.10$\pm$8.43 & 36.53$\pm$5.64 & 31.37$\pm$5.55 & 25.30$\pm$2.90 & 21.97$\pm$2.54\\
        & SD         & 98.30$\pm$0.68 & 98.30$\pm$0.48 & 97.70$\pm$0.45 & 91.93$\pm$3.32 & 77.06$\pm$5.25 & 53.63$\pm$8.09 & 34.09$\pm$6.80 & 23.35$\pm$5.27 & 18.32$\pm$3.35 & 15.69$\pm$2.19 & 12.99$\pm$1.88\\
        & GroupDRO   & 82.74$\pm$31.90 & 83.02$\pm$31.17 & 82.20$\pm$31.00 & 74.25$\pm$29.22 & 55.16$\pm$22.73 & 36.65$\pm$16.02 & 24.12$\pm$10.87 & 17.94$\pm$6.06 & 15.24$\pm$3.97 & 13.82$\pm$4.07 & 12.14$\pm$2.44\\
        & RSC        & 97.56$\pm$0.33 & 97.14$\pm$0.70 & 96.57$\pm$1.01 & 91.92$\pm$4.04 & 74.78$\pm$14.11 & 51.21$\pm$14.26 & 33.71$\pm$9.29 & 22.92$\pm$4.36 & 17.94$\pm$1.85 & 14.94$\pm$1.17 & 12.72$\pm$0.63\\
        & Mixup      & 98.68$\pm$0.65 & 98.77$\pm$0.61 & 97.66$\pm$0.40 & 95.64$\pm$1.44 & 87.66$\pm$5.48 & 76.18$\pm$8.01 & 63.46$\pm$6.14 & 50.96$\pm$5.00 & 41.13$\pm$2.20 & 33.84$\pm$1.98 & 29.28$\pm$1.53\\
\bottomrule
    \end{tabular}}
    \caption{Average accuracy of top-3 models of each algorithm at each shift degree of \textsc{NoisyMNIST} (Simple CNN). This table shows the results on shift degrees from $\mathcal{D}_{8}$ to $\mathcal{D}_{10}$.}
    \label{tab:noisymnist-cnn-8-10}
\end{table}
\begin{table}[t]
    \centering
    \adjustbox{max width=\textwidth}{
    \begin{tabular}{llcccccccccccc}
        \toprule
        & Algorithm & $\mathcal{D}_{0}$ & $\mathcal{D}_{1}$ & $\mathcal{D}_{2}$ & $\mathcal{D}_{3}$ & $\mathcal{D}_{4}$ & $\mathcal{D}_{5}$ & $\mathcal{D}_{6}$ & $\mathcal{D}_{7}$ & $\mathcal{D}_{8}$ & $\mathcal{D}_{9}$ & $\mathcal{D}_{10}$\\
        \midrule
        \multirow{23}{*}{$\mathcal{D}_{0}$}
        & ERM        & 98.61$\pm$0.05 & 98.27$\pm$0.21 & 98.46$\pm$0.18 & 87.10$\pm$6.14 & 54.98$\pm$13.89 & 27.71$\pm$11.31 & 17.14$\pm$7.75 & 12.57$\pm$4.23 & 10.53$\pm$2.17 & 9.74$\pm$1.24 & 10.14$\pm$0.66\\
        & VREx       & 98.58$\pm$0.13 & 98.26$\pm$0.12 & 98.53$\pm$0.40 & 86.70$\pm$6.22 & 50.95$\pm$13.92 & 23.98$\pm$7.73 & 13.71$\pm$2.45 & 10.45$\pm$0.92 & 9.36$\pm$0.45 & 9.14$\pm$0.15 & 9.82$\pm$0.15\\
        & IRM        & 98.37$\pm$0.24 & 98.38$\pm$0.27 & 98.30$\pm$0.47 & 84.63$\pm$6.49 & 53.42$\pm$9.34 & 29.23$\pm$9.34 & 18.87$\pm$5.47 & 13.32$\pm$2.39 & 10.81$\pm$1.19 & 9.92$\pm$0.69 & 10.20$\pm$0.41\\
        & SD         & 98.60$\pm$0.04 & 98.35$\pm$0.17 & 98.22$\pm$0.81 & 84.80$\pm$5.94 & 49.07$\pm$8.79 & 22.45$\pm$3.44 & 14.10$\pm$2.63 & 11.96$\pm$2.69 & 10.50$\pm$2.18 & 10.39$\pm$2.39 & 10.76$\pm$1.87\\
        & CORAL      & 98.58$\pm$0.07 & 98.31$\pm$0.13 & 96.46$\pm$0.71 & 67.62$\pm$5.11 & 31.23$\pm$4.64 & 14.92$\pm$3.73 & 12.14$\pm$2.68 & 11.09$\pm$2.31 & 10.31$\pm$1.68 & 10.15$\pm$1.55 & 10.69$\pm$1.35\\
        & GroupDRO   & 98.61$\pm$0.11 & 98.18$\pm$0.22 & 96.81$\pm$1.76 & 74.17$\pm$8.37 & 37.24$\pm$6.84 & 17.73$\pm$4.33 & 13.02$\pm$2.11 & 10.57$\pm$0.89 & 9.74$\pm$0.68 & 9.49$\pm$0.41 & 9.83$\pm$0.17\\
        & RSC        & 98.67$\pm$0.10 & 98.46$\pm$0.22 & 97.96$\pm$0.32 & 81.93$\pm$5.49 & 45.53$\pm$9.40 & 23.62$\pm$7.34 & 16.67$\pm$4.70 & 12.77$\pm$2.77 & 10.51$\pm$1.37 & 10.20$\pm$1.18 & 10.34$\pm$0.55\\
\bottomrule
    \end{tabular}}
    \caption{Average accuracy of top-3 models of each algorithm at each shift degree of \textsc{NoisyMNIST} (EfficientNet-b0). This table shows the results on shift degrees from $\mathcal{D}_{0}$ to $\mathcal{D}_{3}$.}
    \label{tab:noisymnist-enb0-0-3}
\end{table}
\begin{table}[t]
    \centering
    \adjustbox{max width=\textwidth}{
    \begin{tabular}{llcccccccccccc}
        \toprule
        & Algorithm & $\mathcal{D}_{0}$ & $\mathcal{D}_{1}$ & $\mathcal{D}_{2}$ & $\mathcal{D}_{3}$ & $\mathcal{D}_{4}$ & $\mathcal{D}_{5}$ & $\mathcal{D}_{6}$ & $\mathcal{D}_{7}$ & $\mathcal{D}_{8}$ & $\mathcal{D}_{9}$ & $\mathcal{D}_{10}$\\
        \midrule
        \multirow{23}{*}{$\mathcal{D}_{4}$}
        & ERM        & 98.19$\pm$0.26 & 97.99$\pm$0.33 & 98.46$\pm$0.10 & 92.55$\pm$0.44 & 69.59$\pm$1.50 & 39.58$\pm$3.69 & 23.73$\pm$3.67 & 15.52$\pm$2.61 & 12.39$\pm$1.87 & 10.83$\pm$1.49 & 10.86$\pm$0.83\\
        & VREx       & 98.12$\pm$0.33 & 98.07$\pm$0.29 & 98.44$\pm$0.39 & 89.83$\pm$3.34 & 61.48$\pm$5.58 & 31.04$\pm$5.74 & 16.91$\pm$2.93 & 11.26$\pm$0.99 & 9.63$\pm$0.46 & 9.19$\pm$0.13 & 9.82$\pm$0.12\\
        & IRM        & 88.82$\pm$17.73 & 89.41$\pm$16.73 & 91.83$\pm$12.46 & 84.05$\pm$13.99 & 63.11$\pm$7.16 & 38.50$\pm$5.99 & 24.68$\pm$7.73 & 17.83$\pm$6.66 & 14.85$\pm$6.40 & 13.34$\pm$4.95 & 13.00$\pm$4.15\\
        & SD         & 97.77$\pm$0.37 & 97.80$\pm$0.34 & 98.34$\pm$0.24 & 91.32$\pm$1.48 & 67.17$\pm$3.47 & 35.72$\pm$6.69 & 20.78$\pm$5.15 & 14.56$\pm$3.57 & 12.21$\pm$2.67 & 11.15$\pm$1.99 & 10.81$\pm$0.97\\
        & CORAL      & 98.03$\pm$0.39 & 97.66$\pm$0.35 & 97.37$\pm$0.92 & 82.12$\pm$4.07 & 51.39$\pm$2.27 & 27.09$\pm$5.56 & 18.47$\pm$5.57 & 13.87$\pm$4.00 & 12.00$\pm$3.27 & 11.38$\pm$2.66 & 11.25$\pm$1.64\\
        & GroupDRO   & 97.69$\pm$0.27 & 97.77$\pm$0.26 & 97.39$\pm$0.41 & 85.23$\pm$0.73 & 60.15$\pm$2.30 & 37.27$\pm$3.76 & 24.19$\pm$4.60 & 17.44$\pm$3.56 & 14.24$\pm$2.51 & 13.04$\pm$2.20 & 12.85$\pm$1.67\\
        & RSC        & 97.01$\pm$1.41 & 97.01$\pm$1.55 & 97.32$\pm$0.86 & 89.61$\pm$2.01 & 68.38$\pm$3.68 & 38.91$\pm$4.31 & 23.97$\pm$3.42 & 16.40$\pm$2.32 & 12.59$\pm$1.47 & 10.92$\pm$0.97 & 10.93$\pm$0.61\\
\bottomrule
    \end{tabular}}
    \caption{Average accuracy of top-3 models of each algorithm at each shift degree of \textsc{NoisyMNIST} (EfficientNet-b0). This table shows the results on shift degrees from $\mathcal{D}_{4}$ to $\mathcal{D}_{7}$.}
    \label{tab:noisymnist-enb0-4-7}
\end{table}
\begin{table}[t]
    \centering
    \adjustbox{max width=\textwidth}{
    \begin{tabular}{llcccccccccccc}
        \toprule
        & Algorithm & $\mathcal{D}_{0}$ & $\mathcal{D}_{1}$ & $\mathcal{D}_{2}$ & $\mathcal{D}_{3}$ & $\mathcal{D}_{4}$ & $\mathcal{D}_{5}$ & $\mathcal{D}_{6}$ & $\mathcal{D}_{7}$ & $\mathcal{D}_{8}$ & $\mathcal{D}_{9}$ & $\mathcal{D}_{10}$\\
        \midrule
        \multirow{23}{*}{$\mathcal{D}_{8}$}
        & ERM        & 94.22$\pm$0.67 & 94.05$\pm$0.85 & 94.24$\pm$1.45 & 74.94$\pm$7.71 & 48.48$\pm$9.41 & 32.12$\pm$5.72 & 24.30$\pm$3.23 & 18.63$\pm$1.41 & 16.41$\pm$0.52 & 14.46$\pm$0.95 & 13.99$\pm$0.93\\
        & VREx       & 59.43$\pm$35.14 & 59.49$\pm$34.78 & 59.27$\pm$33.53 & 44.48$\pm$22.61 & 28.01$\pm$10.30 & 21.95$\pm$6.36 & 18.59$\pm$3.33 & 16.11$\pm$1.66 & 14.89$\pm$0.44 & 13.47$\pm$1.16 & 12.31$\pm$1.24\\
        & IRM        & 37.77$\pm$8.21 & 38.46$\pm$9.51 & 48.09$\pm$11.27 & 44.61$\pm$7.92 & 39.19$\pm$7.48 & 33.96$\pm$5.99 & 29.62$\pm$5.63 & 25.77$\pm$3.83 & 23.18$\pm$2.74 & 20.64$\pm$2.12 & 19.58$\pm$1.39\\
        & SD         & 97.22$\pm$1.64 & 97.34$\pm$1.47 & 97.28$\pm$1.36 & 86.48$\pm$4.14 & 57.34$\pm$8.95 & 35.91$\pm$6.32 & 26.41$\pm$4.06 & 19.59$\pm$2.32 & 17.64$\pm$0.98 & 15.17$\pm$0.39 & 14.06$\pm$0.27\\
        & CORAL      & 97.07$\pm$1.46 & 96.88$\pm$1.50 & 93.94$\pm$4.65 & 68.02$\pm$9.59 & 40.86$\pm$7.34 & 26.65$\pm$4.98 & 21.77$\pm$3.34 & 17.17$\pm$1.74 & 16.00$\pm$1.06 & 14.23$\pm$0.93 & 13.96$\pm$0.69\\
        & GroupDRO   & 96.76$\pm$1.80 & 96.87$\pm$1.79 & 95.92$\pm$2.65 & 76.20$\pm$10.77 & 51.72$\pm$11.65 & 34.87$\pm$6.79 & 25.40$\pm$2.80 & 19.04$\pm$1.71 & 16.25$\pm$0.47 & 14.81$\pm$0.85 & 14.04$\pm$1.09\\
        & RSC        & 96.25$\pm$2.25 & 95.98$\pm$2.05 & 95.80$\pm$3.19 & 81.32$\pm$7.43 & 54.88$\pm$6.68 & 35.72$\pm$5.13 & 26.49$\pm$3.90 & 19.61$\pm$1.66 & 16.48$\pm$0.97 & 13.85$\pm$0.70 & 12.51$\pm$0.57\\
\bottomrule
    \end{tabular}}
    \caption{Average accuracy of top-3 models of each algorithm at each shift degree of \textsc{NoisyMNIST} (EfficientNet-b0). This table shows the results on shift degrees from $\mathcal{D}_{8}$ to $\mathcal{D}_{10}$.}
    \label{tab:noisymnist-enb0-8-10}
\end{table}
\begin{table}[t]
    \centering
    \adjustbox{max width=\textwidth}{
    \begin{tabular}{llcccccccccccc}
        \toprule
        & Algorithm & $\mathcal{D}_{0}$ & $\mathcal{D}_{1}$ & $\mathcal{D}_{2}$ & $\mathcal{D}_{3}$ & $\mathcal{D}_{4}$ & $\mathcal{D}_{5}$ & $\mathcal{D}_{6}$ & $\mathcal{D}_{7}$ & $\mathcal{D}_{8}$ & $\mathcal{D}_{9}$ & $\mathcal{D}_{10}$\\
        \midrule
        \multirow{28}{*}{$\mathcal{D}_{0}$}
        & ERM        & 99.11$\pm$0.02 & 98.92$\pm$0.09 & 99.13$\pm$0.36 & 87.94$\pm$10.08 & 56.18$\pm$20.90 & 30.53$\pm$15.70 & 18.25$\pm$7.61 & 13.17$\pm$2.92 & 10.90$\pm$1.20 & 10.21$\pm$1.08 & 10.13$\pm$0.66\\
        & VREx       & 99.08$\pm$0.06 & 98.96$\pm$0.15 & 93.25$\pm$8.13 & 56.57$\pm$13.90 & 28.56$\pm$7.81 & 16.96$\pm$2.44 & 12.68$\pm$0.70 & 10.48$\pm$0.45 & 9.94$\pm$0.44 & 9.23$\pm$0.93 & 9.58$\pm$0.38\\
        & IRM        & 98.58$\pm$0.44 & 98.32$\pm$0.66 & 98.97$\pm$0.44 & 90.74$\pm$9.27 & 69.49$\pm$23.37 & 49.34$\pm$26.62 & 30.90$\pm$17.22 & 18.51$\pm$8.49 & 14.77$\pm$5.87 & 11.89$\pm$3.71 & 11.35$\pm$2.54\\
        & SD         & 99.29$\pm$0.05 & 99.14$\pm$0.08 & 99.23$\pm$0.22 & 86.87$\pm$8.17 & 48.54$\pm$11.92 & 21.98$\pm$6.40 & 13.88$\pm$2.21 & 11.27$\pm$1.42 & 10.23$\pm$1.60 & 9.98$\pm$1.32 & 10.27$\pm$0.75\\
        & GroupDRO   & 99.21$\pm$0.04 & 98.99$\pm$0.25 & 96.50$\pm$3.41 & 57.57$\pm$28.26 & 31.37$\pm$19.58 & 16.76$\pm$6.64 & 11.94$\pm$1.76 & 9.74$\pm$0.27 & 9.96$\pm$0.32 & 9.83$\pm$0.63 & 9.95$\pm$0.66\\
        & RSC        & 99.17$\pm$0.04 & 99.03$\pm$0.14 & 97.08$\pm$2.31 & 65.49$\pm$20.73 & 35.51$\pm$15.79 & 18.28$\pm$7.16 & 12.15$\pm$3.18 & 9.73$\pm$1.88 & 9.04$\pm$0.80 & 9.05$\pm$0.93 & 9.63$\pm$0.78\\
        & Mixup      & 99.30$\pm$0.04 & 99.03$\pm$0.11 & 98.50$\pm$0.56 & 77.85$\pm$12.10 & 32.30$\pm$7.37 & 14.96$\pm$1.93 & 11.05$\pm$0.91 & 9.57$\pm$0.22 & 8.89$\pm$0.40 & 8.87$\pm$0.20 & 9.44$\pm$0.55\\
\bottomrule
    \end{tabular}}
    \caption{Average accuracy of top-3 models of each algorithm at each shift degree of \textsc{NoisyMNIST} (ResNet-50). This table shows the results on shift degrees from $\mathcal{D}_{0}$ to $\mathcal{D}_{3}$.}
    \label{tab:noisymnist-rn50-0-3}
\end{table}
\begin{table}[t]
    \centering
    \adjustbox{max width=\textwidth}{
    \begin{tabular}{llcccccccccccc}
        \toprule
        & Algorithm & $\mathcal{D}_{0}$ & $\mathcal{D}_{1}$ & $\mathcal{D}_{2}$ & $\mathcal{D}_{3}$ & $\mathcal{D}_{4}$ & $\mathcal{D}_{5}$ & $\mathcal{D}_{6}$ & $\mathcal{D}_{7}$ & $\mathcal{D}_{8}$ & $\mathcal{D}_{9}$ & $\mathcal{D}_{10}$\\
        \midrule
        \multirow{28}{*}{$\mathcal{D}_{4}$}
        & ERM        & 98.10$\pm$0.62 & 97.80$\pm$0.67 & 98.96$\pm$0.22 & 98.42$\pm$0.47 & 96.84$\pm$0.72 & 84.65$\pm$4.70 & 55.31$\pm$12.41 & 30.96$\pm$9.81 & 19.61$\pm$5.18 & 15.61$\pm$4.10 & 13.17$\pm$2.51\\
        & VREx       & 90.60$\pm$15.14 & 90.21$\pm$15.41 & 91.88$\pm$13.39 & 87.25$\pm$10.51 & 62.43$\pm$2.96 & 33.81$\pm$4.41 & 19.55$\pm$3.18 & 13.34$\pm$2.45 & 11.09$\pm$1.71 & 9.92$\pm$1.00 & 9.77$\pm$0.33\\
        & IRM        & 97.28$\pm$1.04 & 97.22$\pm$0.74 & 98.56$\pm$0.48 & 97.92$\pm$0.42 & 94.42$\pm$2.19 & 75.65$\pm$10.29 & 47.50$\pm$13.40 & 28.62$\pm$9.74 & 21.57$\pm$6.09 & 17.55$\pm$4.50 & 14.66$\pm$2.79\\
        & SD         & 98.41$\pm$0.46 & 98.48$\pm$0.23 & 99.01$\pm$0.33 & 98.72$\pm$0.30 & 96.80$\pm$1.32 & 82.03$\pm$13.74 & 59.44$\pm$22.35 & 40.30$\pm$18.68 & 27.71$\pm$13.28 & 20.90$\pm$8.94 & 17.53$\pm$6.83\\
        & GroupDRO   & 98.55$\pm$0.23 & 98.53$\pm$0.30 & 98.78$\pm$0.39 & 95.56$\pm$2.98 & 77.70$\pm$9.02 & 47.54$\pm$16.06 & 28.06$\pm$9.64 & 18.09$\pm$3.63 & 13.90$\pm$2.36 & 12.14$\pm$1.62 & 11.17$\pm$0.72\\
        & RSC        & 98.43$\pm$0.34 & 98.43$\pm$0.26 & 98.89$\pm$0.17 & 96.45$\pm$2.09 & 83.98$\pm$7.28 & 56.62$\pm$11.88 & 32.84$\pm$10.32 & 19.78$\pm$4.91 & 13.92$\pm$2.73 & 11.84$\pm$1.22 & 11.03$\pm$0.81\\
        & Mixup      & 96.76$\pm$1.60 & 96.63$\pm$1.67 & 97.66$\pm$1.27 & 94.62$\pm$2.32 & 76.02$\pm$11.82 & 40.51$\pm$16.47 & 22.26$\pm$11.68 & 14.79$\pm$5.18 & 12.08$\pm$2.22 & 11.17$\pm$1.59 & 11.40$\pm$1.48\\
\bottomrule
    \end{tabular}}
    \caption{Average accuracy of top-3 models of each algorithm at each shift degree of \textsc{NoisyMNIST} (ResNet-50). This table shows the results on shift degrees from $\mathcal{D}_{4}$ to $\mathcal{D}_{7}$.}
    \label{tab:noisymnist-rn50-4-7}
\end{table}
\begin{table}[t]
    \centering
    \adjustbox{max width=\textwidth}{
    \begin{tabular}{llcccccccccccc}
        \toprule
        & Algorithm & $\mathcal{D}_{0}$ & $\mathcal{D}_{1}$ & $\mathcal{D}_{2}$ & $\mathcal{D}_{3}$ & $\mathcal{D}_{4}$ & $\mathcal{D}_{5}$ & $\mathcal{D}_{6}$ & $\mathcal{D}_{7}$ & $\mathcal{D}_{8}$ & $\mathcal{D}_{9}$ & $\mathcal{D}_{10}$\\
        \midrule
        \multirow{28}{*}{$\mathcal{D}_{8}$}
        & ERM        & 96.56$\pm$0.99 & 96.34$\pm$1.15 & 98.14$\pm$0.64 & 97.48$\pm$0.44 & 93.87$\pm$3.59 & 80.47$\pm$12.14 & 58.96$\pm$15.06 & 38.30$\pm$11.71 & 27.11$\pm$7.83 & 21.21$\pm$5.80 & 17.77$\pm$3.94\\
        & VREx       & 28.86$\pm$17.68 & 28.82$\pm$17.41 & 32.37$\pm$19.99 & 31.83$\pm$20.38 & 28.59$\pm$18.03 & 22.19$\pm$11.03 & 16.81$\pm$5.04 & 14.38$\pm$2.30 & 14.17$\pm$1.23 & 12.13$\pm$1.02 & 12.29$\pm$1.88\\
        & IRM        & 92.49$\pm$7.45 & 92.36$\pm$7.32 & 95.57$\pm$4.87 & 94.69$\pm$5.16 & 92.37$\pm$5.46 & 76.98$\pm$4.77 & 50.67$\pm$8.41 & 32.44$\pm$6.46 & 24.34$\pm$4.02 & 19.33$\pm$3.37 & 15.89$\pm$1.86\\
        & SD         & 97.82$\pm$0.92 & 97.86$\pm$1.04 & 98.64$\pm$0.47 & 97.98$\pm$0.60 & 95.64$\pm$2.74 & 87.08$\pm$6.69 & 70.35$\pm$10.00 & 48.45$\pm$10.54 & 32.74$\pm$9.21 & 23.56$\pm$7.24 & 19.63$\pm$5.70\\
        & GroupDRO   & 98.33$\pm$0.65 & 98.25$\pm$0.68 & 98.57$\pm$0.37 & 91.84$\pm$1.77 & 68.88$\pm$5.26 & 42.05$\pm$5.96 & 27.30$\pm$4.56 & 18.60$\pm$3.33 & 15.51$\pm$2.16 & 13.18$\pm$1.66 & 11.79$\pm$0.95\\
        & RSC        & 96.77$\pm$1.55 & 96.96$\pm$1.57 & 96.97$\pm$1.37 & 87.84$\pm$4.74 & 66.60$\pm$10.24 & 45.28$\pm$8.74 & 30.28$\pm$4.27 & 21.46$\pm$2.16 & 17.17$\pm$1.84 & 13.64$\pm$1.81 & 12.71$\pm$2.11\\
        & Mixup      & 79.28$\pm$34.61 & 79.17$\pm$34.63 & 80.00$\pm$35.00 & 78.49$\pm$34.13 & 64.45$\pm$29.61 & 36.96$\pm$20.33 & 21.69$\pm$12.16 & 15.22$\pm$5.09 & 12.81$\pm$1.49 & 11.60$\pm$1.11 & 11.68$\pm$1.23\\
\bottomrule
    \end{tabular}}
    \caption{Average accuracy of top-3 models of each algorithm at each shift degree of \textsc{NoisyMNIST} (ResNet-50). This table shows the results on shift degrees from $\mathcal{D}_{8}$ to $\mathcal{D}_{10}$.}
    \label{tab:noisymnist-rn50-8-10}
\end{table}
\begin{table}[t]
    \centering
    \adjustbox{max width=\textwidth}{
    \begin{tabular}{llcccccccccccc}
        \toprule
        & Algorithm & $\mathcal{D}_{0}$ & $\mathcal{D}_{1}$ & $\mathcal{D}_{2}$ & $\mathcal{D}_{3}$ & $\mathcal{D}_{4}$ & $\mathcal{D}_{5}$ & $\mathcal{D}_{6}$ & $\mathcal{D}_{7}$ & $\mathcal{D}_{8}$ & $\mathcal{D}_{9}$ & $\mathcal{D}_{10}$\\
        \midrule
        \multirow{11}{*}{$\mathcal{D}_{0}$}
        & ERM        & 83.02$\pm$0.29 & 82.04$\pm$0.57 & 81.29$\pm$0.46 & 79.34$\pm$0.93 & 73.05$\pm$2.19 & 63.89$\pm$3.49 & 53.77$\pm$4.75 & 38.88$\pm$4.62 & 28.90$\pm$3.48 & 19.27$\pm$2.01 & 15.06$\pm$1.34\\
        & VREx       & 82.24$\pm$0.69 & 81.27$\pm$0.53 & 80.73$\pm$0.83 & 79.03$\pm$0.54 & 72.34$\pm$1.17 & 63.48$\pm$1.35 & 51.59$\pm$1.98 & 36.78$\pm$3.48 & 26.05$\pm$3.42 & 18.38$\pm$2.74 & 14.66$\pm$2.04\\
        & IRM        & 78.50$\pm$6.68 & 77.96$\pm$6.76 & 77.66$\pm$7.20 & 76.16$\pm$6.86 & 70.12$\pm$6.59 & 63.43$\pm$5.26 & 52.90$\pm$4.60 & 38.83$\pm$4.75 & 27.85$\pm$4.88 & 19.24$\pm$6.02 & 14.72$\pm$4.04\\
        & SD         & 83.12$\pm$0.17 & 81.90$\pm$0.64 & 81.23$\pm$0.53 & 79.96$\pm$1.15 & 72.21$\pm$1.80 & 63.86$\pm$3.50 & 53.10$\pm$4.83 & 38.79$\pm$5.09 & 28.86$\pm$4.20 & 20.16$\pm$2.51 & 15.85$\pm$1.15\\
        & CORAL      & 82.96$\pm$0.28 & 82.08$\pm$0.86 & 81.57$\pm$0.70 & 80.70$\pm$1.12 & 74.53$\pm$1.43 & 67.62$\pm$2.57 & 56.67$\pm$3.82 & 41.09$\pm$4.75 & 30.44$\pm$5.94 & 20.62$\pm$5.44 & 15.50$\pm$3.03\\
        & GroupDRO   & 82.92$\pm$0.19 & 81.71$\pm$1.01 & 80.98$\pm$1.16 & 79.43$\pm$1.02 & 73.46$\pm$0.91 & 63.80$\pm$1.74 & 53.83$\pm$3.04 & 38.55$\pm$3.65 & 28.81$\pm$3.27 & 20.02$\pm$3.08 & 14.62$\pm$1.48\\
        & RSC        & 82.54$\pm$0.12 & 81.84$\pm$0.59 & 81.93$\pm$1.27 & 79.57$\pm$0.87 & 72.89$\pm$1.80 & 63.54$\pm$1.83 & 52.82$\pm$3.07 & 37.61$\pm$3.93 & 27.69$\pm$4.12 & 19.14$\pm$3.27 & 14.99$\pm$2.24\\
\bottomrule
    \end{tabular}}
    \caption{Average accuracy of top-3 models of each algorithm at each shift degree of \textsc{LowLightCIFAR10} (Simple CNN). This table shows the results on shift degrees from $\mathcal{D}_{0}$ to $\mathcal{D}_{3}$.}
    \label{tab:lowlightcifar10-cnn-0-3}
\end{table}
\begin{table}[t]
    \centering
    \adjustbox{max width=\textwidth}{
    \begin{tabular}{llcccccccccccc}
        \toprule
        & Algorithm & $\mathcal{D}_{0}$ & $\mathcal{D}_{1}$ & $\mathcal{D}_{2}$ & $\mathcal{D}_{3}$ & $\mathcal{D}_{4}$ & $\mathcal{D}_{5}$ & $\mathcal{D}_{6}$ & $\mathcal{D}_{7}$ & $\mathcal{D}_{8}$ & $\mathcal{D}_{9}$ & $\mathcal{D}_{10}$\\
        \midrule
        \multirow{11}{*}{$\mathcal{D}_{4}$}
        & ERM        & 80.86$\pm$0.97 & 81.33$\pm$0.66 & 82.24$\pm$1.23 & 81.41$\pm$0.84 & 76.84$\pm$0.72 & 70.41$\pm$1.45 & 59.93$\pm$2.24 & 44.25$\pm$2.16 & 32.77$\pm$2.72 & 22.49$\pm$2.70 & 16.33$\pm$1.75\\
        & VREx       & 80.05$\pm$1.13 & 80.25$\pm$0.93 & 81.02$\pm$0.97 & 80.25$\pm$0.57 & 75.29$\pm$0.42 & 67.77$\pm$1.68 & 58.44$\pm$2.26 & 43.36$\pm$1.67 & 32.72$\pm$2.59 & 22.47$\pm$1.88 & 15.83$\pm$0.84\\
        & IRM        & 77.50$\pm$6.62 & 77.32$\pm$6.73 & 78.28$\pm$6.50 & 77.88$\pm$6.08 & 72.76$\pm$5.92 & 66.91$\pm$4.78 & 58.23$\pm$2.76 & 44.84$\pm$1.25 & 33.80$\pm$2.12 & 24.06$\pm$2.79 & 17.61$\pm$2.08\\
        & SD         & 80.82$\pm$0.87 & 81.24$\pm$0.70 & 82.62$\pm$0.52 & 81.69$\pm$0.22 & 77.18$\pm$0.69 & 70.29$\pm$0.63 & 60.85$\pm$1.53 & 45.78$\pm$2.17 & 34.70$\pm$1.98 & 23.49$\pm$2.00 & 16.47$\pm$1.19\\
        & CORAL      & 80.38$\pm$0.64 & 80.94$\pm$0.50 & 81.66$\pm$0.61 & 81.08$\pm$0.71 & 77.47$\pm$0.39 & 70.20$\pm$0.60 & 60.61$\pm$0.60 & 45.91$\pm$2.68 & 34.16$\pm$4.42 & 23.45$\pm$5.29 & 17.17$\pm$3.80\\
        & GroupDRO   & 81.40$\pm$0.43 & 81.40$\pm$0.67 & 81.83$\pm$0.42 & 81.25$\pm$0.66 & 76.50$\pm$0.10 & 68.74$\pm$0.33 & 59.41$\pm$0.42 & 45.26$\pm$0.84 & 34.03$\pm$1.02 & 23.10$\pm$1.48 & 15.98$\pm$0.87\\
        & RSC        & 81.73$\pm$0.32 & 81.85$\pm$0.90 & 82.28$\pm$0.65 & 80.97$\pm$0.53 & 76.07$\pm$0.18 & 67.88$\pm$1.84 & 57.74$\pm$1.45 & 43.50$\pm$1.43 & 30.63$\pm$1.33 & 20.54$\pm$1.42 & 15.03$\pm$1.06\\
\bottomrule
    \end{tabular}}
    \caption{Average accuracy of top-3 models of each algorithm at each shift degree of \textsc{LowLightCIFAR10} (Simple CNN). This table shows the results on shift degrees from $\mathcal{D}_{4}$ to $\mathcal{D}_{7}$.}
    \label{tab:lowlightcifar10-cnn-4-7}
\end{table}
\begin{table}[t]
    \centering
    \adjustbox{max width=\textwidth}{
    \begin{tabular}{llcccccccccccc}
        \toprule
        & Algorithm & $\mathcal{D}_{0}$ & $\mathcal{D}_{1}$ & $\mathcal{D}_{2}$ & $\mathcal{D}_{3}$ & $\mathcal{D}_{4}$ & $\mathcal{D}_{5}$ & $\mathcal{D}_{6}$ & $\mathcal{D}_{7}$ & $\mathcal{D}_{8}$ & $\mathcal{D}_{9}$ & $\mathcal{D}_{10}$\\
        \midrule
        \multirow{11}{*}{$\mathcal{D}_{8}$}
        & ERM        & 76.67$\pm$5.69 & 76.66$\pm$5.15 & 77.51$\pm$5.31 & 76.13$\pm$5.54 & 71.71$\pm$4.58 & 66.37$\pm$3.32 & 59.75$\pm$1.68 & 47.76$\pm$1.72 & 38.25$\pm$1.21 & 26.45$\pm$2.45 & 18.69$\pm$1.25\\
        & VREx       & 64.56$\pm$13.89 & 64.58$\pm$13.22 & 65.78$\pm$12.55 & 65.46$\pm$12.44 & 62.40$\pm$11.40 & 59.00$\pm$8.49 & 54.60$\pm$6.64 & 44.65$\pm$3.51 & 37.03$\pm$1.97 & 27.20$\pm$1.73 & 20.62$\pm$2.09\\
        & IRM        & 58.08$\pm$14.34 & 59.18$\pm$14.11 & 61.37$\pm$11.98 & 61.00$\pm$11.41 & 58.71$\pm$9.97 & 58.62$\pm$7.08 & 54.56$\pm$4.14 & 48.18$\pm$1.04 & 41.43$\pm$2.36 & 31.09$\pm$3.85 & 22.49$\pm$3.61\\
        & SD         & 77.66$\pm$1.88 & 77.41$\pm$2.18 & 78.14$\pm$1.36 & 77.72$\pm$1.33 & 72.46$\pm$1.88 & 67.79$\pm$1.72 & 60.39$\pm$1.54 & 48.15$\pm$1.63 & 39.92$\pm$1.37 & 28.88$\pm$1.59 & 20.18$\pm$2.27\\
        & CORAL      & 78.02$\pm$5.21 & 78.84$\pm$4.62 & 79.18$\pm$5.10 & 78.19$\pm$5.20 & 74.13$\pm$5.18 & 68.60$\pm$3.38 & 59.63$\pm$2.12 & 47.55$\pm$1.77 & 38.44$\pm$1.58 & 28.39$\pm$3.05 & 20.48$\pm$2.55\\
        & GroupDRO   & 75.37$\pm$4.50 & 75.31$\pm$3.86 & 75.82$\pm$3.70 & 75.27$\pm$3.60 & 70.63$\pm$3.24 & 66.82$\pm$1.49 & 59.22$\pm$1.85 & 46.67$\pm$2.20 & 38.08$\pm$1.58 & 28.59$\pm$1.65 & 20.98$\pm$0.89\\
        & RSC        & 74.87$\pm$5.81 & 75.34$\pm$5.79 & 75.66$\pm$5.49 & 75.25$\pm$4.95 & 70.39$\pm$4.30 & 66.60$\pm$2.39 & 59.45$\pm$0.66 & 46.96$\pm$1.45 & 37.24$\pm$1.29 & 27.57$\pm$1.36 & 20.63$\pm$1.90\\
\bottomrule
    \end{tabular}}
    \caption{Average accuracy of top-3 models of each algorithm at each shift degree of \textsc{LowLightCIFAR10} (Simple CNN). This table shows the results on shift degrees from $\mathcal{D}_{8}$ to $\mathcal{D}_{10}$.}
    \label{tab:lowlightcifar10-cnn-8-10}
\end{table}
\begin{table}[t]
    \centering
    \adjustbox{max width=\textwidth}{
    \begin{tabular}{llcccccccccccc}
        \toprule
        & Algorithm & $\mathcal{D}_{0}$ & $\mathcal{D}_{1}$ & $\mathcal{D}_{2}$ & $\mathcal{D}_{3}$ & $\mathcal{D}_{4}$ & $\mathcal{D}_{5}$ & $\mathcal{D}_{6}$ & $\mathcal{D}_{7}$ & $\mathcal{D}_{8}$ & $\mathcal{D}_{9}$ & $\mathcal{D}_{10}$\\
        \midrule
        \multirow{19}{*}{$\mathcal{D}_{0}$}
        & ERM        & 80.89$\pm$0.17 & 79.67$\pm$0.92 & 80.27$\pm$0.67 & 77.58$\pm$0.65 & 70.74$\pm$2.09 & 63.27$\pm$2.96 & 50.29$\pm$3.42 & 35.72$\pm$4.30 & 26.66$\pm$3.80 & 19.46$\pm$3.75 & 15.11$\pm$2.61\\
        & VREx       & 77.38$\pm$4.17 & 76.90$\pm$4.02 & 77.94$\pm$3.64 & 75.83$\pm$3.75 & 69.48$\pm$3.69 & 62.26$\pm$4.06 & 49.16$\pm$3.75 & 33.77$\pm$3.67 & 23.30$\pm$3.18 & 15.76$\pm$2.50 & 12.64$\pm$1.69\\
        & IRM        & 73.16$\pm$10.11 & 72.53$\pm$10.27 & 73.25$\pm$9.20 & 70.42$\pm$9.97 & 63.45$\pm$9.62 & 56.59$\pm$8.61 & 45.31$\pm$7.92 & 32.89$\pm$6.08 & 23.40$\pm$4.44 & 15.61$\pm$3.22 & 13.23$\pm$3.01\\
        & SD         & 81.28$\pm$0.13 & 80.28$\pm$0.76 & 79.78$\pm$0.49 & 78.14$\pm$0.72 & 71.88$\pm$1.89 & 63.25$\pm$2.85 & 51.09$\pm$3.23 & 36.83$\pm$2.47 & 26.91$\pm$1.89 & 19.10$\pm$1.73 & 15.95$\pm$1.45\\
        & CORAL      & 79.60$\pm$0.40 & 78.53$\pm$0.50 & 77.97$\pm$0.64 & 74.99$\pm$0.91 & 66.85$\pm$1.49 & 57.74$\pm$1.74 & 43.87$\pm$1.86 & 28.48$\pm$1.90 & 19.22$\pm$1.97 & 13.15$\pm$1.63 & 12.07$\pm$1.55\\
        & GroupDRO   & 80.30$\pm$0.21 & 79.73$\pm$1.00 & 80.22$\pm$0.78 & 77.21$\pm$0.78 & 70.12$\pm$1.50 & 62.25$\pm$2.64 & 48.43$\pm$3.31 & 32.17$\pm$3.64 & 22.26$\pm$3.53 & 15.12$\pm$2.77 & 13.14$\pm$2.69\\
        & RSC        & 79.71$\pm$0.59 & 78.90$\pm$0.82 & 80.13$\pm$1.14 & 77.74$\pm$0.92 & 72.22$\pm$1.67 & 65.34$\pm$1.71 & 53.19$\pm$3.03 & 37.32$\pm$3.07 & 26.92$\pm$2.24 & 17.67$\pm$2.10 & 14.16$\pm$0.95\\
\bottomrule
    \end{tabular}}
    \caption{Average accuracy of top-3 models of each algorithm at each shift degree of \textsc{LowLightCIFAR10} (EfficientNet-b0). This table shows the results on shift degrees from $\mathcal{D}_{0}$ to $\mathcal{D}_{3}$.}
    \label{tab:lowlightcifar10-enb0-0-3}
\end{table}
\begin{table}[t]
    \centering
    \adjustbox{max width=\textwidth}{
    \begin{tabular}{llcccccccccccc}
        \toprule
        & Algorithm & $\mathcal{D}_{0}$ & $\mathcal{D}_{1}$ & $\mathcal{D}_{2}$ & $\mathcal{D}_{3}$ & $\mathcal{D}_{4}$ & $\mathcal{D}_{5}$ & $\mathcal{D}_{6}$ & $\mathcal{D}_{7}$ & $\mathcal{D}_{8}$ & $\mathcal{D}_{9}$ & $\mathcal{D}_{10}$\\
        \midrule
        \multirow{19}{*}{$\mathcal{D}_{4}$}
        & ERM        & 80.30$\pm$0.48 & 80.24$\pm$0.36 & 81.05$\pm$0.80 & 78.96$\pm$0.84 & 73.98$\pm$0.47 & 67.33$\pm$1.69 & 55.16$\pm$2.63 & 39.38$\pm$4.92 & 27.93$\pm$4.70 & 18.38$\pm$3.48 & 13.74$\pm$2.19\\
        & VREx       & 76.32$\pm$4.94 & 75.78$\pm$4.74 & 77.48$\pm$4.05 & 75.75$\pm$4.07 & 70.35$\pm$2.80 & 63.90$\pm$2.47 & 52.31$\pm$4.59 & 38.37$\pm$6.80 & 27.26$\pm$6.61 & 18.82$\pm$4.88 & 14.69$\pm$3.57\\
        & IRM        & 72.27$\pm$9.72 & 72.01$\pm$10.05 & 73.20$\pm$9.27 & 70.25$\pm$9.93 & 64.81$\pm$10.14 & 58.27$\pm$9.21 & 46.43$\pm$8.12 & 32.93$\pm$5.75 & 23.39$\pm$4.23 & 15.90$\pm$2.97 & 13.61$\pm$2.77\\
        & SD         & 80.46$\pm$0.53 & 80.66$\pm$0.28 & 81.12$\pm$0.63 & 79.90$\pm$0.60 & 74.67$\pm$0.88 & 67.16$\pm$0.84 & 55.02$\pm$0.87 & 39.38$\pm$1.38 & 27.74$\pm$1.22 & 18.27$\pm$1.82 & 15.31$\pm$1.44\\
        & CORAL      & 77.26$\pm$1.39 & 77.69$\pm$1.61 & 78.69$\pm$0.92 & 77.03$\pm$0.42 & 71.46$\pm$0.53 & 63.20$\pm$2.23 & 51.66$\pm$3.20 & 36.49$\pm$2.70 & 25.76$\pm$1.45 & 17.46$\pm$0.87 & 13.98$\pm$1.00\\
        & GroupDRO   & 79.43$\pm$0.78 & 79.43$\pm$0.80 & 79.97$\pm$1.29 & 77.80$\pm$0.86 & 71.92$\pm$0.93 & 65.20$\pm$1.43 & 52.90$\pm$2.09 & 36.21$\pm$2.47 & 25.19$\pm$2.02 & 16.55$\pm$1.68 & 13.10$\pm$1.19\\
        & RSC        & 78.06$\pm$1.60 & 78.72$\pm$0.88 & 79.95$\pm$1.29 & 78.32$\pm$0.60 & 73.29$\pm$0.72 & 66.84$\pm$0.97 & 55.37$\pm$1.53 & 39.32$\pm$1.51 & 27.81$\pm$1.74 & 17.42$\pm$2.15 & 13.89$\pm$1.46\\
\bottomrule
    \end{tabular}}
    \caption{Average accuracy of top-3 models of each algorithm at each shift degree of \textsc{LowLightCIFAR10} (EfficientNet-b0). This table shows the results on shift degrees from $\mathcal{D}_{4}$ to $\mathcal{D}_{7}$.}
    \label{tab:lowlightcifar10-enb0-4-7}
\end{table}
\begin{table}[t]
    \centering
    \adjustbox{max width=\textwidth}{
    \begin{tabular}{llcccccccccccc}
        \toprule
        & Algorithm & $\mathcal{D}_{0}$ & $\mathcal{D}_{1}$ & $\mathcal{D}_{2}$ & $\mathcal{D}_{3}$ & $\mathcal{D}_{4}$ & $\mathcal{D}_{5}$ & $\mathcal{D}_{6}$ & $\mathcal{D}_{7}$ & $\mathcal{D}_{8}$ & $\mathcal{D}_{9}$ & $\mathcal{D}_{10}$\\
        \midrule
        \multirow{19}{*}{$\mathcal{D}_{8}$}
        & ERM        & 74.49$\pm$5.14 & 74.48$\pm$5.14 & 75.80$\pm$4.92 & 74.00$\pm$3.99 & 68.86$\pm$4.22 & 63.68$\pm$3.71 & 54.67$\pm$3.78 & 43.07$\pm$2.36 & 34.40$\pm$2.13 & 24.31$\pm$2.00 & 18.03$\pm$2.27\\
        & VREx       & 70.14$\pm$7.18 & 69.66$\pm$6.78 & 72.25$\pm$6.03 & 69.74$\pm$6.60 & 65.14$\pm$6.72 & 60.34$\pm$6.06 & 51.35$\pm$5.22 & 39.69$\pm$5.25 & 29.66$\pm$4.59 & 20.57$\pm$3.37 & 15.73$\pm$3.07\\
        & IRM        & 68.22$\pm$10.11 & 67.98$\pm$9.83 & 69.07$\pm$9.40 & 66.26$\pm$10.67 & 61.68$\pm$10.63 & 56.16$\pm$9.59 & 47.76$\pm$8.06 & 37.10$\pm$5.32 & 28.35$\pm$3.64 & 19.67$\pm$1.91 & 16.30$\pm$2.06\\
        & SD         & 71.38$\pm$4.07 & 71.03$\pm$4.33 & 73.23$\pm$3.96 & 71.05$\pm$3.62 & 65.96$\pm$3.51 & 62.28$\pm$2.24 & 52.85$\pm$1.74 & 41.63$\pm$0.53 & 31.98$\pm$0.31 & 22.05$\pm$1.28 & 16.83$\pm$1.08\\
        & CORAL      & 67.89$\pm$6.24 & 67.87$\pm$6.19 & 69.55$\pm$5.98 & 67.24$\pm$7.11 & 61.90$\pm$6.83 & 57.28$\pm$7.16 & 49.56$\pm$5.66 & 38.81$\pm$2.99 & 30.75$\pm$0.96 & 21.53$\pm$0.66 & 17.93$\pm$1.19\\
        & GroupDRO   & 72.87$\pm$8.17 & 72.66$\pm$8.13 & 73.34$\pm$8.77 & 71.07$\pm$8.27 & 65.65$\pm$7.45 & 60.06$\pm$6.38 & 51.09$\pm$4.24 & 40.28$\pm$1.80 & 32.46$\pm$1.90 & 23.48$\pm$2.67 & 19.05$\pm$2.65\\
        & RSC        & 69.86$\pm$6.08 & 70.78$\pm$5.45 & 71.55$\pm$5.30 & 69.28$\pm$5.34 & 65.29$\pm$4.83 & 60.60$\pm$4.35 & 53.76$\pm$3.86 & 42.95$\pm$2.58 & 34.78$\pm$2.24 & 23.92$\pm$1.35 & 17.91$\pm$0.62\\
\bottomrule
    \end{tabular}}
    \caption{Average accuracy of top-3 models of each algorithm at each shift degree of \textsc{LowLightCIFAR10} (EfficientNet-b0). This table shows the results on shift degrees from $\mathcal{D}_{8}$ to $\mathcal{D}_{10}$.}
    \label{tab:lowlightcifar10-enb0-8-10}
\end{table}
\begin{table}[t]
    \centering
    \adjustbox{max width=\textwidth}{
    \begin{tabular}{llcccccccccccc}
        \toprule
        & Algorithm & $\mathcal{D}_{0}$ & $\mathcal{D}_{1}$ & $\mathcal{D}_{2}$ & $\mathcal{D}_{3}$ & $\mathcal{D}_{4}$ & $\mathcal{D}_{5}$ & $\mathcal{D}_{6}$ & $\mathcal{D}_{7}$ & $\mathcal{D}_{8}$ & $\mathcal{D}_{9}$ & $\mathcal{D}_{10}$\\
        \midrule
        \multirow{11}{*}{$\mathcal{D}_{0}$}
        & ERM        & 83.73$\pm$0.21 & 82.91$\pm$0.98 & 14.62$\pm$1.27 & 15.14$\pm$1.57 & 14.80$\pm$1.17 & 14.93$\pm$1.37 & 13.86$\pm$1.93 & 14.65$\pm$2.07 & 13.26$\pm$1.97 & 13.01$\pm$2.35 & 12.96$\pm$1.77\\
        & VREx       & 78.76$\pm$4.22 & 78.33$\pm$4.15 & 16.51$\pm$0.77 & 16.97$\pm$1.17 & 16.27$\pm$1.46 & 16.76$\pm$1.58 & 15.99$\pm$2.16 & 16.35$\pm$1.79 & 14.82$\pm$1.64 & 13.16$\pm$1.24 & 12.99$\pm$1.64\\
        & IRM        & 54.36$\pm$16.65 & 54.08$\pm$16.43 & 13.99$\pm$2.64 & 15.15$\pm$3.08 & 14.54$\pm$3.19 & 15.17$\pm$2.56 & 15.97$\pm$3.05 & 14.92$\pm$2.86 & 14.86$\pm$2.19 & 13.24$\pm$2.88 & 13.19$\pm$2.80\\
        & SD         & 84.10$\pm$0.43 & 83.41$\pm$1.10 & 16.65$\pm$2.52 & 17.01$\pm$1.67 & 16.33$\pm$1.53 & 16.40$\pm$0.80 & 15.76$\pm$0.74 & 15.67$\pm$0.48 & 14.60$\pm$0.68 & 13.68$\pm$0.95 & 13.59$\pm$0.33\\
        & CORAL      & 84.54$\pm$0.40 & 83.94$\pm$0.60 & 16.35$\pm$0.94 & 16.98$\pm$1.16 & 15.50$\pm$0.94 & 16.55$\pm$1.73 & 15.60$\pm$0.84 & 15.54$\pm$0.79 & 14.99$\pm$0.45 & 13.78$\pm$0.60 & 13.97$\pm$0.58\\
        & GroupDRO   & 84.72$\pm$0.23 & 83.90$\pm$0.34 & 16.04$\pm$0.78 & 15.86$\pm$0.66 & 15.24$\pm$0.91 & 15.30$\pm$1.01 & 14.70$\pm$0.64 & 13.95$\pm$1.68 & 13.79$\pm$0.77 & 13.17$\pm$0.72 & 13.21$\pm$0.82\\
        & RSC        & 84.56$\pm$0.41 & 83.70$\pm$0.37 & 11.25$\pm$0.91 & 10.55$\pm$1.04 & 10.04$\pm$1.23 & 10.88$\pm$1.06 & 11.28$\pm$1.19 & 10.78$\pm$1.38 & 11.19$\pm$1.09 & 10.56$\pm$1.56 & 10.39$\pm$1.00\\
\bottomrule
    \end{tabular}}
    \caption{Average accuracy of top-3 models of each algorithm at each shift degree of \textsc{LowLightCIFAR10} (ResNet-50). This table shows the results on shift degrees from $\mathcal{D}_{0}$ to $\mathcal{D}_{3}$.}
    \label{tab:lowlightcifar10-rn50-0-3}
\end{table}
\begin{table}[t]
    \centering
    \adjustbox{max width=\textwidth}{
    \begin{tabular}{llcccccccccccc}
        \toprule
        & Algorithm & $\mathcal{D}_{0}$ & $\mathcal{D}_{1}$ & $\mathcal{D}_{2}$ & $\mathcal{D}_{3}$ & $\mathcal{D}_{4}$ & $\mathcal{D}_{5}$ & $\mathcal{D}_{6}$ & $\mathcal{D}_{7}$ & $\mathcal{D}_{8}$ & $\mathcal{D}_{9}$ & $\mathcal{D}_{10}$\\
        \midrule
        \multirow{11}{*}{$\mathcal{D}_{4}$}
        & ERM        & 75.43$\pm$4.16 & 75.43$\pm$3.88 & 19.92$\pm$1.04 & 20.44$\pm$0.97 & 19.99$\pm$0.57 & 19.85$\pm$0.37 & 18.16$\pm$1.39 & 17.60$\pm$1.07 & 17.14$\pm$1.50 & 15.74$\pm$1.86 & 15.59$\pm$1.24\\
        & VREx       & 70.04$\pm$6.37 & 69.72$\pm$7.41 & 17.66$\pm$1.51 & 17.79$\pm$0.77 & 18.36$\pm$0.41 & 18.49$\pm$1.24 & 17.96$\pm$1.05 & 18.00$\pm$1.39 & 16.71$\pm$1.07 & 14.74$\pm$1.52 & 15.00$\pm$1.56\\
        & IRM        & 46.99$\pm$20.23 & 47.23$\pm$20.15 & 15.76$\pm$2.40 & 16.39$\pm$1.91 & 16.37$\pm$2.81 & 16.97$\pm$2.34 & 16.72$\pm$1.56 & 15.90$\pm$2.33 & 15.52$\pm$1.76 & 13.45$\pm$2.17 & 13.17$\pm$2.37\\
        & SD         & 72.47$\pm$6.68 & 72.93$\pm$6.66 & 19.61$\pm$1.59 & 20.13$\pm$0.93 & 19.91$\pm$0.73 & 19.20$\pm$0.73 & 19.02$\pm$0.83 & 17.37$\pm$1.44 & 16.00$\pm$1.13 & 15.24$\pm$0.58 & 14.57$\pm$1.23\\
        & CORAL      & 70.90$\pm$5.34 & 70.37$\pm$5.55 & 18.96$\pm$1.63 & 19.49$\pm$1.67 & 19.64$\pm$0.54 & 19.42$\pm$0.72 & 18.30$\pm$0.87 & 16.92$\pm$1.22 & 16.59$\pm$0.77 & 15.15$\pm$0.86 & 14.82$\pm$1.42\\
        & GroupDRO   & 69.94$\pm$9.64 & 70.20$\pm$9.04 & 18.79$\pm$1.47 & 19.24$\pm$1.00 & 19.51$\pm$0.37 & 19.40$\pm$1.73 & 17.82$\pm$1.87 & 16.54$\pm$2.61 & 16.02$\pm$3.00 & 14.56$\pm$1.88 & 13.52$\pm$2.26\\
        & RSC        & 55.72$\pm$23.63 & 56.03$\pm$23.83 & 15.84$\pm$1.92 & 15.72$\pm$1.74 & 15.48$\pm$1.73 & 16.07$\pm$2.00 & 15.64$\pm$2.52 & 14.87$\pm$2.04 & 14.47$\pm$2.39 & 13.36$\pm$1.70 & 13.26$\pm$2.46\\
\bottomrule
    \end{tabular}}
    \caption{Average accuracy of top-3 models of each algorithm at each shift degree of \textsc{LowLightCIFAR10} (ResNet-50). This table shows the results on shift degrees from $\mathcal{D}_{4}$ to $\mathcal{D}_{7}$.}
    \label{tab:lowlightcifar10-rn50-4-7}
\end{table}
\begin{table}[t]
    \centering
    \adjustbox{max width=\textwidth}{
    \begin{tabular}{llcccccccccccc}
        \toprule
        & Algorithm & $\mathcal{D}_{0}$ & $\mathcal{D}_{1}$ & $\mathcal{D}_{2}$ & $\mathcal{D}_{3}$ & $\mathcal{D}_{4}$ & $\mathcal{D}_{5}$ & $\mathcal{D}_{6}$ & $\mathcal{D}_{7}$ & $\mathcal{D}_{8}$ & $\mathcal{D}_{9}$ & $\mathcal{D}_{10}$\\
        \midrule
        \multirow{11}{*}{$\mathcal{D}_{8}$}
        & ERM        & 74.12$\pm$6.53 & 73.54$\pm$7.17 & 18.34$\pm$1.26 & 19.28$\pm$0.86 & 19.20$\pm$0.90 & 19.48$\pm$0.56 & 19.19$\pm$0.77 & 19.28$\pm$0.72 & 18.93$\pm$0.40 & 17.92$\pm$0.99 & 17.07$\pm$0.61\\
        & VREx       & 62.97$\pm$5.02 & 62.59$\pm$5.32 & 16.74$\pm$2.74 & 17.26$\pm$1.62 & 17.14$\pm$1.36 & 18.14$\pm$1.36 & 18.51$\pm$0.85 & 18.03$\pm$1.73 & 17.43$\pm$0.86 & 15.73$\pm$0.96 & 15.27$\pm$1.43\\
        & IRM        & 39.06$\pm$25.31 & 39.60$\pm$25.33 & 14.35$\pm$3.30 & 14.97$\pm$2.60 & 14.83$\pm$3.23 & 15.99$\pm$2.58 & 16.47$\pm$1.57 & 15.82$\pm$1.84 & 16.05$\pm$1.53 & 13.82$\pm$1.90 & 13.70$\pm$2.03\\
        & SD         & 72.73$\pm$6.99 & 73.30$\pm$7.15 & 16.67$\pm$0.51 & 17.96$\pm$0.88 & 17.83$\pm$0.76 & 18.49$\pm$1.57 & 19.46$\pm$1.77 & 19.65$\pm$1.64 & 19.02$\pm$0.23 & 17.62$\pm$0.12 & 17.82$\pm$1.10\\
        & CORAL      & 75.53$\pm$8.04 & 75.70$\pm$7.97 & 18.17$\pm$1.41 & 18.70$\pm$0.85 & 17.91$\pm$0.64 & 18.99$\pm$0.51 & 18.81$\pm$0.88 & 18.26$\pm$0.99 & 18.12$\pm$0.63 & 16.19$\pm$1.49 & 15.86$\pm$1.83\\
        & GroupDRO   & 73.10$\pm$8.32 & 73.76$\pm$8.55 & 17.09$\pm$0.74 & 18.29$\pm$0.63 & 18.30$\pm$1.36 & 18.99$\pm$0.82 & 18.61$\pm$1.13 & 18.29$\pm$0.98 & 18.49$\pm$0.68 & 16.58$\pm$0.81 & 15.94$\pm$1.03\\
        & RSC        & 73.43$\pm$9.53 & 73.68$\pm$9.44 & 14.80$\pm$1.82 & 14.67$\pm$1.84 & 14.16$\pm$1.59 & 15.32$\pm$1.71 & 15.89$\pm$1.46 & 15.31$\pm$1.39 & 16.08$\pm$1.12 & 14.75$\pm$1.39 & 14.83$\pm$1.19\\
\bottomrule
    \end{tabular}}
    \caption{Average accuracy of top-3 models of each algorithm at each shift degree of \textsc{LowLightCIFAR10} (ResNet-50). This table shows the results on shift degrees from $\mathcal{D}_{8}$ to $\mathcal{D}_{10}$.}
    \label{tab:lowlightcifar10-rn50-8-10}
\end{table}

\clearpage
\subsection{Complete results of linear probing}
\label{sec:lp-details}
\begin{table}[ht]
    \centering
    \adjustbox{max width=\textwidth}{
    \begin{tabular}{llccccccccccc}
        \toprule
        & & $\mathcal{D}_0$ & $\mathcal{D}_1$ & $\mathcal{D}_2$ & $\mathcal{D}_3$  & $\mathcal{D}_4$ & $\mathcal{D}_5$ & $\mathcal{D}_6$ & $\mathcal{D}_7$ & $\mathcal{D}_8$ & $\mathcal{D}_9$ & $\mathcal{D}_{10}$ \\
        \midrule
        \multirow{10}{*}{\rotatebox[origin=c]{90}{ResNet-50}} & IN$_0$ & 98.4\scriptsize{$\pm$0.1} & 39.8\scriptsize{$\pm$4.2} & 24.8\scriptsize{$\pm$4.7} & 16.7\scriptsize{$\pm$3.2} & 12.4\scriptsize{$\pm$2.1} & 10.9\scriptsize{$\pm$2.0} & 10.3\scriptsize{$\pm$1.5} & 10.1\scriptsize{$\pm$1.4} & 9.8\scriptsize{$\pm$1.1} & 9.6\scriptsize{$\pm$0.8} & 9.5\scriptsize{$\pm$0.6}\\
        & IN$_1$ & 97.9\scriptsize{$\pm$0.1} & 97.3\scriptsize{$\pm$0.2} & 90.2\scriptsize{$\pm$0.7} & 62.9\scriptsize{$\pm$2.0} & 33.1\scriptsize{$\pm$1.7} & 20.1\scriptsize{$\pm$1.9} & 15.4\scriptsize{$\pm$1.7} & 13.3\scriptsize{$\pm$1.1} & 11.7\scriptsize{$\pm$1.1} & 10.8\scriptsize{$\pm$0.9} & 10.5\scriptsize{$\pm$1.0}\\
        & IN$_2$ & 97.6\scriptsize{$\pm$0.2} & 97.6\scriptsize{$\pm$0.4} & 95.8\scriptsize{$\pm$0.4} & 88.2\scriptsize{$\pm$0.5} & 65.4\scriptsize{$\pm$0.8} & 39.9\scriptsize{$\pm$1.8} & 25.6\scriptsize{$\pm$2.6} & 19.1\scriptsize{$\pm$2.8} & 15.7\scriptsize{$\pm$2.1} & 13.5\scriptsize{$\pm$1.3} & 12.2\scriptsize{$\pm$1.0}\\
        & IN$_3$ & 97.6\scriptsize{$\pm$0.2} & 97.2\scriptsize{$\pm$0.1} & 95.9\scriptsize{$\pm$0.1} & 92.6\scriptsize{$\pm$0.8} & 81.8\scriptsize{$\pm$0.4} & 57.9\scriptsize{$\pm$1.3} & 36.2\scriptsize{$\pm$1.6} & 23.5\scriptsize{$\pm$1.7} & 17.4\scriptsize{$\pm$1.9} & 14.3\scriptsize{$\pm$1.6} & 12.7\scriptsize{$\pm$1.1}\\
        & IN$_4$ & 96.6\scriptsize{$\pm$0.3} & 96.6\scriptsize{$\pm$0.5} & 96.1\scriptsize{$\pm$0.6} & 93.5\scriptsize{$\pm$0.6} & 87.4\scriptsize{$\pm$1.0} & 74.4\scriptsize{$\pm$0.6} & 54.4\scriptsize{$\pm$1.4} & 37.3\scriptsize{$\pm$2.0} & 26.5\scriptsize{$\pm$1.7} & 20.4\scriptsize{$\pm$1.1} & 16.9\scriptsize{$\pm$0.9}\\
        & CLIP$_0$ & 98.1\scriptsize{$\pm$0.1} & 55.4\scriptsize{$\pm$4.1} & 35.9\scriptsize{$\pm$4.4} & 26.4\scriptsize{$\pm$3.3} & 21.1\scriptsize{$\pm$2.7} & 18.1\scriptsize{$\pm$1.4} & 15.8\scriptsize{$\pm$0.4} & 14.1\scriptsize{$\pm$0.4} & 13.0\scriptsize{$\pm$0.4} & 12.3\scriptsize{$\pm$0.5} & 12.1\scriptsize{$\pm$0.2}\\
        & CLIP$_1$ & 97.9\scriptsize{$\pm$0.2} & 95.9\scriptsize{$\pm$0.2} & 86.3\scriptsize{$\pm$0.7} & 63.0\scriptsize{$\pm$2.1} & 41.2\scriptsize{$\pm$2.4} & 26.2\scriptsize{$\pm$1.2} & 18.3\scriptsize{$\pm$1.0} & 14.5\scriptsize{$\pm$1.0} & 12.8\scriptsize{$\pm$0.8} & 11.8\scriptsize{$\pm$0.8} & 11.4\scriptsize{$\pm$0.7}\\
        & CLIP$_2$ & 97.4\scriptsize{$\pm$0.2} & 96.2\scriptsize{$\pm$0.2} & 92.7\scriptsize{$\pm$0.5} & 81.0\scriptsize{$\pm$0.6} & 58.3\scriptsize{$\pm$1.7} & 35.9\scriptsize{$\pm$1.7} & 22.9\scriptsize{$\pm$0.6} & 16.9\scriptsize{$\pm$0.4} & 14.0\scriptsize{$\pm$0.3} & 12.8\scriptsize{$\pm$0.6} & 11.8\scriptsize{$\pm$0.4}\\
        & CLIP$_3$ & 97.0\scriptsize{$\pm$0.3} & 95.7\scriptsize{$\pm$0.4} & 93.0\scriptsize{$\pm$0.7} & 85.9\scriptsize{$\pm$0.6} & 70.6\scriptsize{$\pm$0.2} & 45.9\scriptsize{$\pm$1.4} & 24.0\scriptsize{$\pm$2.1} & 15.0\scriptsize{$\pm$1.6} & 11.7\scriptsize{$\pm$0.8} & 10.8\scriptsize{$\pm$0.5} & 10.3\scriptsize{$\pm$0.2}\\
        & CLIP$_4$ & 96.1\scriptsize{$\pm$0.4} & 95.7\scriptsize{$\pm$0.5} & 92.9\scriptsize{$\pm$0.1} & 87.1\scriptsize{$\pm$0.7} & 76.9\scriptsize{$\pm$1.0} & 59.3\scriptsize{$\pm$0.5} & 38.9\scriptsize{$\pm$0.6} & 24.6\scriptsize{$\pm$1.0} & 17.0\scriptsize{$\pm$1.0} & 13.6\scriptsize{$\pm$0.6} & 12.2\scriptsize{$\pm$0.4}\\
        \midrule
        \multirow{10}{*}{\rotatebox[origin=c]{90}{ViT-B/32}} & IN$_0$ & 98.4\scriptsize{$\pm$0.3} & 69.7\scriptsize{$\pm$1.4} & 51.7\scriptsize{$\pm$1.7} & 36.1\scriptsize{$\pm$1.6} & 25.6\scriptsize{$\pm$1.5} & 17.6\scriptsize{$\pm$0.8} & 13.5\scriptsize{$\pm$0.4} & 11.8\scriptsize{$\pm$0.4} & 11.6\scriptsize{$\pm$0.7} & 11.1\scriptsize{$\pm$0.5} & 10.8\scriptsize{$\pm$0.6}\\
        & IN$_1$ & 98.1\scriptsize{$\pm$0.2} & 97.6\scriptsize{$\pm$0.2} & 94.2\scriptsize{$\pm$0.2} & 82.2\scriptsize{$\pm$1.6} & 58.4\scriptsize{$\pm$4.7} & 35.2\scriptsize{$\pm$4.7} & 23.0\scriptsize{$\pm$2.3} & 18.0\scriptsize{$\pm$1.0} & 15.1\scriptsize{$\pm$0.5} & 13.9\scriptsize{$\pm$0.4} & 13.1\scriptsize{$\pm$0.7}\\
        & IN$_2$ & 98.1\scriptsize{$\pm$0.1} & 97.7\scriptsize{$\pm$0.4} & 96.3\scriptsize{$\pm$0.3} & 92.2\scriptsize{$\pm$0.2} & 78.5\scriptsize{$\pm$0.6} & 52.0\scriptsize{$\pm$1.1} & 30.0\scriptsize{$\pm$1.5} & 19.3\scriptsize{$\pm$1.7} & 14.5\scriptsize{$\pm$1.0} & 12.5\scriptsize{$\pm$0.4} & 12.1\scriptsize{$\pm$0.5}\\
        & IN$_3$ & 98.0\scriptsize{$\pm$0.5} & 97.4\scriptsize{$\pm$0.1} & 96.5\scriptsize{$\pm$0.3} & 94.7\scriptsize{$\pm$0.5} & 85.7\scriptsize{$\pm$0.4} & 64.0\scriptsize{$\pm$1.2} & 39.4\scriptsize{$\pm$2.2} & 24.3\scriptsize{$\pm$1.8} & 17.6\scriptsize{$\pm$1.5} & 14.6\scriptsize{$\pm$1.5} & 13.4\scriptsize{$\pm$1.3}\\
        & IN$_4$ & 97.4\scriptsize{$\pm$0.6} & 96.6\scriptsize{$\pm$0.5} & 96.7\scriptsize{$\pm$0.2} & 94.3\scriptsize{$\pm$0.4} & 89.1\scriptsize{$\pm$0.6} & 75.2\scriptsize{$\pm$0.6} & 53.0\scriptsize{$\pm$2.1} & 32.4\scriptsize{$\pm$3.0} & 21.1\scriptsize{$\pm$2.8} & 15.8\scriptsize{$\pm$2.3} & 13.3\scriptsize{$\pm$1.6}\\
        & CLIP$_0$ & 98.8\scriptsize{$\pm$0.1} & 89.7\scriptsize{$\pm$1.9} & 71.2\scriptsize{$\pm$3.7} & 48.2\scriptsize{$\pm$2.3} & 31.2\scriptsize{$\pm$1.3} & 21.7\scriptsize{$\pm$1.0} & 16.3\scriptsize{$\pm$1.0} & 13.7\scriptsize{$\pm$1.0} & 12.3\scriptsize{$\pm$1.1} & 11.6\scriptsize{$\pm$1.1} & 11.2\scriptsize{$\pm$1.2}\\
        & CLIP$_1$ & 98.5\scriptsize{$\pm$0.2} & 97.8\scriptsize{$\pm$0.2} & 91.5\scriptsize{$\pm$1.0} & 69.3\scriptsize{$\pm$4.2} & 45.0\scriptsize{$\pm$5.6} & 29.9\scriptsize{$\pm$5.5} & 22.1\scriptsize{$\pm$4.1} & 17.8\scriptsize{$\pm$3.1} & 15.4\scriptsize{$\pm$2.0} & 14.2\scriptsize{$\pm$1.9} & 12.9\scriptsize{$\pm$1.3}\\
        & CLIP$_2$ & 98.7\scriptsize{$\pm$0.1} & 98.0\scriptsize{$\pm$0.4} & 95.6\scriptsize{$\pm$0.4} & 86.3\scriptsize{$\pm$0.6} & 65.7\scriptsize{$\pm$1.3} & 43.1\scriptsize{$\pm$0.6} & 29.7\scriptsize{$\pm$0.4} & 22.6\scriptsize{$\pm$0.6} & 18.8\scriptsize{$\pm$0.5} & 16.8\scriptsize{$\pm$0.1} & 15.8\scriptsize{$\pm$0.3}\\
        & CLIP$_3$ & 98.3\scriptsize{$\pm$0.2} & 97.5\scriptsize{$\pm$0.1} & 96.0\scriptsize{$\pm$0.1} & 90.4\scriptsize{$\pm$0.6} & 77.2\scriptsize{$\pm$0.4} & 56.7\scriptsize{$\pm$0.6} & 38.8\scriptsize{$\pm$0.2} & 27.6\scriptsize{$\pm$0.5} & 21.8\scriptsize{$\pm$0.4} & 18.1\scriptsize{$\pm$0.5} & 16.3\scriptsize{$\pm$0.5}\\
        & CLIP$_4$ & 98.0\scriptsize{$\pm$0.6} & 97.4\scriptsize{$\pm$0.3} & 95.8\scriptsize{$\pm$0.6} & 91.0\scriptsize{$\pm$0.4} & 81.3\scriptsize{$\pm$0.8} & 67.1\scriptsize{$\pm$0.7} & 50.1\scriptsize{$\pm$0.9} & 36.6\scriptsize{$\pm$0.3} & 27.6\scriptsize{$\pm$0.5} & 22.3\scriptsize{$\pm$0.4} & 19.1\scriptsize{$\pm$0.3}\\
        \bottomrule
    \end{tabular}}
    \caption{Linear probing results on \textsc{NoisyMNIST}.}
    
\end{table}

\begin{table}[ht]
    \centering
    \adjustbox{max width=\textwidth}{
    \begin{tabular}{llccccccccccc}
        \toprule
        & & $\mathcal{D}_0$ & $\mathcal{D}_1$ & $\mathcal{D}_2$ & $\mathcal{D}_3$  & $\mathcal{D}_4$ & $\mathcal{D}_5$ & $\mathcal{D}_6$ & $\mathcal{D}_7$ & $\mathcal{D}_8$ & $\mathcal{D}_9$ & $\mathcal{D}_{10}$ \\
        \midrule
        \multirow{10}{*}{\rotatebox[origin=c]{90}{ResNet-50}} & IN$_0$ & 98.5\scriptsize{$\pm$0.1} & 97.3\scriptsize{$\pm$0.1} & 94.7\scriptsize{$\pm$0.1} & 89.2\scriptsize{$\pm$0.2} & 80.3\scriptsize{$\pm$0.2} & 68.1\scriptsize{$\pm$0.0} & 55.0\scriptsize{$\pm$0.3} & 44.5\scriptsize{$\pm$0.4} & 38.6\scriptsize{$\pm$0.7} & 36.5\scriptsize{$\pm$0.8} & 34.3\scriptsize{$\pm$1.2}\\
        & IN$_1$ & 98.4\scriptsize{$\pm$0.1} & 98.6\scriptsize{$\pm$0.2} & 97.4\scriptsize{$\pm$0.1} & 94.5\scriptsize{$\pm$0.1} & 88.3\scriptsize{$\pm$0.2} & 77.4\scriptsize{$\pm$0.2} & 64.2\scriptsize{$\pm$0.6} & 53.2\scriptsize{$\pm$0.9} & 46.2\scriptsize{$\pm$1.1} & 42.2\scriptsize{$\pm$1.4} & 38.6\scriptsize{$\pm$1.2}\\
        & IN$_2$ & 98.2\scriptsize{$\pm$0.2} & 98.5\scriptsize{$\pm$0.3} & 98.3\scriptsize{$\pm$0.1} & 97.0\scriptsize{$\pm$0.1} & 93.8\scriptsize{$\pm$0.3} & 87.2\scriptsize{$\pm$0.5} & 77.1\scriptsize{$\pm$1.1} & 67.1\scriptsize{$\pm$0.9} & 58.6\scriptsize{$\pm$1.4} & 52.4\scriptsize{$\pm$2.0} & 48.6\scriptsize{$\pm$2.1}\\
        & IN$_3$ & 98.1\scriptsize{$\pm$0.2} & 98.6\scriptsize{$\pm$0.3} & 98.5\scriptsize{$\pm$0.1} & 98.6\scriptsize{$\pm$0.2} & 96.7\scriptsize{$\pm$0.1} & 93.4\scriptsize{$\pm$0.0} & 86.7\scriptsize{$\pm$0.2} & 78.1\scriptsize{$\pm$0.8} & 68.5\scriptsize{$\pm$1.4} & 58.9\scriptsize{$\pm$1.6} & 54.6\scriptsize{$\pm$0.7}\\
        & IN$_4$ & 97.2\scriptsize{$\pm$0.2} & 98.0\scriptsize{$\pm$0.3} & 98.4\scriptsize{$\pm$0.2} & 98.5\scriptsize{$\pm$0.6} & 97.8\scriptsize{$\pm$0.4} & 96.4\scriptsize{$\pm$0.1} & 92.9\scriptsize{$\pm$0.2} & 86.8\scriptsize{$\pm$0.1} & 77.6\scriptsize{$\pm$0.9} & 65.9\scriptsize{$\pm$1.4} & 61.3\scriptsize{$\pm$1.5}\\
        & CLIP$_0$ & 98.2\scriptsize{$\pm$0.1} & 96.6\scriptsize{$\pm$0.1} & 93.4\scriptsize{$\pm$0.2} & 83.6\scriptsize{$\pm$1.0} & 69.6\scriptsize{$\pm$1.4} & 55.4\scriptsize{$\pm$1.4} & 44.8\scriptsize{$\pm$0.7} & 40.5\scriptsize{$\pm$0.9} & 38.0\scriptsize{$\pm$1.7} & 37.9\scriptsize{$\pm$0.6} & 34.2\scriptsize{$\pm$1.9}\\
        & CLIP$_1$ & 98.0\scriptsize{$\pm$0.1} & 98.2\scriptsize{$\pm$0.1} & 96.3\scriptsize{$\pm$0.2} & 90.0\scriptsize{$\pm$0.3} & 78.6\scriptsize{$\pm$0.6} & 64.7\scriptsize{$\pm$0.9} & 53.1\scriptsize{$\pm$0.6} & 48.1\scriptsize{$\pm$0.8} & 43.0\scriptsize{$\pm$0.8} & 40.7\scriptsize{$\pm$0.7} & 37.6\scriptsize{$\pm$0.6}\\
        & CLIP$_2$ & 97.7\scriptsize{$\pm$0.1} & 98.0\scriptsize{$\pm$0.1} & 97.5\scriptsize{$\pm$0.1} & 95.4\scriptsize{$\pm$0.2} & 90.1\scriptsize{$\pm$0.2} & 81.0\scriptsize{$\pm$0.7} & 69.9\scriptsize{$\pm$0.9} & 60.7\scriptsize{$\pm$1.1} & 50.5\scriptsize{$\pm$0.7} & 44.2\scriptsize{$\pm$0.7} & 39.6\scriptsize{$\pm$1.5}\\
        & CLIP$_3$ & 97.6\scriptsize{$\pm$0.2} & 98.1\scriptsize{$\pm$0.1} & 97.9\scriptsize{$\pm$0.5} & 97.6\scriptsize{$\pm$0.4} & 95.3\scriptsize{$\pm$0.1} & 90.5\scriptsize{$\pm$0.3} & 82.1\scriptsize{$\pm$0.5} & 71.5\scriptsize{$\pm$0.6} & 58.3\scriptsize{$\pm$0.8} & 50.8\scriptsize{$\pm$0.6} & 44.1\scriptsize{$\pm$1.4}\\
        & CLIP$_4$ & 97.2\scriptsize{$\pm$0.4} & 97.6\scriptsize{$\pm$0.2} & 98.0\scriptsize{$\pm$0.5} & 97.8\scriptsize{$\pm$0.3} & 96.7\scriptsize{$\pm$0.4} & 94.7\scriptsize{$\pm$0.1} & 89.2\scriptsize{$\pm$0.2} & 80.6\scriptsize{$\pm$0.5} & 68.0\scriptsize{$\pm$0.9} & 56.4\scriptsize{$\pm$0.9} & 49.3\scriptsize{$\pm$1.0}\\
        \midrule
        \multirow{10}{*}{\rotatebox[origin=c]{90}{ViT-B/32}} & IN$_0$ & 98.3\scriptsize{$\pm$0.2} & 97.7\scriptsize{$\pm$0.0} & 96.1\scriptsize{$\pm$0.1} & 90.3\scriptsize{$\pm$0.3} & 80.8\scriptsize{$\pm$0.5} & 68.4\scriptsize{$\pm$0.7} & 55.7\scriptsize{$\pm$0.8} & 46.7\scriptsize{$\pm$0.1} & 41.2\scriptsize{$\pm$0.3} & 39.8\scriptsize{$\pm$0.6} & 38.9\scriptsize{$\pm$0.4}\\
        & IN$_1$ & 98.3\scriptsize{$\pm$0.0} & 98.6\scriptsize{$\pm$0.1} & 97.8\scriptsize{$\pm$0.1} & 95.0\scriptsize{$\pm$0.2} & 89.2\scriptsize{$\pm$0.4} & 79.5\scriptsize{$\pm$0.5} & 67.7\scriptsize{$\pm$0.6} & 58.2\scriptsize{$\pm$0.9} & 51.1\scriptsize{$\pm$0.3} & 45.9\scriptsize{$\pm$0.9} & 45.0\scriptsize{$\pm$0.9}\\
        & IN$_2$ & 98.2\scriptsize{$\pm$0.1} & 98.7\scriptsize{$\pm$0.3} & 98.7\scriptsize{$\pm$0.2} & 97.4\scriptsize{$\pm$0.1} & 94.4\scriptsize{$\pm$0.2} & 88.3\scriptsize{$\pm$0.5} & 78.0\scriptsize{$\pm$1.6} & 66.6\scriptsize{$\pm$1.9} & 57.7\scriptsize{$\pm$1.6} & 50.8\scriptsize{$\pm$1.5} & 48.9\scriptsize{$\pm$0.7}\\
        & IN$_3$ & 97.9\scriptsize{$\pm$0.2} & 98.4\scriptsize{$\pm$0.3} & 98.7\scriptsize{$\pm$0.4} & 98.6\scriptsize{$\pm$0.1} & 96.9\scriptsize{$\pm$0.1} & 94.1\scriptsize{$\pm$0.3} & 87.5\scriptsize{$\pm$0.3} & 77.7\scriptsize{$\pm$0.2} & 67.7\scriptsize{$\pm$0.6} & 59.7\scriptsize{$\pm$1.0} & 56.6\scriptsize{$\pm$0.8}\\
        & IN$_4$ & 97.4\scriptsize{$\pm$0.7} & 98.2\scriptsize{$\pm$0.5} & 99.0\scriptsize{$\pm$0.2} & 98.5\scriptsize{$\pm$0.3} & 98.4\scriptsize{$\pm$0.4} & 96.7\scriptsize{$\pm$0.1} & 92.8\scriptsize{$\pm$0.1} & 85.4\scriptsize{$\pm$0.2} & 76.7\scriptsize{$\pm$0.3} & 66.6\scriptsize{$\pm$0.1} & 62.0\scriptsize{$\pm$0.3}\\
        & CLIP$_0$ & 98.8\scriptsize{$\pm$0.1} & 97.8\scriptsize{$\pm$0.1} & 95.1\scriptsize{$\pm$0.1} & 87.5\scriptsize{$\pm$0.4} & 73.9\scriptsize{$\pm$0.7} & 58.0\scriptsize{$\pm$0.6} & 44.5\scriptsize{$\pm$0.5} & 37.6\scriptsize{$\pm$0.4} & 33.3\scriptsize{$\pm$0.1} & 32.1\scriptsize{$\pm$0.5} & 33.3\scriptsize{$\pm$0.2}\\
        & CLIP$_1$ & 98.6\scriptsize{$\pm$0.2} & 98.9\scriptsize{$\pm$0.1} & 97.7\scriptsize{$\pm$0.1} & 93.3\scriptsize{$\pm$0.3} & 84.2\scriptsize{$\pm$0.6} & 71.0\scriptsize{$\pm$0.7} & 57.1\scriptsize{$\pm$1.0} & 47.3\scriptsize{$\pm$0.9} & 39.7\scriptsize{$\pm$0.5} & 35.3\scriptsize{$\pm$0.4} & 36.9\scriptsize{$\pm$0.5}\\
        & CLIP$_2$ & 98.6\scriptsize{$\pm$0.2} & 98.9\scriptsize{$\pm$0.2} & 98.4\scriptsize{$\pm$0.2} & 96.7\scriptsize{$\pm$0.0} & 92.2\scriptsize{$\pm$0.2} & 82.5\scriptsize{$\pm$0.7} & 69.1\scriptsize{$\pm$1.5} & 57.8\scriptsize{$\pm$1.8} & 48.4\scriptsize{$\pm$1.3} & 41.0\scriptsize{$\pm$1.9} & 41.1\scriptsize{$\pm$1.0}\\
        & CLIP$_3$ & 98.0\scriptsize{$\pm$0.2} & 98.6\scriptsize{$\pm$0.1} & 98.9\scriptsize{$\pm$0.1} & 98.0\scriptsize{$\pm$0.2} & 96.2\scriptsize{$\pm$0.1} & 91.9\scriptsize{$\pm$0.2} & 82.9\scriptsize{$\pm$0.4} & 72.3\scriptsize{$\pm$0.3} & 60.8\scriptsize{$\pm$0.4} & 49.8\scriptsize{$\pm$0.6} & 47.5\scriptsize{$\pm$1.0}\\
        & CLIP$_4$ & 98.1\scriptsize{$\pm$0.5} & 98.9\scriptsize{$\pm$0.1} & 98.7\scriptsize{$\pm$0.2} & 98.4\scriptsize{$\pm$0.2} & 97.8\scriptsize{$\pm$0.4} & 95.6\scriptsize{$\pm$0.1} & 90.2\scriptsize{$\pm$0.2} & 81.5\scriptsize{$\pm$0.5} & 69.7\scriptsize{$\pm$1.2} & 56.9\scriptsize{$\pm$1.1} & 52.8\scriptsize{$\pm$1.7}\\
        \bottomrule
    \end{tabular}}
    \caption{Linear probing results on \textsc{RotatedMNIST}.}
    
\end{table}

\begin{table}[ht]
    \centering
    \adjustbox{max width=\textwidth}{
    \begin{tabular}{llccccccccccc}
        \toprule
        & & $\mathcal{D}_0$ & $\mathcal{D}_1$ & $\mathcal{D}_2$ & $\mathcal{D}_3$  & $\mathcal{D}_4$ & $\mathcal{D}_5$ & $\mathcal{D}_6$ & $\mathcal{D}_7$ & $\mathcal{D}_8$ & $\mathcal{D}_9$ & $\mathcal{D}_{10}$ \\
        \midrule
        \multirow{10}{*}{\rotatebox[origin=c]{90}{ResNet-50}} & IN$_0$ & 97.7\scriptsize{$\pm$0.1} & 92.7\scriptsize{$\pm$0.2} & 85.5\scriptsize{$\pm$0.5} & 75.5\scriptsize{$\pm$1.1} & 61.7\scriptsize{$\pm$1.3} & 47.8\scriptsize{$\pm$1.0} & 37.1\scriptsize{$\pm$1.0} & 29.4\scriptsize{$\pm$1.5} & 23.9\scriptsize{$\pm$0.5} & 20.2\scriptsize{$\pm$0.8} & 17.2\scriptsize{$\pm$1.1}\\
        & IN$_1$ & 98.1\scriptsize{$\pm$0.5} & 96.2\scriptsize{$\pm$0.6} & 88.2\scriptsize{$\pm$0.8} & 78.2\scriptsize{$\pm$0.8} & 65.9\scriptsize{$\pm$0.9} & 52.4\scriptsize{$\pm$0.5} & 39.7\scriptsize{$\pm$1.1} & 32.1\scriptsize{$\pm$1.2} & 25.9\scriptsize{$\pm$0.6} & 20.1\scriptsize{$\pm$0.6} & 18.1\scriptsize{$\pm$1.0}\\
        & IN$_2$ & 97.1\scriptsize{$\pm$0.6} & 96.6\scriptsize{$\pm$0.3} & 90.7\scriptsize{$\pm$0.4} & 81.7\scriptsize{$\pm$0.8} & 71.7\scriptsize{$\pm$0.3} & 59.4\scriptsize{$\pm$1.6} & 47.8\scriptsize{$\pm$1.3} & 37.5\scriptsize{$\pm$1.2} & 29.0\scriptsize{$\pm$0.9} & 23.0\scriptsize{$\pm$1.9} & 19.3\scriptsize{$\pm$0.7}\\
        & IN$_3$ & 97.4\scriptsize{$\pm$0.5} & 96.0\scriptsize{$\pm$0.7} & 92.6\scriptsize{$\pm$1.4} & 85.5\scriptsize{$\pm$0.7} & 75.5\scriptsize{$\pm$1.5} & 65.6\scriptsize{$\pm$1.5} & 52.8\scriptsize{$\pm$0.6} & 41.7\scriptsize{$\pm$0.4} & 34.1\scriptsize{$\pm$0.8} & 26.4\scriptsize{$\pm$1.6} & 20.8\scriptsize{$\pm$1.0}\\
        & IN$_4$ & 97.7\scriptsize{$\pm$0.2} & 96.0\scriptsize{$\pm$0.3} & 93.1\scriptsize{$\pm$0.8} & 85.6\scriptsize{$\pm$2.4} & 76.6\scriptsize{$\pm$1.8} & 67.1\scriptsize{$\pm$0.4} & 56.6\scriptsize{$\pm$0.9} & 47.1\scriptsize{$\pm$2.0} & 39.2\scriptsize{$\pm$0.7} & 31.7\scriptsize{$\pm$0.6} & 25.2\scriptsize{$\pm$0.8}\\
        & CLIP$_0$ & 93.8\scriptsize{$\pm$0.3} & 87.7\scriptsize{$\pm$0.3} & 77.4\scriptsize{$\pm$0.9} & 62.5\scriptsize{$\pm$1.5} & 46.5\scriptsize{$\pm$1.4} & 34.5\scriptsize{$\pm$0.4} & 26.5\scriptsize{$\pm$1.6} & 19.9\scriptsize{$\pm$0.9} & 16.0\scriptsize{$\pm$1.4} & 14.5\scriptsize{$\pm$1.1} & 12.4\scriptsize{$\pm$1.0}\\
        & CLIP$_1$ & 93.4\scriptsize{$\pm$0.9} & 90.5\scriptsize{$\pm$0.7} & 81.4\scriptsize{$\pm$1.0} & 67.8\scriptsize{$\pm$0.2} & 50.0\scriptsize{$\pm$1.2} & 33.5\scriptsize{$\pm$1.9} & 23.2\scriptsize{$\pm$0.7} & 15.7\scriptsize{$\pm$0.4} & 12.2\scriptsize{$\pm$0.4} & 10.2\scriptsize{$\pm$0.6} & 8.7\scriptsize{$\pm$0.5}\\
        & CLIP$_2$ & 93.9\scriptsize{$\pm$0.7} & 90.9\scriptsize{$\pm$0.6} & 84.4\scriptsize{$\pm$0.5} & 76.2\scriptsize{$\pm$1.1} & 64.2\scriptsize{$\pm$1.1} & 49.3\scriptsize{$\pm$1.9} & 34.2\scriptsize{$\pm$1.6} & 23.9\scriptsize{$\pm$1.2} & 18.0\scriptsize{$\pm$1.0} & 13.2\scriptsize{$\pm$0.8} & 10.9\scriptsize{$\pm$1.0}\\
        & CLIP$_3$ & 93.0\scriptsize{$\pm$0.8} & 90.5\scriptsize{$\pm$1.2} & 85.5\scriptsize{$\pm$1.8} & 77.6\scriptsize{$\pm$1.3} & 68.6\scriptsize{$\pm$1.0} & 56.9\scriptsize{$\pm$1.1} & 43.8\scriptsize{$\pm$0.5} & 31.3\scriptsize{$\pm$1.7} & 22.7\scriptsize{$\pm$1.2} & 17.2\scriptsize{$\pm$0.9} & 14.0\scriptsize{$\pm$0.4}\\
        & CLIP$_4$ & 92.9\scriptsize{$\pm$1.3} & 91.0\scriptsize{$\pm$1.6} & 84.8\scriptsize{$\pm$1.5} & 76.8\scriptsize{$\pm$1.8} & 71.0\scriptsize{$\pm$1.4} & 60.6\scriptsize{$\pm$1.8} & 52.3\scriptsize{$\pm$1.3} & 40.7\scriptsize{$\pm$1.1} & 32.3\scriptsize{$\pm$1.7} & 24.4\scriptsize{$\pm$0.7} & 19.0\scriptsize{$\pm$0.6}\\
        \midrule
        \multirow{10}{*}{\rotatebox[origin=c]{90}{ViT-B/32}} & IN$_0$ & 98.3\scriptsize{$\pm$0.3} & 92.9\scriptsize{$\pm$0.2} & 92.2\scriptsize{$\pm$0.7} & 90.9\scriptsize{$\pm$0.3} & 89.3\scriptsize{$\pm$0.3} & 85.9\scriptsize{$\pm$0.8} & 82.2\scriptsize{$\pm$0.4} & 77.9\scriptsize{$\pm$0.5} & 71.8\scriptsize{$\pm$0.4} & 67.7\scriptsize{$\pm$1.5} & 61.5\scriptsize{$\pm$0.9}\\
        & IN$_1$ & 98.4\scriptsize{$\pm$0.3} & 98.4\scriptsize{$\pm$0.3} & 92.6\scriptsize{$\pm$0.4} & 91.5\scriptsize{$\pm$0.3} & 89.0\scriptsize{$\pm$0.7} & 86.6\scriptsize{$\pm$0.6} & 82.6\scriptsize{$\pm$0.8} & 76.7\scriptsize{$\pm$1.4} & 72.3\scriptsize{$\pm$1.3} & 65.7\scriptsize{$\pm$1.0} & 59.6\scriptsize{$\pm$1.2}\\
        & IN$_2$ & 98.2\scriptsize{$\pm$0.4} & 98.3\scriptsize{$\pm$0.3} & 96.9\scriptsize{$\pm$0.3} & 90.7\scriptsize{$\pm$0.2} & 89.0\scriptsize{$\pm$0.3} & 87.1\scriptsize{$\pm$0.6} & 81.4\scriptsize{$\pm$0.9} & 77.8\scriptsize{$\pm$1.2} & 71.3\scriptsize{$\pm$0.2} & 64.2\scriptsize{$\pm$0.9} & 57.7\scriptsize{$\pm$2.0}\\
        & IN$_3$ & 98.5\scriptsize{$\pm$0.5} & 98.7\scriptsize{$\pm$0.6} & 97.4\scriptsize{$\pm$0.8} & 95.9\scriptsize{$\pm$0.8} & 89.9\scriptsize{$\pm$0.3} & 86.7\scriptsize{$\pm$0.9} & 83.2\scriptsize{$\pm$0.3} & 79.8\scriptsize{$\pm$0.9} & 73.7\scriptsize{$\pm$0.8} & 67.8\scriptsize{$\pm$0.8} & 60.9\scriptsize{$\pm$0.5}\\
        & IN$_4$ & 98.8\scriptsize{$\pm$0.3} & 98.5\scriptsize{$\pm$0.8} & 97.6\scriptsize{$\pm$0.2} & 95.5\scriptsize{$\pm$1.2} & 92.9\scriptsize{$\pm$1.4} & 87.2\scriptsize{$\pm$0.5} & 84.7\scriptsize{$\pm$0.1} & 79.2\scriptsize{$\pm$1.3} & 75.3\scriptsize{$\pm$0.7} & 69.1\scriptsize{$\pm$0.8} & 62.6\scriptsize{$\pm$0.6}\\
        & CLIP$_0$ & 94.4\scriptsize{$\pm$0.1} & 92.1\scriptsize{$\pm$0.3} & 88.8\scriptsize{$\pm$0.6} & 81.9\scriptsize{$\pm$0.4} & 72.7\scriptsize{$\pm$1.1} & 62.9\scriptsize{$\pm$1.0} & 53.7\scriptsize{$\pm$1.0} & 45.5\scriptsize{$\pm$1.1} & 39.7\scriptsize{$\pm$1.7} & 32.3\scriptsize{$\pm$1.8} & 27.6\scriptsize{$\pm$0.7}\\
        & CLIP$_1$ & 94.6\scriptsize{$\pm$0.5} & 93.7\scriptsize{$\pm$0.4} & 90.0\scriptsize{$\pm$0.4} & 85.5\scriptsize{$\pm$0.6} & 76.6\scriptsize{$\pm$1.7} & 69.7\scriptsize{$\pm$2.6} & 60.2\scriptsize{$\pm$2.2} & 50.9\scriptsize{$\pm$4.0} & 42.4\scriptsize{$\pm$3.2} & 35.0\scriptsize{$\pm$1.8} & 28.9\scriptsize{$\pm$2.9}\\
        & CLIP$_2$ & 94.5\scriptsize{$\pm$0.8} & 93.8\scriptsize{$\pm$0.6} & 91.8\scriptsize{$\pm$0.6} & 87.0\scriptsize{$\pm$0.4} & 81.4\scriptsize{$\pm$0.9} & 73.5\scriptsize{$\pm$0.7} & 65.4\scriptsize{$\pm$2.3} & 55.0\scriptsize{$\pm$1.3} & 44.5\scriptsize{$\pm$1.7} & 38.1\scriptsize{$\pm$1.4} & 30.3\scriptsize{$\pm$1.5}\\
        & CLIP$_3$ & 94.8\scriptsize{$\pm$1.1} & 94.6\scriptsize{$\pm$0.7} & 90.7\scriptsize{$\pm$0.6} & 87.4\scriptsize{$\pm$1.2} & 82.3\scriptsize{$\pm$0.9} & 75.7\scriptsize{$\pm$1.0} & 69.1\scriptsize{$\pm$0.6} & 60.8\scriptsize{$\pm$0.6} & 51.9\scriptsize{$\pm$1.9} & 42.8\scriptsize{$\pm$1.3} & 35.7\scriptsize{$\pm$0.8}\\
        & CLIP$_4$ & 94.4\scriptsize{$\pm$1.3} & 94.2\scriptsize{$\pm$1.0} & 90.4\scriptsize{$\pm$0.5} & 86.2\scriptsize{$\pm$1.5} & 82.1\scriptsize{$\pm$1.4} & 77.7\scriptsize{$\pm$1.1} & 71.5\scriptsize{$\pm$0.2} & 64.4\scriptsize{$\pm$1.5} & 55.2\scriptsize{$\pm$1.1} & 46.6\scriptsize{$\pm$1.3} & 40.7\scriptsize{$\pm$1.8}\\
        \bottomrule
    \end{tabular}}
    \caption{Linear probing results on \textsc{NoisyImageNet15}.}
    
\end{table}

\begin{table}[ht]
    \centering
    \adjustbox{max width=\textwidth}{
    \begin{tabular}{llccccccccccc}
        \toprule
        & & $\mathcal{D}_0$ & $\mathcal{D}_1$ & $\mathcal{D}_2$ & $\mathcal{D}_3$  & $\mathcal{D}_4$ & $\mathcal{D}_5$ & $\mathcal{D}_6$ & $\mathcal{D}_7$ & $\mathcal{D}_8$ & $\mathcal{D}_9$ & $\mathcal{D}_{10}$ \\
        \midrule
        \multirow{10}{*}{\rotatebox[origin=c]{90}{ResNet-50}} & IN$_0$ & 97.9\scriptsize{$\pm$0.2} & 93.8\scriptsize{$\pm$0.5} & 91.6\scriptsize{$\pm$0.3} & 87.9\scriptsize{$\pm$0.5} & 84.1\scriptsize{$\pm$0.5} & 78.5\scriptsize{$\pm$1.3} & 71.5\scriptsize{$\pm$2.0} & 64.2\scriptsize{$\pm$2.8} & 55.8\scriptsize{$\pm$3.6} & 43.4\scriptsize{$\pm$4.6} & 32.9\scriptsize{$\pm$2.9}\\
        & IN$_1$ & 98.0\scriptsize{$\pm$0.3} & 97.4\scriptsize{$\pm$0.4} & 92.8\scriptsize{$\pm$0.3} & 90.3\scriptsize{$\pm$0.5} & 87.9\scriptsize{$\pm$0.3} & 82.9\scriptsize{$\pm$0.3} & 76.4\scriptsize{$\pm$1.2} & 69.6\scriptsize{$\pm$1.5} & 59.3\scriptsize{$\pm$2.0} & 50.3\scriptsize{$\pm$1.8} & 39.7\scriptsize{$\pm$1.6}\\
        & IN$_2$ & 96.8\scriptsize{$\pm$0.4} & 97.3\scriptsize{$\pm$0.2} & 96.4\scriptsize{$\pm$0.4} & 91.4\scriptsize{$\pm$0.3} & 89.4\scriptsize{$\pm$0.1} & 85.6\scriptsize{$\pm$0.5} & 81.0\scriptsize{$\pm$0.8} & 75.7\scriptsize{$\pm$0.5} & 67.4\scriptsize{$\pm$1.4} & 57.0\scriptsize{$\pm$1.1} & 45.8\scriptsize{$\pm$0.8}\\
        & IN$_3$ & 97.4\scriptsize{$\pm$0.6} & 96.7\scriptsize{$\pm$0.7} & 97.6\scriptsize{$\pm$0.7} & 95.3\scriptsize{$\pm$1.0} & 89.7\scriptsize{$\pm$0.3} & 86.4\scriptsize{$\pm$0.6} & 82.5\scriptsize{$\pm$0.2} & 78.1\scriptsize{$\pm$0.4} & 69.0\scriptsize{$\pm$0.9} & 58.1\scriptsize{$\pm$0.9} & 47.4\scriptsize{$\pm$0.3}\\
        & IN$_4$ & 97.3\scriptsize{$\pm$0.4} & 97.4\scriptsize{$\pm$1.1} & 96.6\scriptsize{$\pm$0.5} & 95.3\scriptsize{$\pm$0.3} & 94.6\scriptsize{$\pm$0.9} & 87.1\scriptsize{$\pm$0.4} & 82.6\scriptsize{$\pm$0.3} & 78.4\scriptsize{$\pm$0.9} & 69.9\scriptsize{$\pm$0.9} & 58.2\scriptsize{$\pm$1.2} & 46.5\scriptsize{$\pm$2.1}\\
        & CLIP$_0$ & 93.5\scriptsize{$\pm$0.3} & 90.8\scriptsize{$\pm$0.7} & 89.2\scriptsize{$\pm$0.5} & 86.1\scriptsize{$\pm$0.8} & 83.0\scriptsize{$\pm$0.2} & 78.3\scriptsize{$\pm$0.8} & 71.3\scriptsize{$\pm$1.5} & 60.2\scriptsize{$\pm$1.5} & 49.2\scriptsize{$\pm$1.4} & 38.0\scriptsize{$\pm$1.9} & 27.3\scriptsize{$\pm$2.5}\\
        & CLIP$_1$ & 93.3\scriptsize{$\pm$0.3} & 92.4\scriptsize{$\pm$0.2} & 90.7\scriptsize{$\pm$0.4} & 88.8\scriptsize{$\pm$0.5} & 86.0\scriptsize{$\pm$0.7} & 81.3\scriptsize{$\pm$0.4} & 73.8\scriptsize{$\pm$0.5} & 64.9\scriptsize{$\pm$0.9} & 54.5\scriptsize{$\pm$1.2} & 44.3\scriptsize{$\pm$2.2} & 34.3\scriptsize{$\pm$1.3}\\
        & CLIP$_2$ & 93.7\scriptsize{$\pm$1.4} & 92.7\scriptsize{$\pm$0.8} & 91.8\scriptsize{$\pm$1.1} & 88.7\scriptsize{$\pm$0.1} & 85.7\scriptsize{$\pm$0.9} & 81.7\scriptsize{$\pm$0.7} & 74.1\scriptsize{$\pm$0.4} & 63.6\scriptsize{$\pm$1.6} & 52.5\scriptsize{$\pm$1.2} & 42.1\scriptsize{$\pm$0.8} & 31.1\scriptsize{$\pm$2.1}\\
        & CLIP$_3$ & 93.2\scriptsize{$\pm$0.3} & 92.2\scriptsize{$\pm$0.9} & 92.3\scriptsize{$\pm$0.3} & 91.0\scriptsize{$\pm$2.0} & 86.4\scriptsize{$\pm$0.7} & 83.0\scriptsize{$\pm$0.4} & 75.8\scriptsize{$\pm$0.3} & 66.3\scriptsize{$\pm$1.2} & 55.4\scriptsize{$\pm$1.9} & 44.7\scriptsize{$\pm$1.0} & 33.1\scriptsize{$\pm$1.8}\\
        & CLIP$_4$ & 92.2\scriptsize{$\pm$0.5} & 93.4\scriptsize{$\pm$0.9} & 91.5\scriptsize{$\pm$0.2} & 91.3\scriptsize{$\pm$1.4} & 89.4\scriptsize{$\pm$1.0} & 83.5\scriptsize{$\pm$0.2} & 77.3\scriptsize{$\pm$0.9} & 68.4\scriptsize{$\pm$0.3} & 58.0\scriptsize{$\pm$0.6} & 46.7\scriptsize{$\pm$0.5} & 34.2\scriptsize{$\pm$1.1}\\
        \midrule
        \multirow{10}{*}{\rotatebox[origin=c]{90}{ViT-B/32}} & IN$_0$ & 98.5\scriptsize{$\pm$0.2} & 94.2\scriptsize{$\pm$0.1} & 93.6\scriptsize{$\pm$0.2} & 92.2\scriptsize{$\pm$0.1} & 90.3\scriptsize{$\pm$0.2} & 88.4\scriptsize{$\pm$0.3} & 85.6\scriptsize{$\pm$0.1} & 81.2\scriptsize{$\pm$0.8} & 77.0\scriptsize{$\pm$1.1} & 70.9\scriptsize{$\pm$1.3} & 59.8\scriptsize{$\pm$1.4}\\
        & IN$_1$ & 98.1\scriptsize{$\pm$0.4} & 98.6\scriptsize{$\pm$0.3} & 93.7\scriptsize{$\pm$0.1} & 92.4\scriptsize{$\pm$0.1} & 91.1\scriptsize{$\pm$0.3} & 89.2\scriptsize{$\pm$0.2} & 86.5\scriptsize{$\pm$0.2} & 83.4\scriptsize{$\pm$0.3} & 79.5\scriptsize{$\pm$0.2} & 73.6\scriptsize{$\pm$0.3} & 62.7\scriptsize{$\pm$0.5}\\
        & IN$_2$ & 98.2\scriptsize{$\pm$0.8} & 98.3\scriptsize{$\pm$0.4} & 97.9\scriptsize{$\pm$0.3} & 92.8\scriptsize{$\pm$0.2} & 91.0\scriptsize{$\pm$0.2} & 89.2\scriptsize{$\pm$0.2} & 86.9\scriptsize{$\pm$0.2} & 84.3\scriptsize{$\pm$0.2} & 80.5\scriptsize{$\pm$0.0} & 74.6\scriptsize{$\pm$0.3} & 63.6\scriptsize{$\pm$0.7}\\
        & IN$_3$ & 98.5\scriptsize{$\pm$1.3} & 98.3\scriptsize{$\pm$0.8} & 98.4\scriptsize{$\pm$0.0} & 98.2\scriptsize{$\pm$0.2} & 91.3\scriptsize{$\pm$0.4} & 90.0\scriptsize{$\pm$0.5} & 87.4\scriptsize{$\pm$0.3} & 83.9\scriptsize{$\pm$0.0} & 80.4\scriptsize{$\pm$0.4} & 74.4\scriptsize{$\pm$0.3} & 63.6\scriptsize{$\pm$0.5}\\
        & IN$_4$ & 98.9\scriptsize{$\pm$0.3} & 98.3\scriptsize{$\pm$0.5} & 98.3\scriptsize{$\pm$0.7} & 98.3\scriptsize{$\pm$0.1} & 97.6\scriptsize{$\pm$0.6} & 90.6\scriptsize{$\pm$0.1} & 88.1\scriptsize{$\pm$0.3} & 85.1\scriptsize{$\pm$0.4} & 81.1\scriptsize{$\pm$0.5} & 75.2\scriptsize{$\pm$0.3} & 66.2\scriptsize{$\pm$0.7}\\
        & CLIP$_0$ & 94.5\scriptsize{$\pm$0.3} & 93.3\scriptsize{$\pm$0.1} & 92.1\scriptsize{$\pm$0.4} & 91.0\scriptsize{$\pm$0.1} & 89.9\scriptsize{$\pm$0.4} & 88.3\scriptsize{$\pm$0.3} & 85.7\scriptsize{$\pm$0.3} & 80.0\scriptsize{$\pm$0.7} & 73.6\scriptsize{$\pm$1.0} & 62.7\scriptsize{$\pm$1.2} & 54.1\scriptsize{$\pm$1.9}\\
        & CLIP$_1$ & 94.1\scriptsize{$\pm$0.7} & 94.8\scriptsize{$\pm$0.5} & 92.4\scriptsize{$\pm$0.1} & 91.6\scriptsize{$\pm$0.1} & 90.3\scriptsize{$\pm$0.1} & 88.6\scriptsize{$\pm$0.8} & 86.2\scriptsize{$\pm$1.8} & 79.3\scriptsize{$\pm$1.6} & 73.0\scriptsize{$\pm$2.1} & 63.2\scriptsize{$\pm$1.4} & 54.0\scriptsize{$\pm$0.8}\\
        & CLIP$_2$ & 94.8\scriptsize{$\pm$1.1} & 94.2\scriptsize{$\pm$1.0} & 93.9\scriptsize{$\pm$1.0} & 92.1\scriptsize{$\pm$0.1} & 91.0\scriptsize{$\pm$0.2} & 89.9\scriptsize{$\pm$0.3} & 86.6\scriptsize{$\pm$0.3} & 79.8\scriptsize{$\pm$0.3} & 74.0\scriptsize{$\pm$0.3} & 64.6\scriptsize{$\pm$0.5} & 55.2\scriptsize{$\pm$0.5}\\
        & CLIP$_3$ & 94.0\scriptsize{$\pm$0.7} & 94.4\scriptsize{$\pm$0.9} & 94.9\scriptsize{$\pm$0.3} & 94.1\scriptsize{$\pm$0.8} & 90.8\scriptsize{$\pm$0.1} & 89.3\scriptsize{$\pm$0.4} & 86.0\scriptsize{$\pm$1.2} & 80.0\scriptsize{$\pm$0.9} & 73.7\scriptsize{$\pm$1.4} & 64.4\scriptsize{$\pm$1.1} & 55.1\scriptsize{$\pm$2.6}\\
        & CLIP$_4$ & 94.9\scriptsize{$\pm$0.4} & 93.6\scriptsize{$\pm$1.0} & 94.3\scriptsize{$\pm$1.2} & 92.9\scriptsize{$\pm$0.2} & 92.1\scriptsize{$\pm$0.6} & 90.0\scriptsize{$\pm$0.2} & 86.4\scriptsize{$\pm$0.3} & 81.0\scriptsize{$\pm$0.7} & 75.6\scriptsize{$\pm$0.6} & 64.6\scriptsize{$\pm$0.3} & 56.2\scriptsize{$\pm$0.3}\\
        \bottomrule
    \end{tabular}}
    \caption{Linear probing results on \textsc{LR-ImageNet15}.}
    
\end{table}

\end{document}